\documentclass[runningheads]{llncs}
\usepackage{graphicx}

\usepackage{tikz}
\usepackage{comment}
\usepackage{amsmath,amssymb, bm} 

\usepackage[accsupp]{axessibility}  

\usepackage[width=122mm,left=12mm,paperwidth=146mm,height=193mm,top=12mm,paperheight=217mm]{geometry} 

\usepackage{epsfig}
\usepackage{xcolor}
\usepackage{graphicx}
\usepackage{float}
\usepackage[utf8]{inputenc}

\usepackage{caption}
\usepackage{subcaption}
\usepackage{array}
\usepackage{algorithm}
\usepackage{algpseudocode}
\usepackage{multirow}
\usepackage{etoolbox}

\newtoggle{arXiv}

\usepackage[pagebackref=true,breaklinks=true,colorlinks,bookmarks=false]{hyperref}
\usepackage{cleveref}

\usepackage{cite}

\usepackage{siunitx}
\def\onedot{.\ }

\def\eg{\emph{e.g}\onedot} 
\def\ie{\emph{i.e}\onedot}

\def\wrt{w.r.t\onedot}

\newcommand{\boldparagraph}[1]{\vspace{0.2cm}\noindent{\bf #1:}}
\def\wrt{wrt\onedot}

\begin{document}
\pagestyle{headings}
\mainmatter
\def\ECCVSubNumber{1469}  

\toggletrue{arXiv}

\title{ARAH: Animatable Volume Rendering of Articulated Human SDFs} 

\titlerunning{ARAH: Animatable Volume Rendering of Articulated Human SDFs}
%
\author{Shaofei Wang\inst{1} \and
Katja Schwarz\inst{2,3} \and
Andreas Geiger\inst{2,3} \and
Siyu Tang \inst{1}}
\authorrunning{Wang et al.}
%
\institute{ETH Z\"{u}rich \and
Max Planck Institute for Intelligent Systems, T\"{u}bingen \and
University of T\"{u}bingen}
\maketitle

\begin{abstract}
Combining human body models with differentiable rendering has recently enabled animatable avatars of clothed humans from sparse sets of multi-view RGB videos.
While state-of-the-art approaches achieve a realistic appearance with neural radiance fields (NeRF), the inferred geometry often lacks detail due to missing geometric constraints.
Further, animating avatars in out-of-distribution poses is not yet possible because the mapping from observation space to canonical space does not generalize faithfully to unseen poses.
In this work, we address these shortcomings and propose a model to create animatable clothed human avatars with detailed geometry that generalize well to out-of-distribution poses.
To achieve detailed geometry, we combine an articulated implicit surface representation with volume rendering. 
For generalization, we propose a novel joint root-finding algorithm for simultaneous ray-surface intersection search and correspondence search. 
Our algorithm enables efficient point sampling and accurate point canonicalization while generalizing well to unseen poses.
We demonstrate that our proposed pipeline can generate clothed avatars with high-quality pose-dependent geometry and appearance from a sparse set of multi-view RGB videos.
Our method achieves state-of-the-art performance on geometry and appearance reconstruction while creating animatable avatars that generalize well to out-of-distribution poses beyond the small number of training poses.

\end{abstract}

\section{Introduction}
\label{sec:intro}
\begin{figure}[t]
\centering
  \includegraphics[width=1.0\textwidth]{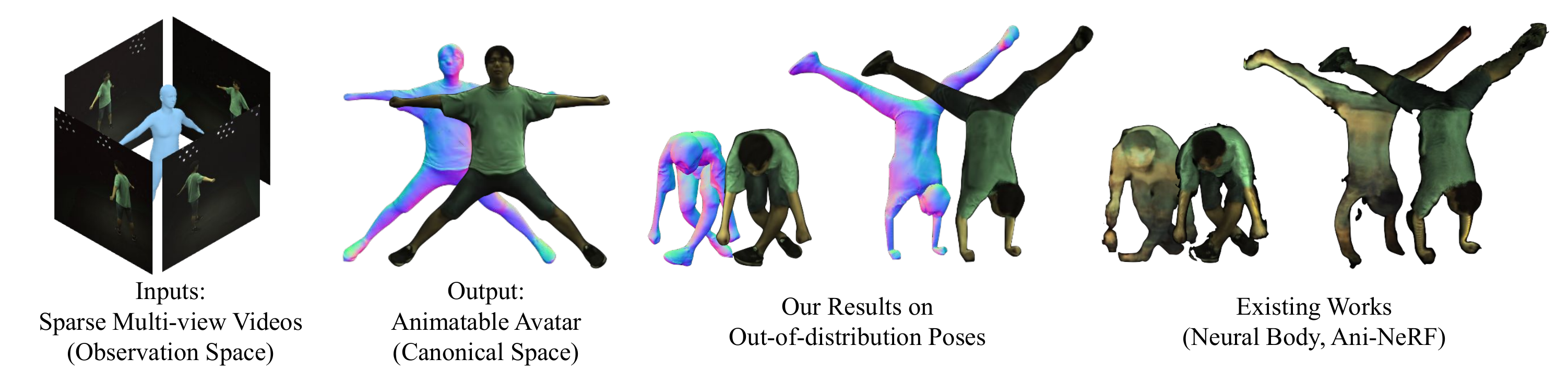}
  \caption{\textbf{Detailed Geometry and Generalization to Extreme Poses.} Given sparse multi-view videos with SMPL fittings and foreground masks, our approach synthesizes animatable clothed avatars with realistic pose-dependent geometry and appearance. While existing works, \eg\ Neural Body~\cite{peng2020neural} and Ani-NeRF~\cite{peng2021animatable}, struggle with generalizing to unseen poses, our approach enables avatars that can be animated in extreme out-of-distribution poses.}
  \label{fig:teaser}
\end{figure}
Reconstruction and animation of clothed human avatars is a rising topic in computer vision research. It is of particular interest for various applications in AR/VR and the future metaverse. Various sensors can be used to create clothed human avatars, ranging from 4D scanners over depth sensors to simple RGB cameras. 
Among these data sources, RGB videos are by far the most accessible and user-friendly choice. However, they also provide the least supervision, making this setup the most challenging for the reconstruction and animation of clothed humans.

Traditional works in clothed human modeling use explicit mesh~\cite{alldieck2019learning,alldieck2018video,Bhatnagar_ECCV2020,bhatnagar2020loopreg,deepwrinkles:2018:ECCV,Yang:ECCV:18,BUFF,Guan12drape:dressing,gundogdu19garnet,patel20tailornet,santesteban2021garmentcollisions,PhysCloth:3DV:2021,Tiwari_2021_ICCV} or truncated signed distance fields (TSDFs) of fixed grid resolution~\cite{DoubleFusion_CVPR_2018,RobustFusion_ECCV_2020,UnstructuredFusion_PAMI_2020,Li2020portrait,li2021posefusion} to represent the geometry of humans. Textures are often represented by vertex colors or UV-maps.
With the recent success of neural implicit representations, significant progress has been made towards modeling articulated clothed humans. PIFu~\cite{Saito_ICCV2019} and PIFuHD~\cite{Saito_CVPR2020} are among the first works that propose to model clothed humans as continuous neural implicit functions. ARCH~\cite{ARCH_CVPR_2020} extends this idea and develops animatable clothed human avatars from monocular images. However, this line of works does not handle dynamic pose-dependent cloth deformations. Further, they require ground-truth geometry for training. Such ground-truth data is expensive to acquire, limiting the generalization of these methods.

Another line of works removes the need for ground-truth geometry by utilizing differentiable neural rendering. These methods aim to reconstruct humans from a sparse set of multi-view videos with only image supervision. Many of them use NeRF~\cite{mildenhall2020nerf} as the underlying representation and achieve impressive visual fidelity on novel view synthesis tasks. However, there are two fundamental drawbacks of these existing approaches: (1) the NeRF-based representation lacks proper geometric regularization, leading to inaccurate geometry. This is particularly detrimental in a sparse multi-view setup and often results in artifacts in the form of erroneous color blobs under novel views or poses. (2) Existing approaches condition their NeRF networks~\cite{peng2020neural} or canonicalization networks~\cite{peng2021animatable} on inputs in observation space. Thus, they cannot generalize to unseen out-of-distribution poses.

In this work, we address these two major drawbacks of existing approaches. (1) We improve geometry by building an articulated signed-distance-field (SDF) representation for clothed human bodies to better capture the geometry of clothed humans and improve the rendering quality. (2) In order to render the SDF, we develop an efficient joint root-finding algorithm for the conversion from observation space to canonical space. Specifically, we represent clothed human avatars as a combination of a forward linear blend skinning (LBS) network, an implicit SDF network, and a color network, all defined in canonical space and do not condition on inputs in observation space. Given these networks and camera rays in observation space, we apply our novel joint root-finding algorithm that can efficiently find the iso-surface points in observation space and their correspondences in canonical space. This enables us to perform efficient sampling on camera rays around the iso-surface. All network modules can be trained with a photometric loss in image space and regularization losses in canonical space.

We validate our approach on the ZJU-MoCap~\cite{peng2020neural} and the H36M~\cite{h36m_pami} dataset. 
Our approach generalizes well to unseen poses, enabling robust animation of clothed avatars even under out-of-distribution poses where existing works fail, as shown in Fig.~\ref{fig:teaser}.
We achieve significant improvements over state-of-the-arts for novel pose synthesis and geometry reconstruction, while also outperforming state-of-the-arts in the novel view synthesis task on training poses. Code and data are available at \href{https://neuralbodies.github.io/arah/}{\color{black}{https://neuralbodies.github.io/arah/}}.
\section{Related Works}
\label{sec:related}
\boldparagraph{Clothed Human Modeling with Explicit Representations} Many explicit mesh-based approaches represent cloth deformations as deformation layers~\cite{alldieck2019learning,alldieck2018video,Bhatnagar_ECCV2020,bhatnagar2020loopreg,DSFN:ICCV:2021} added to minimally clothed parametric human body models~\cite{SCAPE,Hasler2009CGF,Joo_2018_CVPR,SMPL:2015,SMPL-X:2019,Xu_2020_CVPR,STAR:ECCV:2020}. Such approaches enjoy compatibility with parametric human body models but have difficulties in modeling large garment deformations. Other mesh-based approaches model garments as separate meshes~\cite{deepwrinkles:2018:ECCV,Yang:ECCV:18,BUFF,Guan12drape:dressing,gundogdu19garnet,patel20tailornet,santesteban2021garmentcollisions,PhysCloth:3DV:2021,Tiwari_2021_ICCV} in order to represent more detailed and physically plausible cloth deformations. However, such methods often require accurate 3D-surface registration, synthetic 3D data or dense multi-view images for training and the garment meshes need to be pre-defined for each cloth type. More recently, point-cloud-based explicit methods~\cite{SCALE:CVPR:21,POP:ICCV:2021,Zakharkin_2021_ICCV} also showed promising results in modeling clothed humans. However, they still require explicit 3D or depth supervision for training, while our goal is to train using sparse multi-view RGB supervision alone.

\boldparagraph{Clothed Humans as Implicit Functions} Neural implicit functions~\cite{chen2018implicit_decoder,Occupancy_Networks,Michalkiewicz_2019_ICCV,DeepSDF,Peng2021SAP} have been used to model clothed humans from various sensor inputs including monocular images~\cite{ARCH_CVPR_2020,ARCH++:ICCV:2021,li2020monocular,tong2020geo-pifu,Saito_ICCV2019,Saito_CVPR2020,raj2020anr,pamir2020,xiu2022icon,ANeRF:NeurIPS:2021}, multi-view videos~\cite{peng2020neural,peng2021animatable,kwon2021neural,HNeRF:NeurIPS:2021,NARF:ICCV:2021,liu2021neural}, sparse point clouds~\cite{Bhatnagar_ECCV2020,IFNet,PTF:CVPR:2021,zuo2021selfsupervised,MetaAvatar:NeurIPS:2021,dong2022pina}, or 3D meshes~\cite{Chen2021ICCV,SCANimate:CVPR:21,tiwari21neuralgif,corona2021smplicit,chen2022gdna,LEAP:CVPR:21,Mihajlovic:CVPR:2022}. Among the image-based methods,~\cite{ARCH_CVPR_2020,ARCH++:ICCV:2021,alldieck2022phorhum} obtain animatable reconstructions of clothed humans from a single image. However, they do not model pose-dependent cloth deformations and require ground-truth geometry for training.~\cite{kwon2021neural} learns generalizable NeRF models for human performance capture and only requires multi-view images as supervision. But it needs images as inputs for synthesizing novel poses.~\cite{peng2020neural,peng2021animatable,HNeRF:NeurIPS:2021,NARF:ICCV:2021,liu2021neural} take multi-view videos as inputs and do not need ground-truth geometry during training. These methods generate personalized per-subject avatars and only need 2D supervision. Our approach follows this line of work and also learns a personalized avatar for each subject.

\boldparagraph{Neural Rendering of Animatable Clothed Humans} Differentiable neural rendering has been extended to model animatable human bodies by a number of recent works~\cite{ANeRF:NeurIPS:2021,HNeRF:NeurIPS:2021,peng2020neural,peng2021animatable,NARF:ICCV:2021,SMPLpix:WACV:2020}. 
Neural Body~\cite{peng2020neural} proposes to diffuse latent per-vertex codes associated with SMPL meshes in observation space and condition NeRF~\cite{mildenhall2020nerf} on such latent codes.
However, the conditional inputs of Neural Body are in the observation space. Therefore, it does not generalize well to out-of-distribution poses. Several recent works~\cite{peng2021animatable,NARF:ICCV:2021,ANeRF:NeurIPS:2021} propose to model the radiance field in canonical space and use a pre-defined or learned backward mapping to map query points from observation space to this canonical space. A-NeRF~\cite{ANeRF:NeurIPS:2021} uses a deterministic backward mapping defined by piecewise rigid bone transformations. This mapping is very coarse and the model has to use a complicated bone-relative embedding to compensate for that.
Ani-NeRF~\cite{peng2021animatable} trains a backward LBS network that does not generalize well to out-of-distribution poses, even when fine-tuned with a cycle consistency loss for its backward LBS network for each test pose. Further, all aforementioned methods utilize a volumetric radiance representation and hence suffer from noisy geometry~\cite{yariv2020multiview,Oechsle2021ICCV,volsdf:NeurIPS:2021,wang2021neus}. In contrast to these works, we improve geometry by combining an implicit surface representation with volume rendering and improve pose generalization via iterative root-finding. H-NeRF~\cite{HNeRF:NeurIPS:2021} achieves large improvements in geometric reconstruction by co-training SDF and NeRF networks. However, code and models of H-NeRF are not publicly available. Furthermore, H-NeRF's canonicalization process relies on imGHUM~\cite{alldieck2021imghum} to predict an accurate signed distance in \textit{observation space}.
Therefore, imGHUM needs to be trained on a large corpus of posed human scans and it is unclear whether the learned signed distance fields generalize to out-of-distribution poses beyond the training set.
In contrast, our approach does not need to be trained on any posed scans and it can generalize to extreme out-of-distribution poses.

\boldparagraph{Concurrent Works} Several concurrent works extend NeRF-based articulated models to improve novel view synthesis, geometry reconstruction, or animation quality~\cite{hu2021hvtr,xu2022sanerf,chen2021animatable,weng2022humannerf,zheng2022imavatar,jiang2022selfrecon,Mihajlovic:KeypointNeRF:ECCV2022,su2022danbo,li2022tava,peng2022animatable}. ~\cite{zheng2022imavatar} proposes to jointly learn forward blending weights, a canonical occupancy network, and a canonical color network using differentiable surface rendering for head-avatars. In contrast to human heads, human bodies show much more articulation. Abrupt changes in depth also occur more frequently when rendering human bodies, which is difficult to capture with surface rendering~\cite{wang2021neus}. Furthermore,~\cite{zheng2022imavatar} uses the secant method to find surface points. For each secant step, this needs to solve a root-finding problem from scratch. Instead, we use volume rendering of SDFs and formulate the surface-finding task of articulated SDFs as a joint root-finding problem that only needs to be solved once per ray. We remark that~\cite{jiang2022selfrecon} proposes to formulate surface-finding and correspondence search as a joint root-finding problem to tackle geometry reconstruction from photometric and mask losses. However, they use pre-defined skinning fields and surface rendering. They also require estimated normals from PIFuHD~\cite{Saito_CVPR2020} while our approach achieves detailed geometry reconstructions without such supervision.
\section{Method}
\label{sec:fundamentals}
\begin{figure}[t]
\centering
  \includegraphics[width=1.0\textwidth]{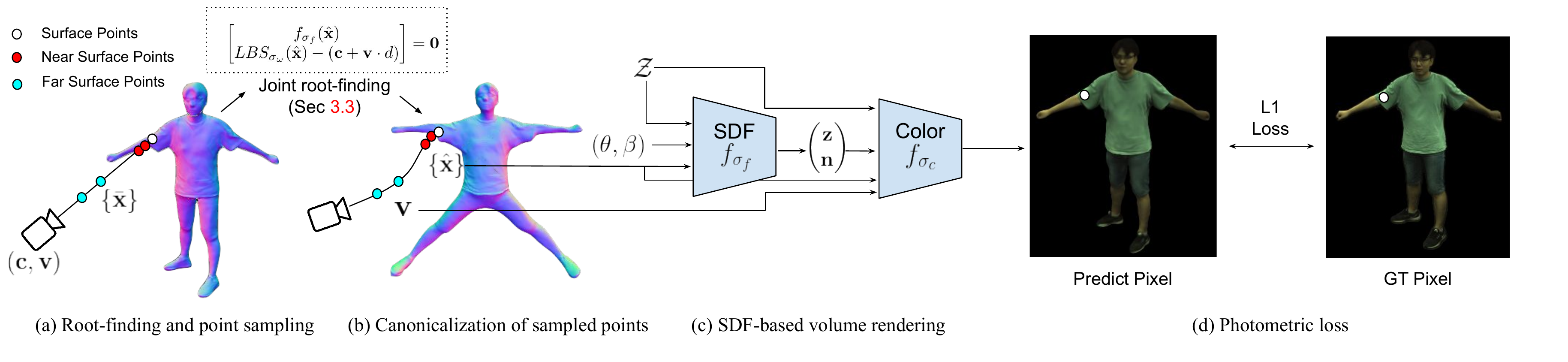}
  \caption{\textbf{Overview of Our Pipeline.} (a) Given a ray $( \mathbf{c}, \mathbf{v} )$ with camera center $\mathbf{c}$ and ray direction $\mathbf{v}$ in observation space, we jointly search for its intersection with the SDF iso-surface and the correspondence of the intersection point via a novel joint root-finding algorithm (Section~\ref{sec:joint_root_finding}). We then sample near/far surface points $\{ \bar{\mathbf{x}} \}$. (b) The sampled points are mapped into canonical space as $\{ \hat{\mathbf{x}} \}$ via root-finding. (c) In canonical space, we run an SDF-based volume rendering with canonicalized points $\{ \hat{\mathbf{x}} \}$, local body poses and shape $(\theta, \beta)$, an SDF network feature $\mathbf{z}$, surface normals $\mathbf{n}$, and a per-frame latent code $\mathcal{Z}$ to predict the corresponding pixel value of the input ray (Section~\ref{sec:vol_sdf}). (d) All network modules, including the forward LBS network $LBS_{\sigma_{\omega}}$, the canonical SDF network $f_{\sigma_f}$, and the canonical color network $f_{\sigma_c}$, are trained end-to-end with a photometric loss in image space and regularization losses in canonical space (Section~\ref{sec:loss}).}
  \label{fig:pipeline}
\end{figure}
Our pipeline is illustrated in Fig.~\ref{fig:pipeline}. Our model consists of a forward linear blend skinning (LBS) network (Section~\ref{sec:LBS}), a canonical SDF network, and a canonical color network (Section~\ref{sec:sdf_and_color}). When rendering a specific pixel of the image in observation space, we first find the intersection of the corresponding camera ray and the observation-space SDF iso-surface. Since we model a canonical SDF and a forward LBS, we propose a novel joint root-finding algorithm that can simultaneously search for the ray-surface intersection and the canonical correspondence of the intersection point (Section~\ref{sec:joint_root_finding}). Such a formulation does not condition the networks on observations in observation space. Consequently, it can generalize to unseen poses.
Once the ray-surface intersection is found, we sample near/far surface points on the camera ray and find their canonical correspondences via forward LBS root-finding. The canonicalized points are used for volume rendering to compose the final RGB value at the pixel (Section~\ref{sec:vol_sdf}). The predicted pixel color is then compared to the observation using a photometric loss (Section~\ref{sec:loss}). The model is trained end-to-end using the photometric loss and regularization losses. The learned networks represent a personalized animatable avatar that can robustly synthesize new geometries and appearances under novel poses (Section~\ref{sec:exp_novel_pose}).

\subsection{Neural Linear Blend Skinning}
\label{sec:LBS}
Traditional parametric human body models~\cite{SCAPE, Hasler2009CGF, SMPL:2015, SMPL-X:2019, STAR:ECCV:2020, Xu_2020_CVPR} often use linear blend skinning (LBS) to deform a template model according to rigid bone transformations and skinning weights. We follow the notations of~\cite{MetaAvatar:NeurIPS:2021} to describe LBS. Given a set of $N$ points in canonical space,  $\hat{\mathbf{X}} = \{ \hat{\mathbf{x}}^{(i)} \}_{i=1}^{N}$, LBS takes a set of rigid bone transformations $\{ \mathbf{B}_b \}_{b=1}^{24}$ as inputs, each $\mathbf{B}_b$ being a $4 \times 4$ rotation-translation matrix. We use 23 local transformations and one global transformation with an underlying SMPL~\cite{SMPL:2015} model. For a 3D point $\hat{\mathbf{x}}^{(i)} \in \hat{\mathbf{X}}$~\footnote{with slight abuse of notation, we also use $\hat{\mathbf{x}}$ to represent points in homogeneous coordinates when necessary.}, a skinning weight vector is defined as $\mathbf{w}^{(i)} \in [ 0, 1 ]^{24}, \text{s.t.} \sum_{b=1}^{24} \mathbf{w}_b^{(i)} = 1$. This vector indicates the affinity of the point $\hat{\mathbf{x}}^{(i)}$ to each of the bone transformations $\{ \mathbf{B}_b \}_{b=1}^{24}$. Following recent works~\cite{LEAP:CVPR:21,SCANimate:CVPR:21,Chen2021ICCV,MetaAvatar:NeurIPS:2021}, we use a neural network $f_{\sigma_{\omega}} (\cdot): \mathbb{R}^3 \mapsto [ 0, 1 ]^{24}$ with parameters $\sigma_{\omega}$ to predict the skinning weights of any point in space. The set of transformed points $\bar{\mathbf{X}} = \{ \bar{\mathbf{x}}^{(i)} \}_{i=1}^{N}$ is related to $\hat{\mathbf{X}}$ via:
\begin{align}
    \label{eqn:LBS}
    & \bar{\mathbf{x}}^{(i)} = LBS_{\sigma_{\omega}}\left(\hat{\mathbf{x}}^{(i)}, \{ \mathbf{B}_b \} \right), \quad \forall i = 1, \ldots, N \nonumber \\
    \Longleftrightarrow & \bar{\mathbf{x}}^{(i)} = \left(\sum_{b=1}^{24} f_{\sigma_{\omega}} (\hat{\mathbf{x}}^{(i)})_{b} \mathbf{B}_b\right) \hat{\mathbf{x}}^{(i)}, \quad \forall i = 1, \ldots, N 
\end{align}
where Eq.~\eqref{eqn:LBS} is referred to as the forward LBS function. The process of applying Eq.~\eqref{eqn:LBS} to all points in $\hat{\mathbf{X}}$ is often referred to as \textit{forward skinning}. For brevity, for the remainder of the paper, we drop $\{ \mathbf{B}_b \}$ from the LBS function and write $LBS_{\sigma_{\omega}} (\hat{\mathbf{x}}^{(i)}, \{ \mathbf{B}_b \})$ as $LBS_{\sigma_{\omega}} (\hat{\mathbf{x}}^{(i)})$.

\subsection{Canonical SDF and Color Networks}
\label{sec:sdf_and_color}
We model an articulated human as a neural SDF $f_{\sigma_f} (\hat{\mathbf{x}}, \mathbf{\theta}, \mathbf{\beta}, \mathcal{Z})$ with parameters $\sigma_f$ in canonical space, where $\hat{\mathbf{x}}$ denotes the canonical query point, $\mathbf{\theta}$ and $\mathbf{\beta}$ denote local poses and body shape of the human which capture pose-dependent cloth deformations, and $\mathcal{Z}$ denotes a per-frame optimizable latent code which compensates for time-dependent dynamic cloth deformations. For brevity, we write this neural SDF as $f_{\sigma_f} (\hat{\mathbf{x}})$ in the remainder of the paper.

Similar to the canonical SDF network, we define a canonical color network with parameters $\sigma_c$ as $f_{\sigma_c} (\hat{\mathbf{x}}, \mathbf{n}, \mathbf{v}, \mathbf{z}, \mathcal{Z}): \mathbb{R}^{9+|\mathbf{z}|+|\mathcal{Z}|} \mapsto \mathbb{R}^3$. Here, $\mathbf{n}$ denotes a normal vector in the observation space. $\mathbf{n}$ is computed by transforming the canonical normal vectors using the rotational part of forward transformations $\sum_{b=1}^{24} f_{\sigma_{\omega}} (\hat{\mathbf{x}}^{(i)})_{b} \mathbf{B}_b$ (Eq.~\eqref{eqn:LBS}). $\mathbf{v}$ denotes viewing direction. Similar to~\cite{yariv2020multiview,wang2021neus,volsdf:NeurIPS:2021}, $\mathbf{z}$ denotes an SDF feature which is extracted from the output of the second-last layer of the neural SDF. $\mathcal{Z}$ denotes a per-frame latent code which is shared with the SDF network. It compensates for time-dependent dynamic lighting effects. The outputs of $f_{\sigma_c}$ are RGB color values in the range $[0, 1]$.

\subsection{Joint Root-Finding}
\label{sec:joint_root_finding}
While surface rendering~\cite{yariv2020multiview,Niemeyer2020CVPR} could be used to learn the network parameters introduced in Sections~\ref{sec:LBS} and~\ref{sec:sdf_and_color}, it cannot handle abrupt changes in depth, as demonstrated in~\cite{wang2021neus}. We also observe severe geometric artifacts when applying surface rendering to our setup, we refer readers to \iftoggle{arXiv}{Appendix~\ref{appx:ablation}}{the Supp. Mat.} for such an ablation. On the other hand, volume rendering can better handle abrupt depth changes in articulated human rendering. However, volume rendering requires multi-step dense sampling on camera rays~\cite{wang2021neus,volsdf:NeurIPS:2021}, which, when combined naively with the iterative root-finding algorithm~\cite{Chen2021ICCV}, requires significantly more memory and becomes prohibitively slow to train and test. We thus employ a hybrid method similar to~\cite{Oechsle2021ICCV}. We first search the ray-surface intersection and then sample near/far surface points on the ray. In practice, we initialize our SDF network with~\cite{MetaAvatar:NeurIPS:2021}. Thus, we fix the sampling depth interval around the surface to $[-5cm, +5cm]$.

A naive way of finding the ray-surface intersection is to use sphere tracing~\cite{Hart:Sphere:1995} and map each point to canonical space via root-finding~\cite{Chen2021ICCV}. In this case, we need to solve the costly root-finding problem during each step of the sphere tracing. This becomes prohibitively expensive when the number of rays is large. Thus, we propose an alternative solution. We leverage the skinning weights of the nearest neighbor on the registered SMPL mesh to the query point $\bar{\mathbf{x}}$ and use the inverse of the linearly combined forward bone transforms to map $\bar{\mathbf{x}}$ to its rough canonical correspondence. Combining this approximate backward mapping with sphere tracing, we obtain rough estimations of intersection points. Then, starting from these rough estimations, we apply a novel joint root-finding algorithm to search the precise intersection points and their correspondences in canonical space. In practice, we found that using a single initialization for our joint root-finding works well already. Adding more initializations incurs drastic memory and runtime overhead while not achieving any noticeable improvements. We hypothesize that this is due to the fact that our initialization is obtained using inverse transformations with SMPL skinning weights rather than rigid bone transformations (as was done in~\cite{Chen2021ICCV}).

Formally, we define a camera ray as $\mathbf{r} = (\mathbf{c}, \mathbf{v})$ where $\mathbf{c}$ is the camera center and $\mathbf{v}$ is a unit vector that defines the direction of this camera ray. Any point on the camera ray can be expressed as $\mathbf{c} + \mathbf{v} \cdot d$ with $d >= 0$. The joint root-finding aims to find canonical point $\hat{\mathbf{x}}$ and depth $d$ on the ray in observation space, such that:
\begin{align}
    f_{\sigma_f} (\hat{\mathbf{x}}) &= 0 \nonumber \\
    LBS_{\sigma_{\omega}}(\hat{\mathbf{x}}) - (\mathbf{c} + \mathbf{v} \cdot d) &= \mathbf{0}
\end{align}
in which $\mathbf{c}, \mathbf{v}$ are constants per ray. Denoting the joint vector-valued function as $g_{\sigma_f, \sigma_{\omega}} (\hat{\mathbf{x}}, d)$ and the joint root-finding problem as:
\begin{align}
\label{eqn:joint_root_finding}
    g_{\sigma_f, \sigma_{\omega}} (\hat{\mathbf{x}}, d) =
    \begin{bmatrix}
    f_{\sigma_f} (\hat{\mathbf{x}}) \\ LBS_{\sigma_{\omega}}(\hat{\mathbf{x}}) - (\mathbf{c} + \mathbf{v} \cdot d)
    \end{bmatrix} = \mathbf{0}
\end{align}
we can then solve it via Newton's method
\begin{align}
    \begin{bmatrix}
    \hat{\mathbf{x}}_{k+1} \\ d_{k+1}
    \end{bmatrix} =
    \begin{bmatrix}
    \hat{\mathbf{x}}_{k} \\ d_{k}
    \end{bmatrix} - \mathbf{J}^{-1}_{k} \cdot g_{\sigma_f, \sigma_{\omega}} (\hat{\mathbf{x}}_{k}, d_{k})
\end{align}
where:
\begin{align}
    \mathbf{J}_{k} =
    \begin{bmatrix}
    \frac{\partial f_{\sigma_f}}{\partial \hat{\mathbf{x}}} (\hat{\mathbf{x}}_{k}) & 0 \\[6pt] \frac{\partial LBS_{\sigma_{\omega}}}{\partial \hat{\mathbf{x}}} (\hat{\mathbf{x}}_{k}) & -\mathbf{v}
    \end{bmatrix}
\end{align}
Following~\cite{Chen2021ICCV}, we use Broyden's method to avoid computing $\mathbf{J}_{k}$ at each iteration.

\boldparagraph{Amortized Complexity} Given the number of sphere-tracing steps as N and the number of root-finding steps as M, the amortized complexity for joint root-finding is $O(M)$ while naive alternation between sphere-tracing and root-finding is $O(MN)$. In practice, this results in about $5 \times$ speed up of joint root-finding compared to the naive alternation between sphere-tracing and root-finding. We also note that from a theoretical perspective, our proposed joint root-finding converges quadratically while the secant-method-based root-finding in the concurrent work~\cite{zheng2022imavatar} converges only superlinearly.

We describe how to compute implicit gradients \wrt the canonical SDF and the forward LBS in \iftoggle{arXiv}{Appendix~\ref{appx:implicit_gradients}.}{the Supp. Mat.} In the main paper, we use volume rendering which does not need to compute implicit gradients \wrt the canonical SDF.

\subsection{Differentiable Volume Rendering}
\label{sec:vol_sdf}
We employ a recently proposed SDF-based volume rendering formulation~\cite{volsdf:NeurIPS:2021}. Specifically, we convert SDF values into density values $\sigma$ using the scaled CDF of the Laplace distribution with the negated SDF values as input
\begin{align}
\label{eqn:sdf_to_density}
    \sigma (\hat{\mathbf{x}}) = \frac{1}{b} \left(\frac{1}{2} + \frac{1}{2} \text{sign}(-f_{\sigma_f} (\hat{\mathbf{x}})\big) \big(1 - \exp(-\frac{|-f_{\sigma_f} (\hat{\mathbf{x}})|}{b})\right)
\end{align}
where $b$ is a learnable parameter. Given the surface point found via solving Eq.~\eqref{eqn:joint_root_finding}, we sample 16 points around the surface points and another 16 points between the near scene bound and the surface point, and map them to canonical space along with the surface point. For rays that do not intersect with any surface, we uniformly sample 64 points for volume rendering. With $N$ sampled points on a ray $\mathbf{r} = (\mathbf{c}, \mathbf{v})$, we use standard volume rendering~\cite{mildenhall2020nerf} to render the pixel color
\begin{align}
\label{eqn:rendering_eqn}
    \hat{C}(\mathbf{r}) &= \sum_{i=1}^{N} T^{(i)} \left(1 - \exp(-\sigma(\hat{\mathbf{x}}^{(i)})\delta^{(i)})\right) f_{c_{\sigma}} (\hat{\mathbf{x}}^{(i)}, \mathbf{n}^{(i)}, \mathbf{v}, \mathbf{z}, \mathcal{Z}) \\
    T^{(i)} &= \exp \left( -\sum_{j < i} \sigma(\hat{\mathbf{x}}^{(j)}) \delta^{(j)} \right)
\end{align}
where $\delta^{(i)} = |d^{(i+1)} - d^{(i)}|$.

\subsection{Loss Function}
\label{sec:loss}
Our loss consists of a photometric loss in observation space and multiple regularizers in canonical space
\begin{align}
\label{eqn:loss_total}
    \mathcal{L} = \lambda_{C} \cdot \mathcal{L}_{C} + \lambda_{E} \cdot \mathcal{L}_{E} + \lambda_{O} \cdot \mathcal{L}_{O} + \lambda_{I} \cdot \mathcal{L}_{I} + \lambda_{S} \cdot \mathcal{L}_{S}
\end{align}
$\mathcal{L}_{C}$ is the L1 loss for color predictions. $\mathcal{L}_{E}$ is the Eikonal regularization~\cite{Gropp:2020:ICML}. $\mathcal{L}_{O}$ is an off-surface point loss, encouraging points far away from the SMPL mesh to have positive SDF values. Similarly, $\mathcal{L}_{I}$ regularizes points inside the canonical SMPL mesh to have negative SDF values. $\mathcal{L}_{S}$ encourages the forward LBS network to predict similar skinning weights to the canonical SMPL mesh. Different from~\cite{yariv2020multiview,HNeRF:NeurIPS:2021,jiang2022selfrecon}, we do not use an explicit silhouette loss. Instead, we utilize foreground masks and set all background pixel values to zero. In practice, this encourages the SDF network to predict positive SDF values for points on rays that do not intersect with foreground masks. For detailed definitions of loss terms and model architectures, please refer to \iftoggle{arXiv}{Appendix~\ref{appx:loss},~\ref{appx:networks}}{the Supp. Mat}.
\section{Experiments}
\label{sec:experiments}
We validate the generalization ability and reconstruction quality of our proposed method against several recent baselines~\cite{peng2020neural,peng2021animatable,ANeRF:NeurIPS:2021}.
As was done in~\cite{peng2020neural}, we consider a setup with 4 cameras positioned equally spaced around the human subject. 
For an ablation study on different design choices of our model, including ray sampling strategy, LBS networks, and number of initializations for root-finding,  we refer readers to \iftoggle{arXiv}{Appendix~\ref{appx:ablation}.}{the Supp. Mat.}

\boldparagraph{Datasets} We use the ZJU-MoCap~\cite{peng2020neural} dataset as our primary testbed because its setup includes 23 cameras which allows us to extract pseudo-ground-truth geometry to evaluate our model. More specifically, the dataset consists of 9 sequences captured with 23 calibrated cameras. We use the training/testing splits from Neural Body~\cite{peng2020neural} for both the cameras and the poses. As one of our goals is learn to detailed geometry, we collect pseudo-ground-truth geometry for the training poses. We use all 23 cameras and apply NeuS with a background NeRF model~\cite{wang2021neus}, a state-of-the-art method for multi-view reconstruction. Note that we refrain from using the masks provided by Neural Body~\cite{peng2020neural} as these masks are noisy and insufficient for accurate static scene reconstruction. We observe that geometry reconstruction with NeuS~\cite{wang2021neus} fails when subjects wear black clothes or the environmental light is not bright enough. Therefore, we manually exclude bad reconstructions and discard sequences with less than 3 valid reconstructions. For completeness, we also tested our approach on the H36M dataset~\cite{h36m_pami} and report a quantitative comparison to~\cite{peng2021animatable,NARF:ICCV:2021} in \iftoggle{arXiv}{Appendix~\ref{appx:additional_quantitative}.}{the Supp. Mat.}

\boldparagraph{Baselines} We compare against three major baselines: Neural Body~\cite{peng2020neural}(NB), Ani-NeRF~\cite{peng2021animatable}(AniN), and A-NeRF~\cite{ANeRF:NeurIPS:2021}(AN). Neural Body diffuses per-SMPL-vertex latent codes into observation space as additional conditioning for NeRF models to achieve state-of-the-art novel view synthesis results on training poses. Ani-NeRF learns a canonical NeRF model and a backward LBS network which predicts residuals to the deterministic SMPL-based backward LBS. Consequently, the LBS network needs to be re-trained for each test sequence. A-NeRF employs a deterministic backward mapping with bone-relative embeddings for query points and only uses keypoints and joint rotations instead of surface models (\ie\ SMPL surface). For the detailed setups of these baselines, please refer to \iftoggle{arXiv}{Appendix~\ref{appx:impl_baselines}.}{the Supp. Mat.}

\boldparagraph{Benchmark Tasks} We benchmark our approach on three tasks: generalization to unseen poses, geometry reconstruction, and novel-view synthesis. 
To analyze generalization ability, we evaluate the trained models on unseen testing poses. 
Due to the stochastic nature of cloth deformations, we quantify performance via perceptual similarity to the ground-truth images with the LPIPS~\cite{zhang2018perceptual} metric. We report PSNR and SSIM in \iftoggle{arXiv}{Appendix~\ref{appx:additional_quantitative}.}{the Supp. Mat.} We also encourage readers to check out qualitative comparison videos at \href{https://neuralbodies.github.io/arah/}{\color{black}{https://neuralbodies.github.io/arah/}}.

For geometry reconstruction, we evaluate our method and baselines on the training poses.
We report point-based L2 Chamfer distance (CD) and normal consistency (NC) wrt. the pseudo-ground-truth geometry. During the evaluation, we only keep the largest connected component of the reconstructed meshes. Note that is in favor of the baselines as they are more prone to producing floating blob artifacts. 
We also remove any ground-truth or predicted mesh points that are below an estimated ground plane to exclude outliers from the ground plane from the evaluation.
For completeness, we also evaluate novel-view synthesis with PSNR, SSIM, and LPIPS using the poses from the training split.
\begin{figure}[t]
\captionsetup[subfigure]{labelformat=empty}
\centering
\begin{subfigure}[b]{0.19\textwidth}
    \includegraphics [trim=160 120 160 70, clip, width=1.0\textwidth]{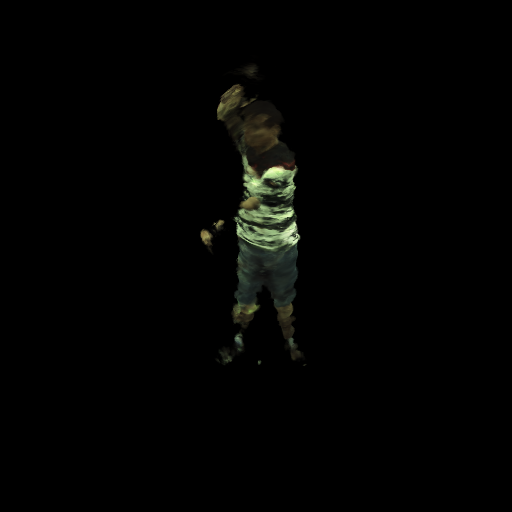}
\end{subfigure}
\begin{subfigure}[b]{0.19\textwidth}
    \includegraphics [trim=160 120 160 70, clip, width=1.0\textwidth]{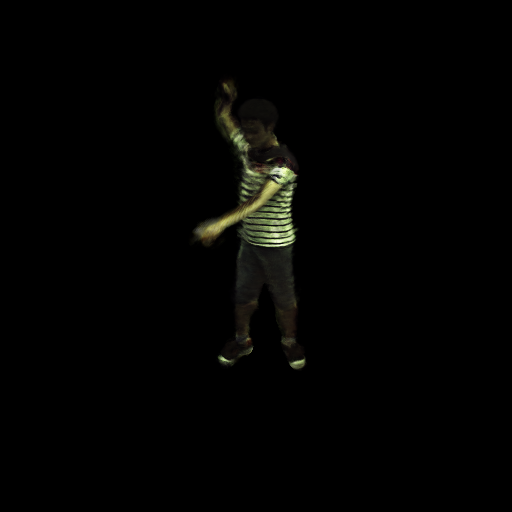}
\end{subfigure}
\begin{subfigure}[b]{0.19\textwidth}
    \includegraphics [trim=160 120 160 70, clip, width=1.0\textwidth]{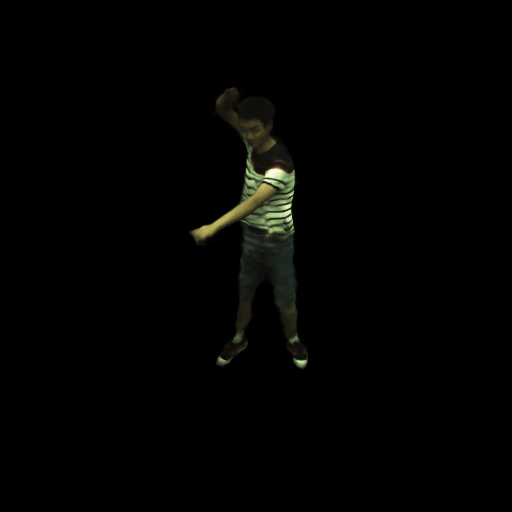}
\end{subfigure}
\begin{subfigure}[b]{0.19\textwidth}
    \includegraphics [trim=160 120 160 70, clip, width=1.0\textwidth]{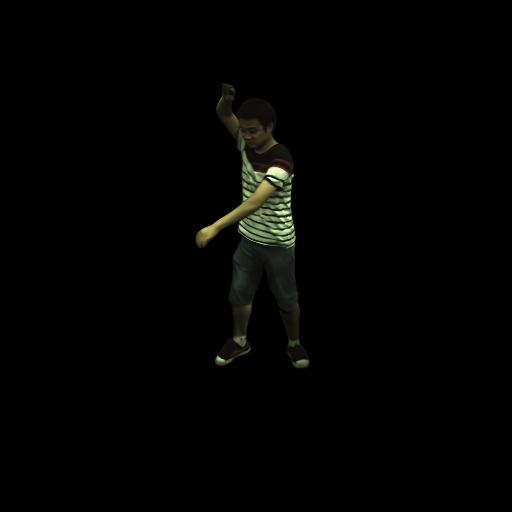}
\end{subfigure}
\begin{subfigure}[b]{0.19\textwidth}
    \includegraphics [trim=160 120 160 70, clip, width=1.0\textwidth]{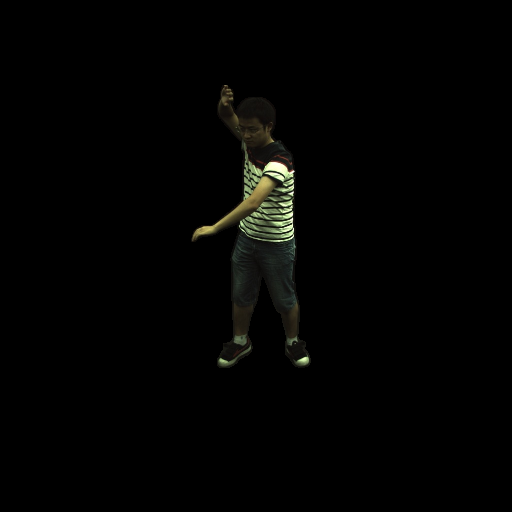}
\end{subfigure} \\
\begin{subfigure}[b]{0.19\textwidth}
    \includegraphics [trim=240 159 110 49, clip, width=1.0\textwidth]{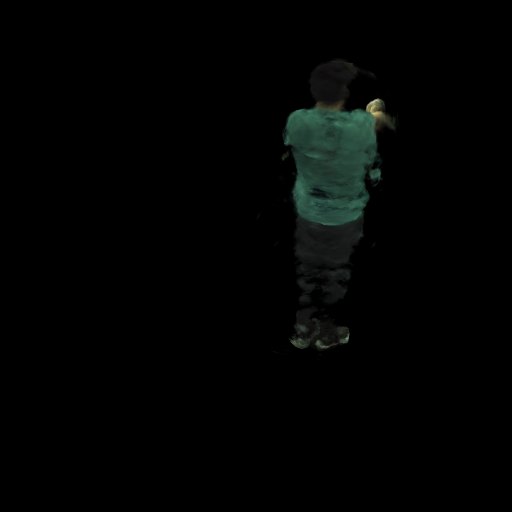}
    \caption{A-NeRF}
\end{subfigure}
\begin{subfigure}[b]{0.19\textwidth}
    \includegraphics [trim=240 159 110 49, clip, width=1.0\textwidth]{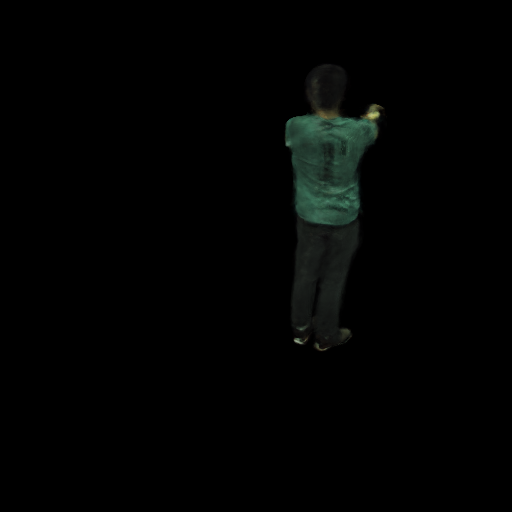}
    \caption{Ani-NeRF}
\end{subfigure}
\begin{subfigure}[b]{0.19\textwidth}
    \includegraphics [trim=240 159 110 49, clip, width=1.0\textwidth]{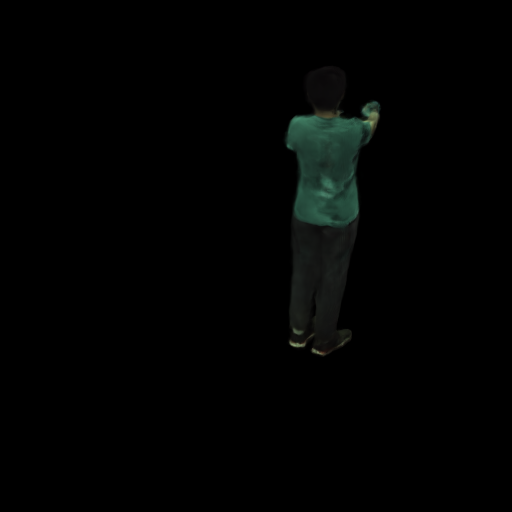}
    \caption{Neural Body}
\end{subfigure}
\begin{subfigure}[b]{0.19\textwidth}
    \includegraphics [trim=240 159 110 49, clip, width=1.0\textwidth]{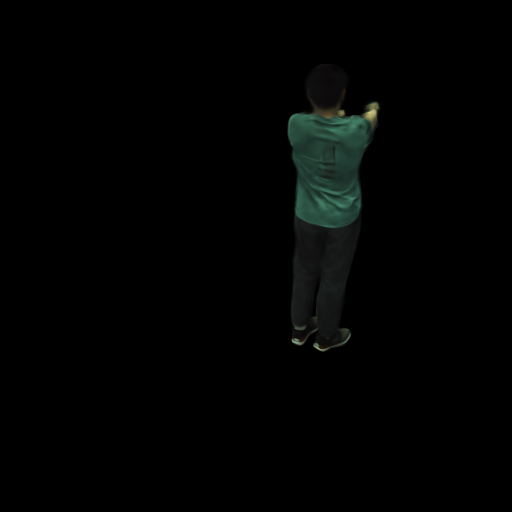}
    \caption{Ours}
\end{subfigure}
\begin{subfigure}[b]{0.19\textwidth}
    \includegraphics [trim=240 159 110 49, clip, width=1.0\textwidth]{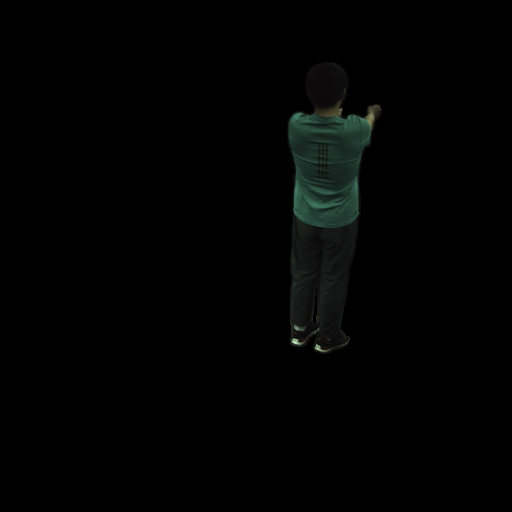}
    \caption{GT}
\end{subfigure}
\caption{\textbf{Generalization to Unseen Poses} on the testing poses of ZJU-MoCap. A-NeRF struggles with unseen poses due to the limited training poses and the lack of a SMPL surface prior. Ani-NeRF produces noisy images as it uses an inaccurate backward mapping function. Neural Body loses details, e.g. wrinkles, because its conditional NeRF is learned in observation space. Our approach generalizes well to unseen poses and can model fine details like wrinkles.}
\label{fig:qualitative_results_zju}
\end{figure}
\begin{figure}[t]
\captionsetup[subfigure]{labelformat=empty}
\scriptsize
\centering
\begin{subfigure}[b]{0.19\textwidth}
    \includegraphics [trim=80 130 180 70, clip, width=1.0\textwidth]{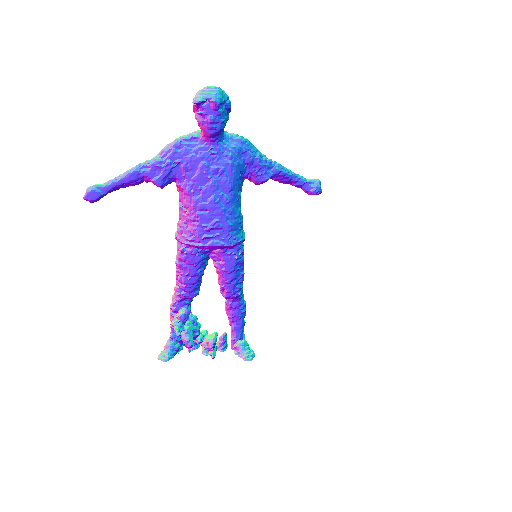}
\end{subfigure}
\begin{subfigure}[b]{0.19\textwidth}
    \includegraphics [trim=80 130 180 70, clip, width=1.0\textwidth]{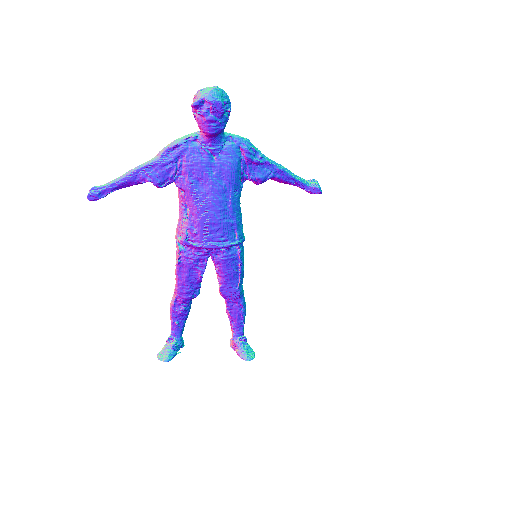}
\end{subfigure}
\begin{subfigure}[b]{0.19\textwidth}
    \includegraphics [trim=80 130 180 70, clip, width=1.0\textwidth]{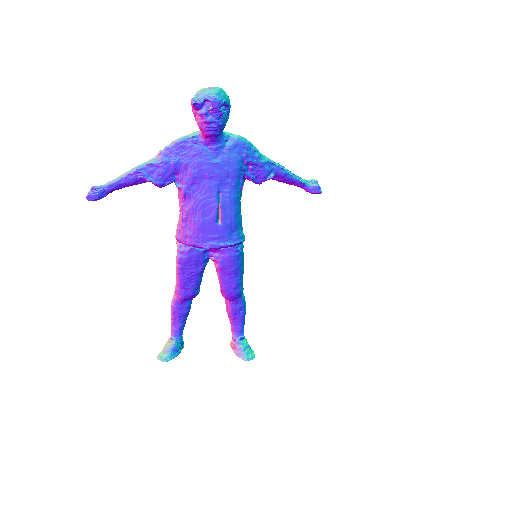}
\end{subfigure}
\begin{subfigure}[b]{0.19\textwidth}
    \includegraphics [trim=80 130 180 70, clip, width=1.0\textwidth]{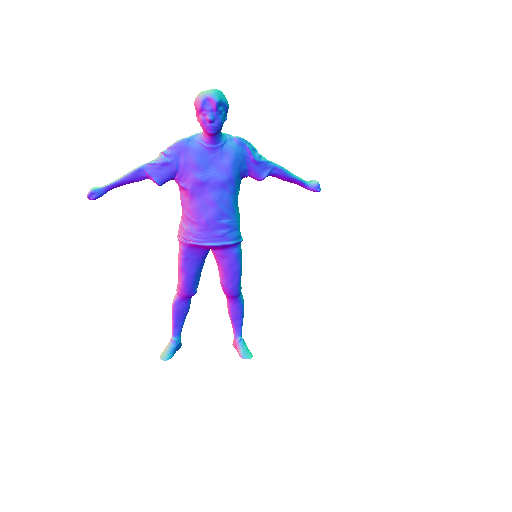}
\end{subfigure}
\begin{subfigure}[b]{0.19\textwidth}
    \includegraphics [trim=80 130 180 70, clip, width=1.0\textwidth]{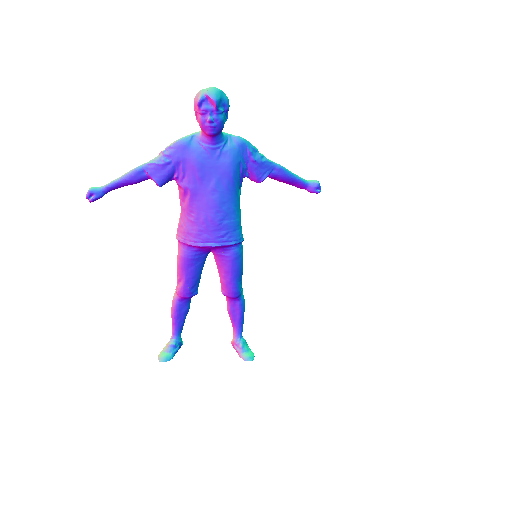}
\end{subfigure} \\
\begin{subfigure}[b]{0.19\textwidth}
    \includegraphics [trim=210 140 50 60, clip, width=1.0\textwidth]{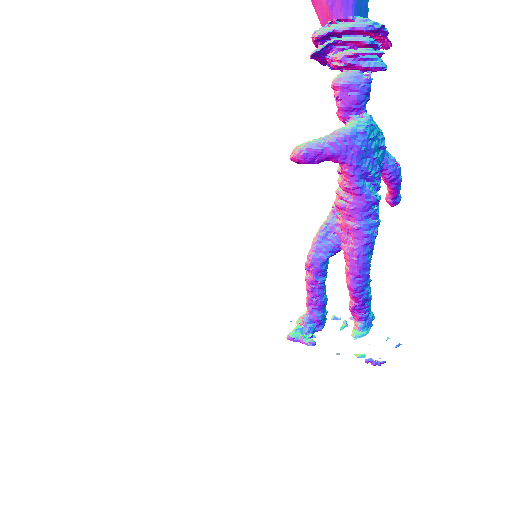}
\end{subfigure}
\begin{subfigure}[b]{0.19\textwidth}
    \includegraphics [trim=210 140 50 60, clip, width=1.0\textwidth]{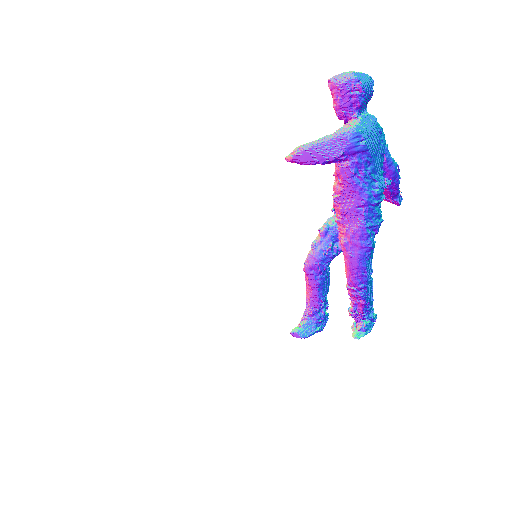}
\end{subfigure}
\begin{subfigure}[b]{0.19\textwidth}
    \includegraphics [trim=210 140 50 60, clip, width=1.0\textwidth]{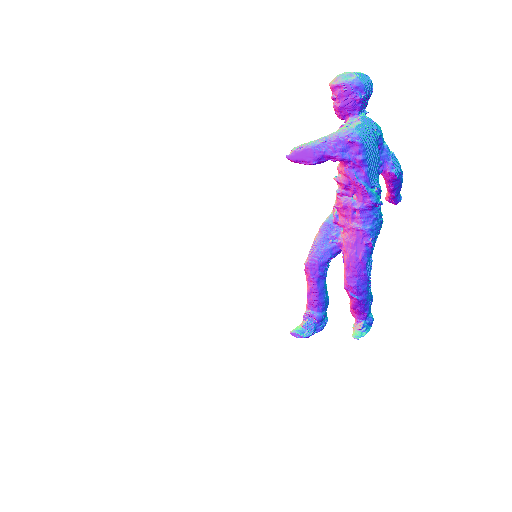}
\end{subfigure}
\begin{subfigure}[b]{0.19\textwidth}
    \includegraphics [trim=210 140 50 60, clip, width=1.0\textwidth]{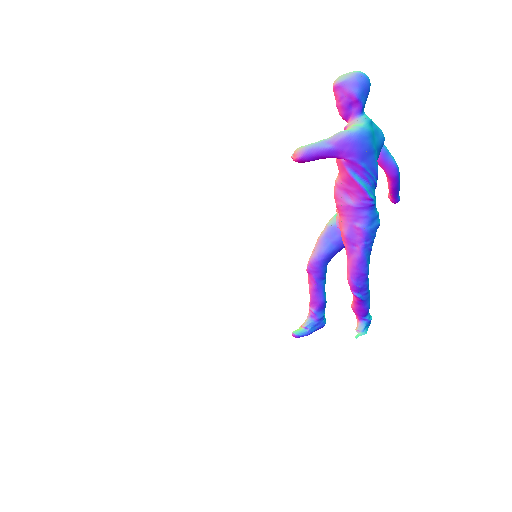}
\end{subfigure}
\begin{subfigure}[b]{0.19\textwidth}
    \includegraphics [trim=210 140 50 60, clip, width=1.0\textwidth]{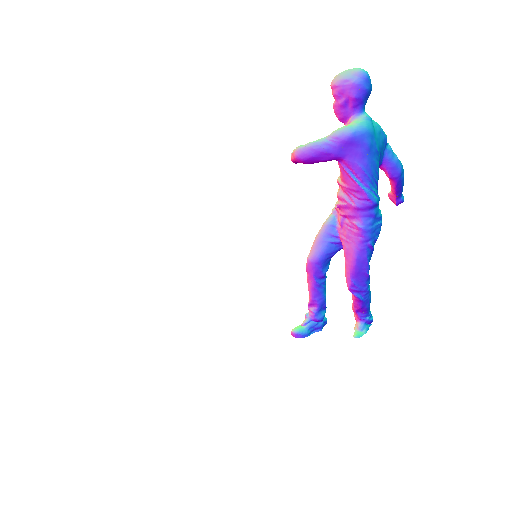}
\end{subfigure} \\
\begin{subfigure}[b]{0.19\textwidth}
    \includegraphics [trim=180 150 70 50, clip, width=1.0\textwidth]{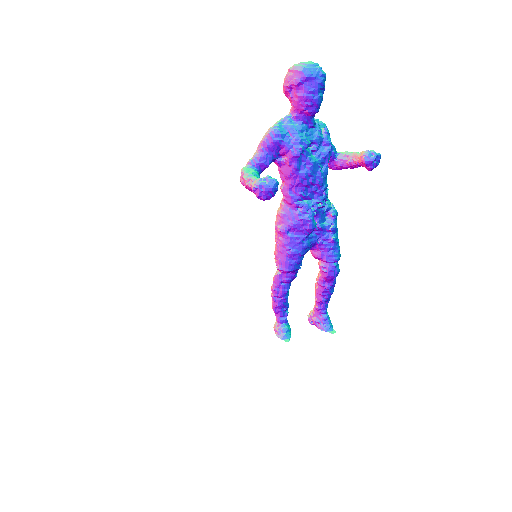}
    \caption{A-NeRF}
\end{subfigure}
\begin{subfigure}[b]{0.19\textwidth}
    \includegraphics [trim=180 150 70 50, clip, width=1.0\textwidth]{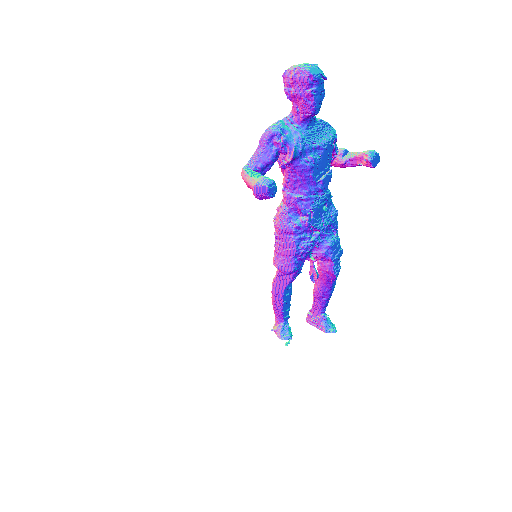}
    \caption{Ani-NeRF}
\end{subfigure}
\begin{subfigure}[b]{0.19\textwidth}
    \includegraphics [trim=180 150 70 50, clip, width=1.0\textwidth]{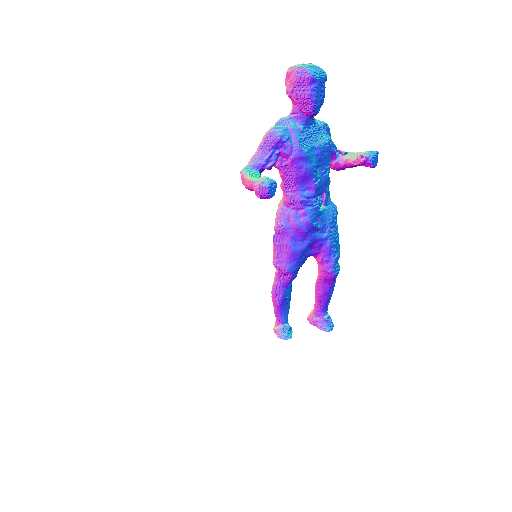}
    \caption{Neural Body}
\end{subfigure}
\begin{subfigure}[b]{0.19\textwidth}
    \includegraphics [trim=180 150 70 50, clip, width=1.0\textwidth]{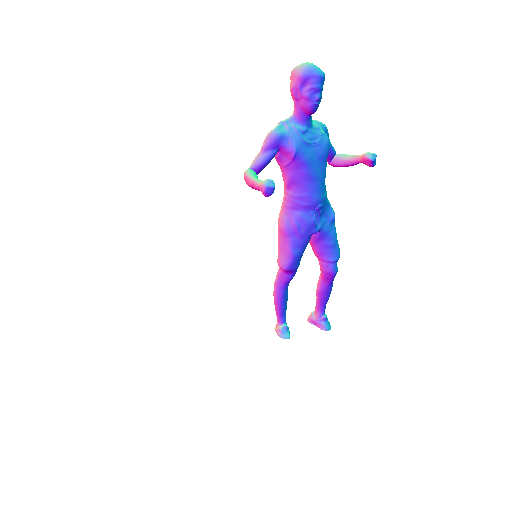}
    \caption{Ours}
\end{subfigure}
\begin{subfigure}[b]{0.19\textwidth}
    \includegraphics [trim=180 150 70 50, clip, width=1.0\textwidth]{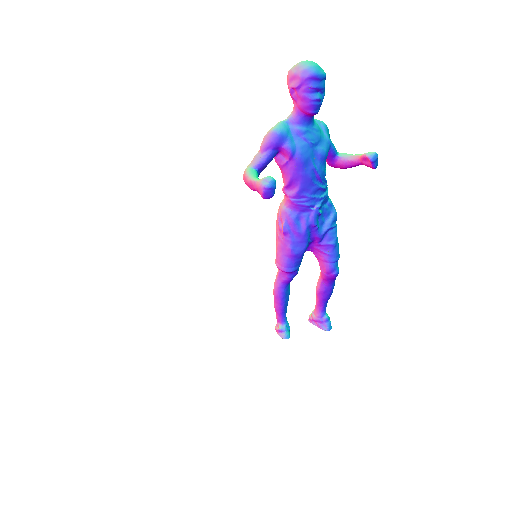}
    \caption{GT}
\end{subfigure}
\caption{\textbf{Geometry Reconstruction}. Our approach reconstructs more fine-grained geometry than the baselines while preserving high-frequency details such as wrinkles. Note that we remove an estimated ground plane from all meshes.}
\label{fig:qualitative_results_geo}
\end{figure}
\begin{table}
\scriptsize
 \renewcommand{\tabcolsep}{2.5pt}
 \begin{minipage}[t]{.48\linewidth}
 \caption{\textbf{Generalization to Unseen Poses}. We report LPIPS~\cite{zhang2018perceptual} on synthesized images under unseen poses from the testset of the ZJU-MoCap dataset~\cite{peng2020neural} (\ie\ all views except 0, 6, 12, and 18). Our approach consistently outperforms the baselines by a large margin. We report PSNR and SSIM \iftoggle{arXiv}{Appendix~\ref{appx:additional_quantitative}.}{in the Supp. Mat.}}
 \iftoggle{arXiv}{\vspace{0.05in}}{}
 \label{tab:novel_pose}
 \centering
 \renewcommand{\arraystretch}{1.41}
 \begin{tabular}{ | c | c | c | c | c | c |}
 \hline
 Sequence & Metric & NB & AniN & AN & Ours \\\hline
 \multirow{1}{*}{313} & LPIPS $\downarrow$ & 0.126 & 0.115 & 0.209 & \textbf{0.092} \\\hline
 \multirow{1}{*}{315} & LPIPS $\downarrow$ & 0.152 & 0.167 & 0.232 & \textbf{0.105} \\\hline
 \multirow{1}{*}{377} & LPIPS $\downarrow$ & 0.119 & 0.153 & 0.165 & \textbf{0.093} \\\hline
 \multirow{1}{*}{386} & LPIPS $\downarrow$ & 0.171 & 0.187 & 0.241 & \textbf{0.127} \\\hline
 \multirow{1}{*}{387} & LPIPS $\downarrow$ & 0.135 & 0.145 & 0.162 & \textbf{0.099} \\\hline
 \multirow{1}{*}{390} & LPIPS $\downarrow$ & 0.163  & 0.173 & 0.226 & \textbf{0.126} \\\hline
 \multirow{1}{*}{392} & LPIPS $\downarrow$ & 0.135 & 0.169 & 0.183 & \textbf{0.106} \\\hline
 \multirow{1}{*}{393} & LPIPS $\downarrow$ & 0.132 & 0.155 & 0.175 & \textbf{0.104} \\\hline
 \multirow{1}{*}{394} & LPIPS $\downarrow$ & 0.150 & 0.171 & 0.199 & \textbf{0.111} \\\hline
 \end{tabular}
 \end{minipage} \quad
 \begin{minipage}[t]{.48\linewidth}
 \caption{\textbf{Geometry Reconstruction}. We report L2 Chamfer Distance (CD) and Normal Consistency (NC) on the training poses of the ZJU-MoCap dataset~\cite{peng2020neural}. Note that AniN and AN occasionally produce large background blobs that are connected to the body resulting in large deviations from the ground truth.}
 \iftoggle{arXiv}{\vspace{0.05in}}{}
 \label{tab:reconstruction}
 \centering
  \renewcommand{\arraystretch}{1.1}
 \begin{tabular}{ | c | c | c | c | c | c |}
 \hline
 Sequence & Metric & NB & AniN & AN & Ours \\\hline
 \multirow{2}{*}{313} & CD $\downarrow$ & 1.258 & 1.242 & 9.174 & \textbf{0.707} \\
 & NC $\uparrow$ & 0.700 & 0.599 & 0.691 & \textbf{0.809} \\\hline
 \multirow{2}{*}{315} & CD $\downarrow$ & 2.167 & 2.860 & 1.524 & \textbf{0.779} \\
 & NC $\uparrow$ & 0.636 & 0.450 & 0.610 & \textbf{0.753} \\\hline
 \multirow{2}{*}{377} & CD $\downarrow$ & 1.062 & 1.649 & 1.008 & \textbf{0.840} \\
 & NC $\uparrow$ & 0.672 & 0.541 & 0.682 & \textbf{0.786} \\\hline
 \multirow{2}{*}{386} & CD $\downarrow$ & 2.938 & 23.53 & 3.632 & \textbf{2.880} \\
 & NC $\uparrow$ & 0.607 & 0.325 & 0.596 & \textbf{0.741} \\\hline
 \multirow{2}{*}{393} & CD $\downarrow$ & 1.753 & 3.252 & 1.696 & \textbf{1.342} \\
 & NC $\uparrow$ & 0.600 & 0.481 & 0.605 & \textbf{0.739} \\\hline
 \multirow{2}{*}{394} & CD $\downarrow$ & 1.510 & 2.813 & 558.8 & \textbf{1.177} \\
 & NC $\uparrow$ & 0.628 & 0.540 & 0.639 & \textbf{0.762} \\\hline
 \end{tabular}
 \end{minipage}
\end{table}
\begin{table}
\scriptsize
 \caption{\textbf{Novel View Synthesis.} We report PSNR, SSIM, and LPIPS~\cite{zhang2018perceptual} for novel views of training poses of the ZJU-MoCap dataset~\cite{peng2020neural}. Due to better geometry, our approach produces more consistent rendering results across novel views than the baselines. We include qualitative comparisons in \iftoggle{arXiv}{Appendix~\ref{appx:additional_qualitative_train}.}{the Supp. Mat.} Note that we crop slightly larger bounding boxes than Neural Body~\cite{peng2020neural} to better capture loose clothes, \eg\ sequence 387 and 390. Therefore, the reported numbers vary slightly from their evaluation.}
 \iftoggle{arXiv}{\vspace{0.05in}}{}
 \label{tab:novel_view}
 \centering
 \setlength{\tabcolsep}{3pt}
 \renewcommand{\arraystretch}{1.1}
 \begin{tabular}{ | c | c | c | c |  c | c | c  | c | c | c | }
 \hline
 & \multicolumn{3}{c|}{313} & \multicolumn{3}{c|}{315} & \multicolumn{3}{c|}{377}
 \\\hline
  Method & PSNR $\uparrow$ & SSIM $\uparrow$ & LPIPS $\downarrow$ & PSNR $\uparrow$ & SSIM $\uparrow$ & LPIPS $\downarrow$ & PSNR $\uparrow$ & SSIM $\uparrow$ & LPIPS $\downarrow$ \\
 \hline
 NB & 30.5 & 0.967 & 0.068 & 26.4 & 0.958 & 0.079 & \textbf{28.1} & \textbf{0.956} & 0.080 \\
 Ani-N & 29.8 & 0.963 & 0.075 & 23.1 & 0.917 & 0.138 & 24.2 & 0.925 & 0.124 \\
 A-NeRF & 29.2 & 0.954 & 0.075 & 25.1 & 0.948 & 0.087 & 27.2 & 0.951 & 0.080 \\
 Ours & \textbf{31.6} & \textbf{0.973} & \textbf{0.050} & \textbf{27.0} & \textbf{0.965} & \textbf{0.058} & 27.8 & \textbf{0.956} & \textbf{0.071} \\
 \hline
 & \multicolumn{3}{c|}{386} & \multicolumn{3}{c|}{387} & \multicolumn{3}{c|}{390}
 \\\hline
  Method & PSNR $\uparrow$ & SSIM $\uparrow$ & LPIPS $\downarrow$ & PSNR $\uparrow$ & SSIM $\uparrow$ & LPIPS $\downarrow$ & PSNR $\uparrow$ & SSIM $\uparrow$ & LPIPS $\downarrow$ \\
 \hline
 NB & 29.0 & \textbf{0.935} & 0.112 & 26.7 & 0.942 & 0.101 & \textbf{27.9} & 0.928 & 0.112 \\
 Ani-N & 25.6 & 0.878 & 0.199 & 25.4 & 0.926 & 0.131 & 26.0 & 0.912 & 0.148 \\
 A-NeRF & 28.5 & 0.928 & 0.127 & 26.3 & 0.937 & 0.100 & 27.0 & 0.914 & 0.126 \\
 Ours & \textbf{29.2} & 0.934 & \textbf{0.105}  & \textbf{27.0} & \textbf{0.945} & \textbf{0.079} & \textbf{27.9} & \textbf{0.929} & \textbf{0.102} \\
 \hline
 & \multicolumn{3}{c|}{392} & \multicolumn{3}{c|}{393} & \multicolumn{3}{c|}{394}
 \\\hline
  Method & PSNR $\uparrow$ & SSIM $\uparrow$ & LPIPS $\downarrow$ & PSNR $\uparrow$ & SSIM $\uparrow$ & LPIPS $\downarrow$ & PSNR $\uparrow$ & SSIM $\uparrow$ & LPIPS $\downarrow$ \\
 \hline
 NB & \textbf{29.7} & \textbf{0.949} & 0.101 & \textbf{27.7} & 0.939 & 0.105 & 28.7 & 0.942 & 0.098 \\
 Ani-N & 28.0 & 0.931 & 0.151 & 26.1 & 0.916 & 0.151 & 27.5 & 0.924 & 0.142 \\
 A-NeRF & 28.7 & 0.942 & 0.106 & 26.8 & 0.931 & 0.113 & 28.1 & 0.936 & 0.103 \\
 Ours & 29.5 & 0.948 & \textbf{0.090} & \textbf{27.7} & \textbf{0.940} & \textbf{0.093} & \textbf{28.9} & \textbf{0.945} & \textbf{0.084} \\
 \hline
 \end{tabular}
\end{table}

\subsection{Generalization to Unseen Poses}
\label{sec:exp_novel_pose}
We first analyze the generalization ability of our approach in comparison to the baselines. Given a trained model and a pose from the test set, we render images of the human subject in the given pose. We show qualitative results in Fig.~\ref{fig:qualitative_results_zju} and quantitative results in Table~\ref{tab:novel_pose}. 
We significantly outperform the baselines both qualitatively and quantitatively. 
The training poses of the ZJU-MoCap dataset are extremely limited, usually comprising just 60-300 frames of repetitive motion. This limited training data results in severe overfitting for the baselines. In contrast, our method generalizes well to unseen poses, even when training data is limited.

\begin{figure}[t]
\captionsetup[subfigure]{labelformat=empty}
\centering
\begin{subfigure}[b]{0.16\textwidth}
    \includegraphics [trim=220 40 80 80, clip, height=3.6cm]{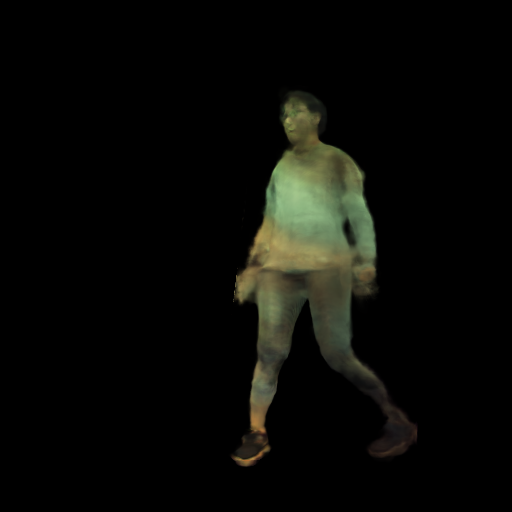}
\end{subfigure}
\begin{subfigure}[b]{0.24\textwidth}
    \includegraphics [trim=100 50 80 60, clip, height=3.6cm]{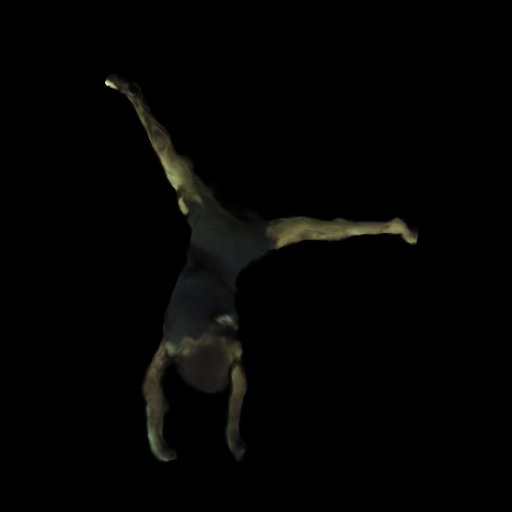}
\end{subfigure}
\begin{subfigure}[b]{0.31\textwidth}
    \includegraphics [trim=60 80 120 120, clip, height=3.6cm]{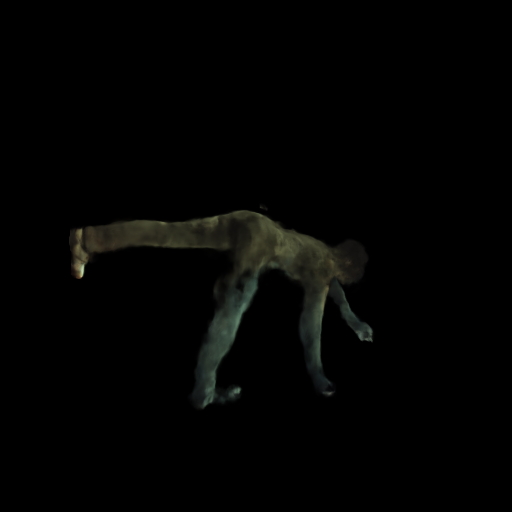}
\end{subfigure}
\begin{subfigure}[b]{0.23\textwidth}
    \includegraphics [trim=160 120 160 40, clip, height=3.6cm]{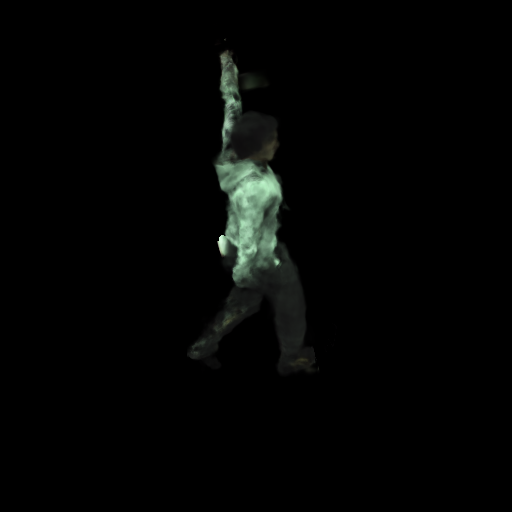}
\end{subfigure} \\
\begin{subfigure}[b]{0.16\textwidth}
    \includegraphics [trim=220 40 80 80, clip, height=3.6cm]{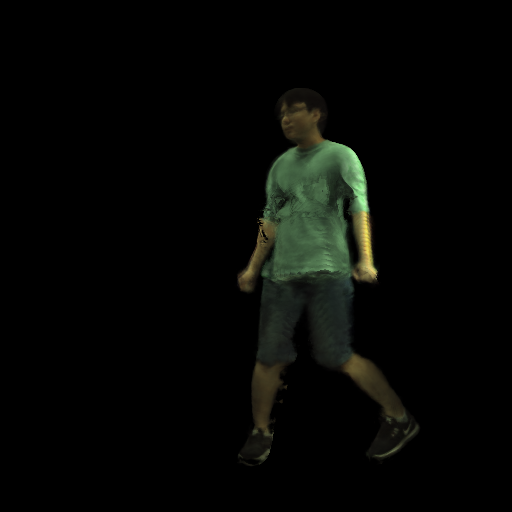}
\end{subfigure}
\begin{subfigure}[b]{0.24\textwidth}
    \includegraphics [trim=100 50 80 60, clip, height=3.6cm]{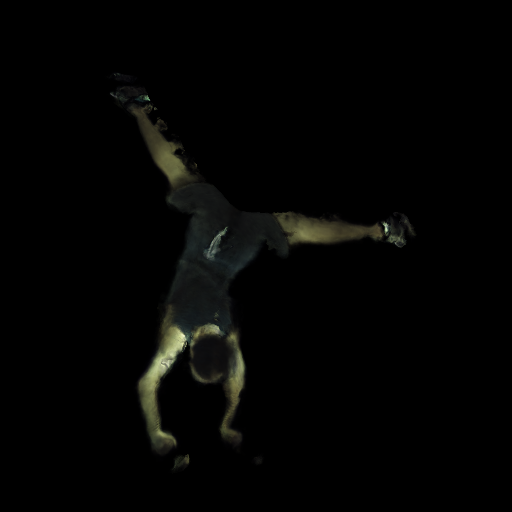}
\end{subfigure}
\begin{subfigure}[b]{0.31\textwidth}
    \includegraphics [trim=60 80 120 120, clip, height=3.6cm]{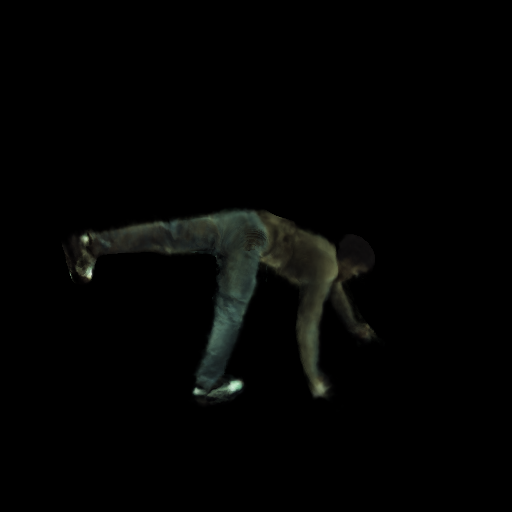}
\end{subfigure}
\begin{subfigure}[b]{0.23\textwidth}
    \includegraphics [trim=160 120 160 40, clip, height=3.6cm]{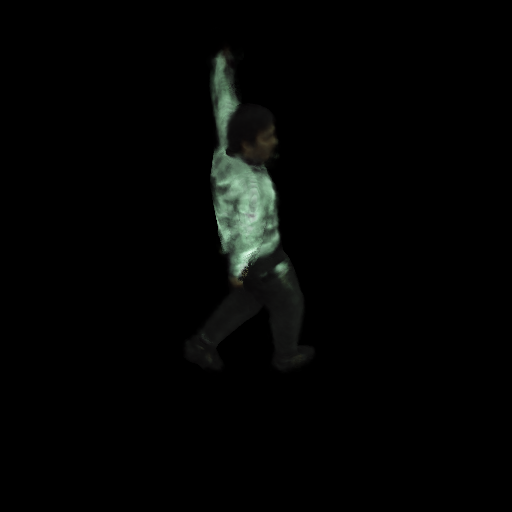}
\end{subfigure} \\
\begin{subfigure}[b]{0.16\textwidth}
    \includegraphics [trim=220 40 80 80, clip, height=3.6cm]{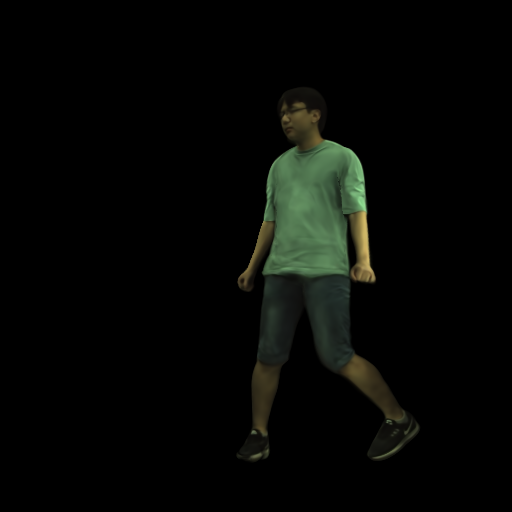}
\end{subfigure}
\begin{subfigure}[b]{0.24\textwidth}
    \includegraphics [trim=100 50 80 60, clip, height=3.6cm]{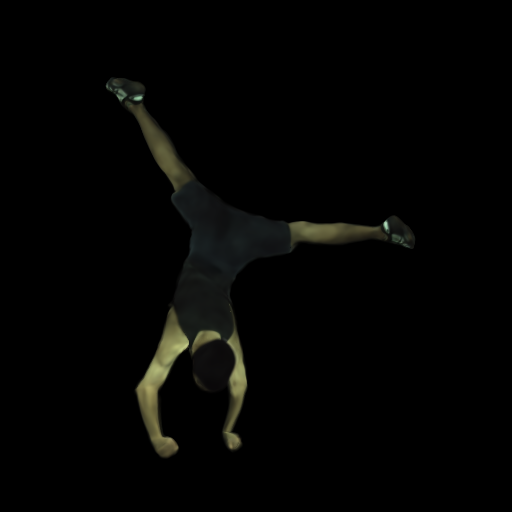}
\end{subfigure}
\begin{subfigure}[b]{0.31\textwidth}
    \includegraphics [trim=60 80 120 120, clip, height=3.6cm]{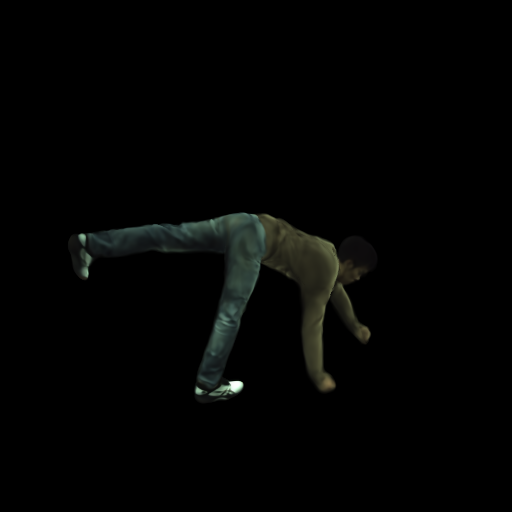}
\end{subfigure}
\begin{subfigure}[b]{0.23\textwidth}
    \includegraphics [trim=160 120 160 40, clip, height=3.6cm]{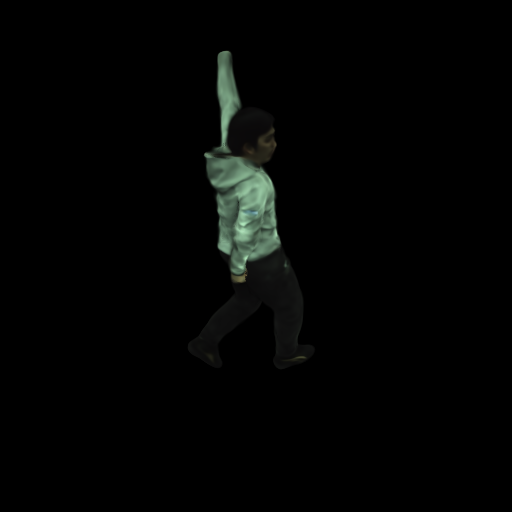}
\end{subfigure} \\
\begin{subfigure}[b]{0.16\textwidth}
    \includegraphics [trim=220 40 80 80, clip, height=3.6cm]{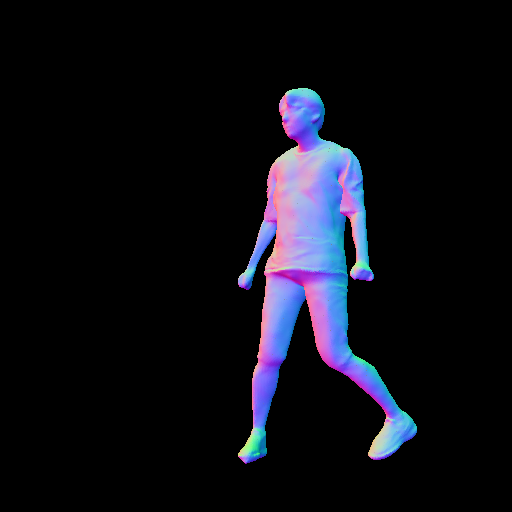}
\end{subfigure}
\begin{subfigure}[b]{0.24\textwidth}
    \includegraphics [trim=100 50 80 60, clip, height=3.6cm]{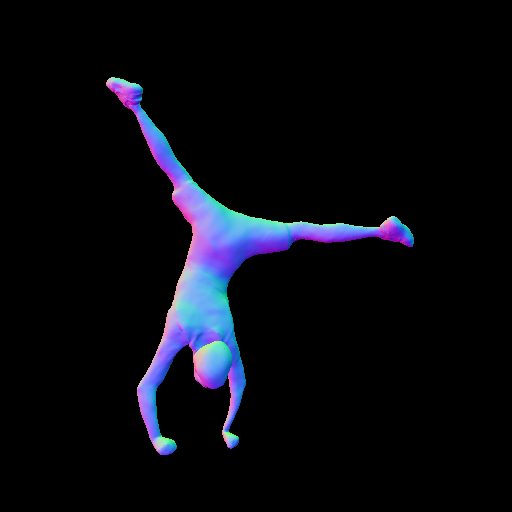}
\end{subfigure}
\begin{subfigure}[b]{0.31\textwidth}
    \includegraphics [trim=60 80 120 120, clip, height=3.6cm]{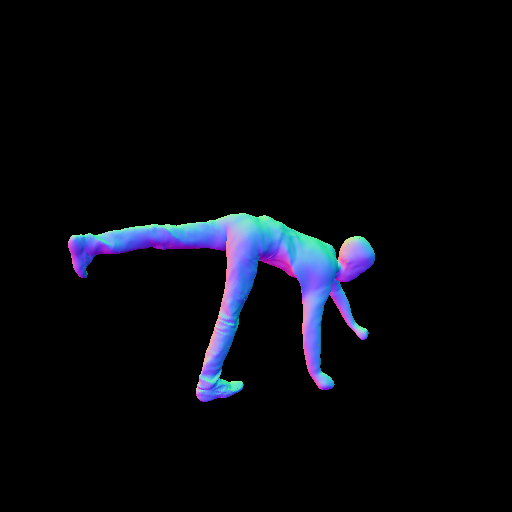}
\end{subfigure}
\begin{subfigure}[b]{0.23\textwidth}
    \includegraphics [trim=160 120 160 40, clip, height=3.6cm]{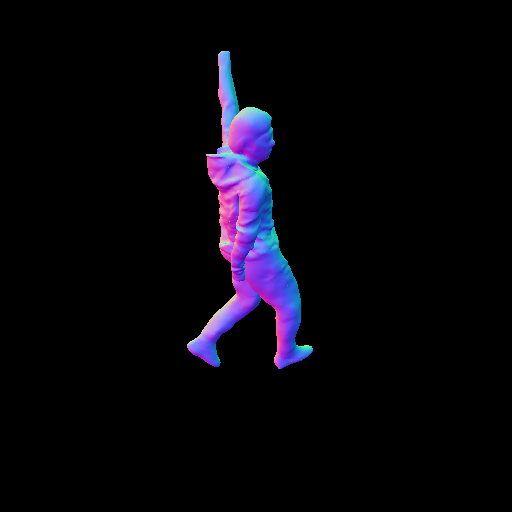}
\end{subfigure}
\caption{\textbf{Qualitative Results on Out-of-distribution Poses} from the AMASS~\cite{AMASS:ICCV:2019} and AIST++~\cite{aist++:ICCV:2021} datasets. From top to bottom row: Neural Body, Ani-NeRF, our rendering, and our geometry. Note that Ani-NeRF requires re-training their backward LBS network on novel pose sequence. We did not show A-NeRF results as it already produces severe overfitting effects on ZJU-MoCap test poses. For more qualitative comparisons, please refer to \iftoggle{arXiv}{Appendix~\ref{appx:additional_qualitative_odp}.}{the Supp. Mat.}}
\label{fig:qualitative_results_odp}
\end{figure}

We additionally animate our models trained on the ZJU-MoCap dataset using extreme out-of-distribution poses from the AMASS~\cite{AMASS:ICCV:2019} and AIST++~\cite{aist++:ICCV:2021} datasets. As shown in Fig.~\ref{fig:qualitative_results_odp}, even under extreme pose variation our approach produces plausible geometry and rendering results while all baselines show severe artifacts. We attribute the large improvement on unseen poses to our root-finding-based backward skinning, as the learned forward skinning weights are constants per subject, while root-finding is a deterministic optimization process that does not rely on learned neural networks that condition on inputs from the observation space. More comparisons can be found in \iftoggle{arXiv}{Appendix~\ref{appx:additional_qualitative_test},~\ref{appx:additional_qualitative_odp}.}{the Supp. Mat.}

\subsection{Geometry Reconstruction on Training Poses}
\label{sec:exp_geo}
Next, we analyze the geometry reconstructed with our approach against reconstructions from the baselines. We compare to the pseudo-ground-truth obtained from NeuS~\cite{wang2021neus}. 
We show qualitative results in Fig.~\ref{fig:qualitative_results_geo} and quantitative results in Table~\ref{tab:reconstruction}. 
Our approach consistently outperforms existing NeRF-based human models on geometry reconstruction. 
As evidenced in Fig.~\ref{fig:qualitative_results_geo}, the geometry obtained with our approach is much cleaner compared to NeRF-based baselines, while preserving high-frequency details such as wrinkles.

\subsection{Novel View Synthesis on Training Poses}
\label{sec:exp_novel_view}
Lastly, we analyze our approach for novel view synthesis on training poses. Table.~\ref{tab:novel_view} provides a quantitative comparison to the baselines. 
While not the main focus of this work, our approach also outperforms existing methods on novel view synthesis. This suggests that more faithful modeling of geometry is also beneficial for the visual fidelity of novel views. Particularly when few training views are available, NeRF-based methods produce blob/cloud artifacts. By removing such artifacts, our approach achieves high image fidelity and better consistency across novel views.
Due to space limitations, we include further qualitative results on novel view synthesis in \iftoggle{arXiv}{Appendix~\ref{appx:additional_qualitative_train}.}{the Supp. Mat.}
\section{Conclusion}
\label{sec:conclusion}
We propose a new approach to create animatable avatars from sparse multi-view videos. We largely improve geometry reconstruction over existing approaches by modeling the geometry as articulated SDFs. 
Further, our novel joint root-finding algorithm enables generalization to extreme out-of-distribution poses. We discuss limitations of our approach in \iftoggle{arXiv}{Appendix~\ref{appx:limitations}.}{the Supp. Mat.}

\boldparagraph{Acknowledgement} Shaofei Wang and Siyu Tang acknowledge the SNF grant 200021 204840. Katja Schwarz was supported by the BMWi in the project KI Delta Learning (project number 19A19013O). Andreas Geiger was supported by the ERC Starting Grant LEGO-3D (850533) and the DFG EXC number 2064/1 - project number 390727645.

\clearpage
%
%
\bibliographystyle{splncs04}
\bibliography{egbib}

\appendix
\numberwithin{equation}{section}
\setcounter{equation}{0}
\numberwithin{figure}{section}
\setcounter{figure}{0}
\numberwithin{table}{section}
\setcounter{table}{0}
\numberwithin{algorithm}{section}
\setcounter{algorithm}{0}

\section{Loss Definition}
\label{appx:loss}
In Section\iftoggle{arXiv}{~\ref{sec:loss}}{~\textcolor{red}{3.5}} of the main paper, we define the loss terms as follows
\begin{align}
\label{appx:eqn:loss_total}
    \mathcal{L} = \lambda_{C} \cdot \mathcal{L}_{C} + \lambda_{E} \cdot \mathcal{L}_{E} + \lambda_{O} \cdot \mathcal{L}_{O} + \lambda_{I} \cdot \mathcal{L}_{I} + \lambda_{S} \cdot \mathcal{L}_{S}
\end{align}
In this section, we elaborate on how each loss term is defined. Let $I_p \in [0, 1]^3$ denote the ground-truth RGB value of a pixel $p$. Further, let $P$ denote the set of all pixels sampled from an image.

\boldparagraph{RGB Color Loss} The RGB color loss is defined as
\begin{align}
\label{appx:eqn:rgb_loss}
\mathcal{L}_{C} = \frac{1}{|P|} \sum_{p \in P} \Big| f_{\sigma_c} (\hat{\mathbf{x}}^{(p)}, \mathbf{n}^{(p)}, \mathbf{v}^{(p)}, \mathbf{z}, \mathcal{Z}) - I_p \Big|
\end{align}
\boldparagraph{Eikonal Regularization} We sample 1024 points, denoted as $ \hat{\mathbf{X}}_{\text{eik}} $, in the range $[ -1, 1 ]^3$ in canonical space, and compute Eikonal loss~\cite{Gropp:2020:ICML} as follows:
\begin{align}
\label{appx:eqn:eikonal_loss}
    \mathcal{L}_{E} = \frac{1}{|P|} \sum_{\hat{\mathbf{x}} \in \hat{\mathbf{X}}_{\text{eik}}} \Big| \lVert \nabla_{\hat{\mathbf{x}}} f_{\sigma_f}(\hat{\mathbf{x}}) \rVert_2 - 1 \Big|
\end{align}
\boldparagraph{Off-surface Point Loss} In canonical space, we sample 1024 points whose distance to the canonical SMPL mesh is greater than 20cm. Let $ \hat{\mathbf{X}}_{\text{off}} $ denote these sampled points, we compute the off-surface point loss as
\begin{align}
\label{appx:eqn:off_surface_loss}
    \mathcal{L}_{O} = \frac{1}{|P|} \sum_{\hat{\mathbf{x}} \in \hat{\mathbf{X}}_{\text{off}}} \exp \left(-1e^2 \cdot f_{\sigma_f}(\hat{\mathbf{x}}) \right) 
\end{align}
\boldparagraph{Inside Point Loss} In canonical space, we sample 1024 points that are inside the canonical SMPL mesh and whose distance to the SMPL surface is greater than 1cm. Let $ \hat{\mathbf{X}}_{\text{in}} $ denote these sampled points, we compute the inside point loss as
\begin{align}
\label{appx:eqn:inside_loss}
    \mathcal{L}_{I} = \frac{1}{|P|} \sum_{\hat{\mathbf{x}} \in \hat{\mathbf{X}}_{\text{in}}} \text{sigmoid} \left( 5e^3 \cdot f_{\sigma_f}(\hat{\mathbf{x}}) \right) 
\end{align}
\boldparagraph{Skinning Loss} Finally, in canonical space, we sample 1024 points on the canonical SMPL surface, $ \hat{\mathbf{X}}_{\text{S}} $, and regularize the forward LBS network with the corresponding SMPL skinning weights $\mathbf{W} = \{ \mathbf{w} \}$:
\begin{align}
\label{appx:eqn:skin_loss}
    \mathcal{L}_{S} = \frac{1}{|P|} \sum_{\substack{\hat{\mathbf{x}} \in \hat{\mathbf{X}}_{\text{S}} \\ \mathbf{w} \in \mathbf{W} }} \sum_{i=1}^{i=24} \Big| f_{\sigma_{\omega}} (\hat{\mathbf{x}})_i - \mathbf{w}_i  \Big|
\end{align}
We set $\lambda_C = 3e^1, \lambda_E = 5e^1, \lambda_O = 1e^2, \lambda_{I} = \lambda_{S} = 10$ throughout all experiments.

\boldparagraph{Mask Loss} As described in Section\iftoggle{arXiv}{~\ref{sec:loss}}{~\textcolor{red}{3.5}} of the main paper, our volume rendering formulation does not need explicit mask loss. Here we describe the mask loss from~\cite{yariv2020multiview} which we use in the ablation study on surface rendering (Section~\ref{appx:ablation}). Given the camera ray $\mathbf{r}^{(p)} = (\mathbf{c}, \mathbf{v}^{(p)})$ of a specific pixel $p$, we first define $S (\alpha, \mathbf{c}, \mathbf{v}^{(p)}) = \text{sigmoid} (-\alpha \min_{d \ge 0} f_{\sigma_f} (LBS_{\sigma_{\omega}}^{-1} (\mathbf{c} + d \mathbf{v}^{(p)}) )$, \ie the Sigmoid of the minimal SDF along a ray. In practice we sample 100 $d$s uniformly between $[d_{\text{min}}, d_{\text{max}}]$ along the ray, where $d_{\text{min}}$ and $d_{\text{max}}$ are determined by the bounding box of the registered SMPL mesh. $\alpha$ is a learnable scalar parameter.

Let $O_p \in \{ 0, 1 \}$ denote the foreground mask value (0 indicates background and 1 indicates foreground) of a pixel $p$. Further, let $P_{in}$ denote the set of pixels for which ray-intersection with the iso-surface of neural SDF is found and $O_p = 1$, while $P_{out} = P \setminus P_{in}$ is the set of pixels for which no ray-intersection with the iso-surface of neural SDF is found or $O_p = 0$. The mask loss is defined as
\begin{align}
\label{eqn:mask_loss}
    \mathcal{L}_{M} = \frac{1}{\alpha|P|} \sum_{p \in P_{out}} \text{BCE} (O_p,  S(\alpha, \mathbf{c}, \mathbf{v}^{(p)})))
\end{align}
where $\text{BCE} (\cdot)$ denotes binary cross entropy loss. We set the weight of $\mathcal{L}_M$ to be $3e^3$ and add this loss term to Eq.~\eqref{appx:eqn:loss_total} for our surface rendering baseline in Section~\ref{appx:ablation}. 
\section{Network Architectures}
\label{appx:networks}
In this section, we describe detailed network architectures for the forward LBS network $f_{\sigma_{\omega}}$, the SDF network $f_{\sigma_f}$ and the color network $f_{\sigma_{c}}$ introduced in Sections\iftoggle{arXiv}{~\ref{sec:LBS}-\ref{sec:sdf_and_color}}{~\textcolor{red}{3.1}-\textcolor{red}{3.2}} of the main paper.

\subsection{Forward LBS Network}
\label{appx:forward_lbs_arch}
We use the same forward LBS network as~\cite{Chen2021ICCV}, which consists of 4 hidden layers with 128 channels and weight normalization~\cite{weightnorm:NeurIPS:16}. It uses Softplus activation with $\beta=100$. $f_{\sigma_{\omega}}$ only takes query points in canonical space as inputs and does not have any conditional inputs.

To initialize this forward LBS network, we meta learn the network on skinning weights of canonical meshes from the CAPE~\cite{CAPE:CVPR:20} dataset. Specifically, we use Reptile~\cite{Reptile:arXiv:2018} with 24 inner steps. The inner learning rate is set to $1e^{-4}$ while the outer learning rate is set to $1e^{-5}$. Adam~\cite{Adam:ICLR:2015} optimizer is used for both the inner and the outer loop. We train with a batch size of 4 for 100k steps of the outer loop. We use the resulting model as the initialization for our per-subject optimization on the ZJU-MoCap~\cite{peng2020neural} dataset.

\subsection{Canonical SDF Network}
\label{appx:sdf_arch}
\begin{figure}[t]
\centering
  \includegraphics[width=1.0\textwidth]{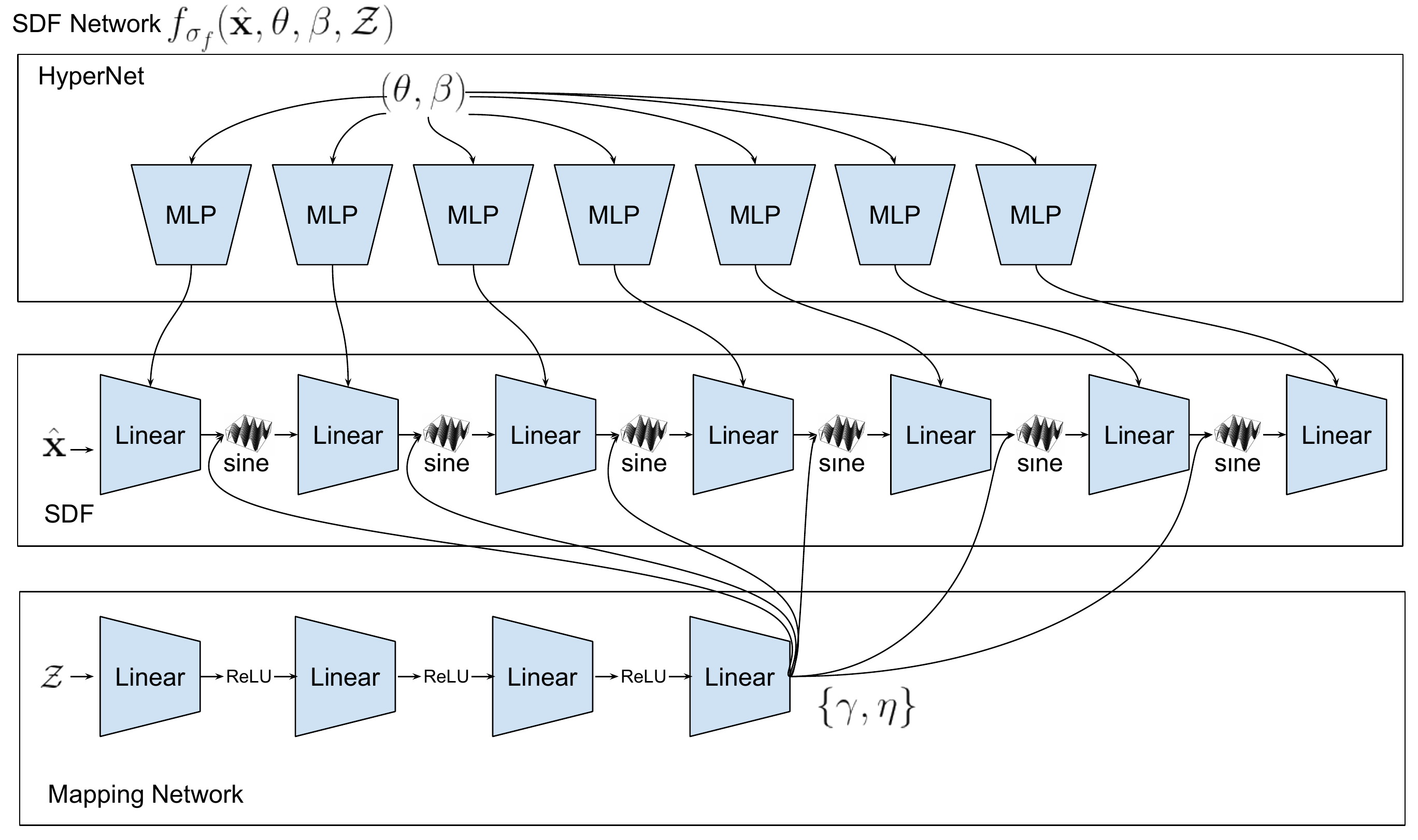}
  \caption{\textbf{Network Architecture for the SDF Network.} Our SDF network builds upon MetaAvatar~\cite{MetaAvatar:NeurIPS:2021} which uses a hypernetwork (top) that conditions on local body poses and shape $( \theta, \beta)$, and predicts the parameters of a neural SDF with periodic activation (middle). Since MetaAvatar does not model per-frame latent codes, we add a mapping network (bottom) that maps the per-frame latent code $\mathcal{Z}$ to scaling factors $\{ \gamma \}$ and offsets $\{ \eta \}$ which are used to modulate the outputs from each linear layer of the neural SDF, except for the last layer.}
  \label{appx:fig:sdf_net_arch}
\end{figure}
We describe our canonical SDF network in Fig.~\ref{appx:fig:sdf_net_arch}. The hypernetwork (top) and neural SDF (middle) are initialized with MetaAvatar~\cite{MetaAvatar:NeurIPS:2021} pre-trained on the CAPE dataset. Note that the SDF network from MetaAvatar can be trained with canonical meshes only and does not need any posed meshes as supervision. Each MLP of the hypernetwork (top) consists of one hidden layer with 256 channels and uses ReLU activation. The neural SDF (middle) consists of 5 hidden layers with 256 channels and uses a periodic activation~\cite{sitzmann2019siren}. In addition to the MetaAvatar SDF, we add a mapping network~\cite{chanmonteiro2020pi-GAN,perez2018film} which consists of 2 hidden layers with 256 channels and a ReLU activation. It maps the per-frame latent code $\mathcal{Z}$ to scaling factors and offsets that modulate the outputs from each layer of the neural SDF. We initialize the last layer of the mapping network to predict scaling factors with value 1 and offsets with value 0. 

\subsection{Canonical Color Network}
\label{appx:color_net_arch}
\begin{figure}[t]
\centering
  \includegraphics[width=1.0\textwidth]{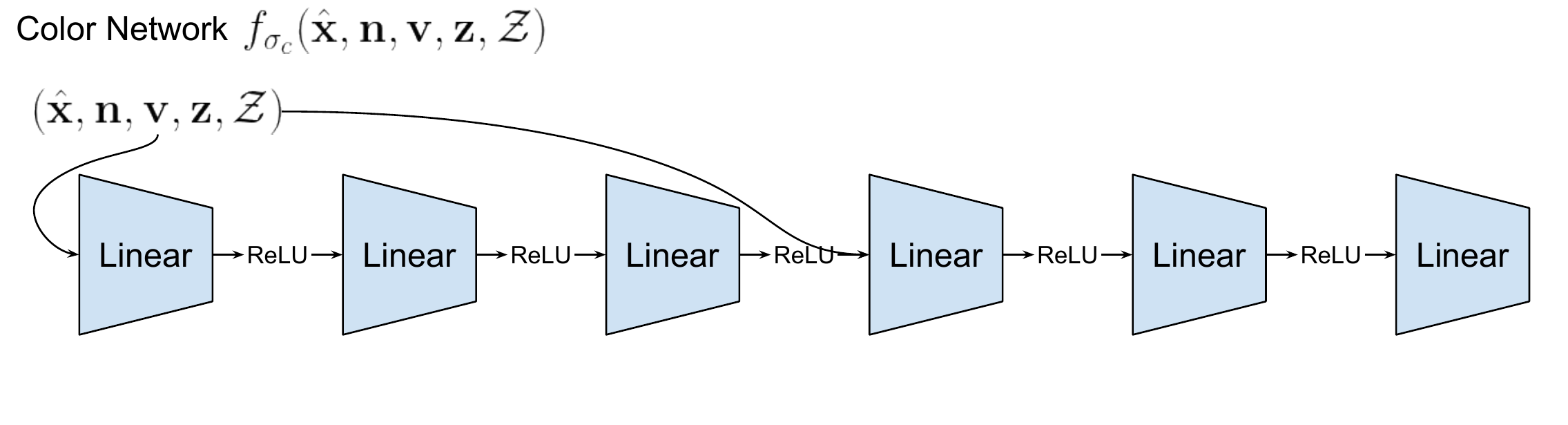}
  \caption{\textbf{Network Architecture for the Color Network.} The color network takes canonicalized query points $\hat{\mathbf{x}}$, normal vectors $\mathbf{n}$, viewing directions $\mathbf{v}$, an SDF feature $\mathbf{z}$, and a per-frame latent code $\mathcal{Z}$ as inputs.}
  \label{appx:fig:color_net_arch}
\end{figure}
We describe our canonical color network in Fig.~\ref{appx:fig:color_net_arch}. The network consists of 4 hidden layers with 256 channels and ReLU activation. The inputs to the network are also concatenated with activations of the third layer and fed into the fourth layer together. 
\section{Implicit Gradients}
\label{appx:implicit_gradients}
In this section, we describe how to compute gradients of the root-finding solutions \wrt the forward LBS network and the SDF network. In the main paper, we use our novel joint root-finding algorithm to find the surface point and sample points around the surface point; these sampled points, along with the surface point, are mapped to canonical space via iterative root-finding~\cite{Chen2021ICCV}. Section~\ref{appx:grad_LBS} describes how to differentiate through these points to compute gradients \wrt the forward LBS network. Section~\ref{appx:grad_joint_root_finding} describes how to compute gradients \wrt the forward LBS network and the SDF network given the surface point and its correspondence. Section~\ref{appx:grad_LBS} is used for volume rendering, which is described in Section\iftoggle{arXiv}{~\ref{sec:vol_sdf}}{~\textcolor{red}{3.4}} of the main paper. Section~\ref{appx:grad_joint_root_finding} is used for surface rendering, which is one of our ablation baselines in Section~\ref{appx:ablation}.

\subsection{Implicit Gradients for Forward LBS}
\label{appx:grad_LBS}
Here we follow~\cite{Chen2021ICCV} and describe how to compute implicit gradients for the forward LBS network given samples on camera rays and their canonical correspondences. Denoting sampled points in observation space as $\bar{\mathbf{X}} = \{ \bar{\mathbf{x}} \}_{i=1}^{N}$, and their canonical correspondences obtained by iterative root-finding~\cite{Chen2021ICCV} as $\hat{\mathbf{X}}^* = \{ \hat{\mathbf{x}}^* \}_{i=1}^{N}$, they should satisfy the following condition
\begin{align}
    \label{appx:eqn:LBS}
    LBS_{\sigma_{\omega}}(\hat{\mathbf{x}}^{*(i)} ) - \bar{\mathbf{x}}^{(i)} = 0, \quad \forall i = 1, \ldots, N
\end{align}
As done in~\cite{yariv2020multiview}, by applying implicit differentiation, we obtain a differentiable point sample $\hat{\mathbf{x}}$ as
\begin{align}
\label{appx:eqn:differentiable_samples_lbs}
    \hat{\mathbf{x}}
    = \hat{\mathbf{x}}^* - (\mathbf{J}^*)^{-1} \cdot \left(LBS_{\sigma_{\omega}}(\hat{\mathbf{x}}^{*(i)} ) - \bar{\mathbf{x}}^{(i)} \right)
\end{align}
where $ \mathbf{J}^* = \frac{\partial LBS_{\sigma_{\omega}}}{\partial \hat{\mathbf{x}}} (\hat{\mathbf{x}}^*) $. $\hat{\mathbf{x}}^*$ and $\mathbf{J}^*$ are detached from the computational graph such that no gradient will flow through them. These differentiable samples can be used as inputs to the SDF and color networks. Gradients \wrt $\sigma_{\omega}$ are computed from photometric loss Eq.~\eqref{appx:eqn:rgb_loss} via standard back-propagation. Taking the derivative \wrt $\sigma_{\omega}$ for both sides of Eq.~\eqref{appx:eqn:differentiable_samples_lbs} results in the same analytical gradient defined in Eq.~(\textcolor{red}{14}) of~\cite{Chen2021ICCV}. 

\begin{figure}
\centering
 \begin{subfigure}[b]{1.0\textwidth}
    \includegraphics [width=1.0\textwidth]{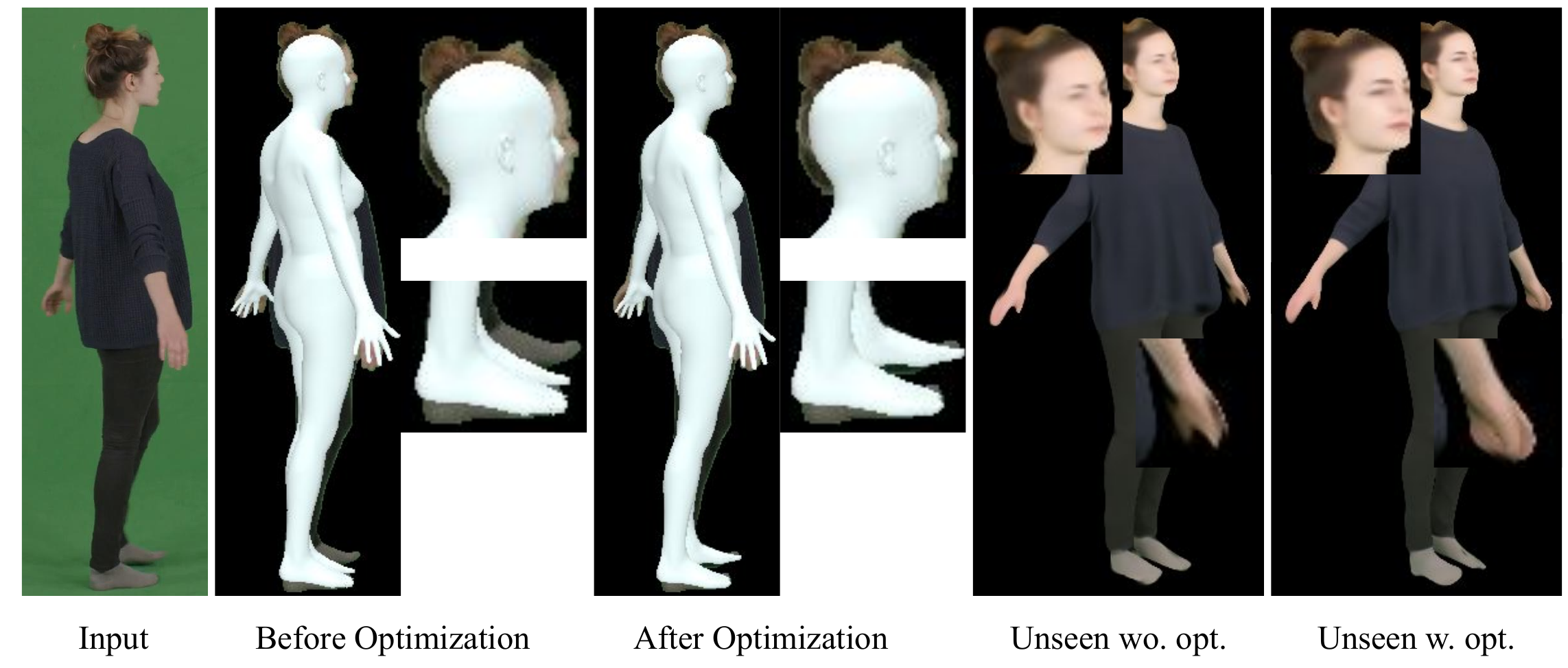}
\end{subfigure}
\caption{\textbf{Result of Pose and Shape Optimization.} We can improve the noisy SMPL estimations on training poses with implicit gradients and improve the rendering quality on unseen poses (see Unseen w. opt.).}
\label{appx:fig:people_snapshot}
\end{figure}

\boldparagraph{Pose and Shape Optimization} We note that implicit gradients can also be back-propagated to SMPL parameters $\{ \mathbf{\theta}, \mathbf{\beta} \}$ as the SMPL model is fully differentiable. We found pose and shape optimization particularly helpful when SMPL estimations are noisy, \eg\ those estimated from monocular videos. In Fig.~\ref{appx:fig:people_snapshot} we show a qualitative sample on the People Snapshot dataset~\cite{alldieck2018video} where the pose is improved while the resulting model also achieves better visual quality.

\subsection{Implicit Gradients for Joint Root-finding}
\label{appx:grad_joint_root_finding}
Now we derive implicit gradients for our joint root-finding algorithm. We denote the joint vector-valued function of the ray-surface intersection and forward LBS as $g_{\sigma_f, \sigma_{\omega}} (\hat{\mathbf{x}}, d)$. The joint root-finding problem is
\begin{align}
\label{appx:eqn:joint_root_finding}
    g_{\sigma_f, \sigma_{\omega}} (\hat{\mathbf{x}}, d) =
    \begin{bmatrix}
    f_{\sigma_f} (\hat{\mathbf{x}}) \\ LBS_{\sigma_{\omega}}(\hat{\mathbf{x}}) - (\mathbf{c} + \mathbf{v} \cdot d)
    \end{bmatrix} = \mathbf{0}
\end{align}
with a slight abuse of notation, we denote the iso-surface point as $\hat{\mathbf{x}}^*$ and their corresponding depth in observation space as $d^*$. We follow~\cite{yariv2020multiview} and use implicit differentiation to obtain a differentiable point sample $\hat{\mathbf{x}}$ and a depth sample $d$:
\begin{align}
\label{appx:eqn:differentiable_samples}
    \begin{bmatrix}
    \hat{\mathbf{x}} \\ d
    \end{bmatrix} =
    \begin{bmatrix}
    \hat{\mathbf{x}}^* \\ d^*
    \end{bmatrix}
    - (\mathbf{J}^*)^{-1} \cdot g_{\sigma_f, \sigma_{\omega}} (\hat{\mathbf{x}}^*, d^*)
\end{align}
where $\mathbf{J}^*$ is defined as
\begin{align}
    \mathbf{J}^* =
    \begin{bmatrix}
    \frac{\partial f_{\sigma_f}}{\partial \hat{\mathbf{x}}} (\hat{\mathbf{x}}^*) & 0 \\[6pt] \frac{\partial LBS_{\sigma_{\omega}}}{\partial \hat{\mathbf{x}}} (\hat{\mathbf{x}}^*) & -\mathbf{v}
    \end{bmatrix}
\end{align}
Similar to Section~\ref{appx:grad_LBS}, these differentiable samples can be used as inputs to the SDF and color networks and gradients \wrt $\sigma_{f}, \sigma_{\omega}$ can be computed from the photometric loss Eq.~\eqref{appx:eqn:rgb_loss}.
\section{Implementation Details}
\label{appx:implementation}
We use Adam~\cite{Adam:ICLR:2015} to optimize our models and the per-frame latent codes $\{ \mathcal{Z} \}$. 
We initialize the SDF network with MetaAvatar~\cite{MetaAvatar:NeurIPS:2021} and set the learning rate to $1e^{-6}$ as suggested in~\cite{MetaAvatar:NeurIPS:2021}. 
For the remaining models and the latent codes, we use a learning rate of $1e^{-4}$. We apply weight decay with a weight of $0.05$ to the per-frame latent codes.

We train our models with a batch size of 4 and 2048 rays per batch, with 1024 rays sampled from the foreground mask and 1024 rays sampled from the background. As mentioned in Section\iftoggle{arXiv}{~\ref{sec:vol_sdf}}{~\textcolor{red}{3.4}} of the main paper, we sample 16 near and 16 far surface points for rays that intersect with a surface and 64 points for rays that do not intersect with a surface. Our model is trained for 250 epochs (except for sequence 313 which we trained for 1250 epochs, due to its training frames being much fewer than other sequences), which corresponds to 60k-80k iterations depending on the amount of training data. This takes about 1.5 days on 4 NVIDIA 2080 Ti GPUs. During training, we follow~\cite{HNeRF:NeurIPS:2021} and add normally distributed noise with zero mean and a standard deviation of 0.1 to the input $\theta$ of the SDF network. This noise ensures that the canonical SDF does not fail when given extreme out-of-distribution poses. We also augment the input viewing directions to the color network during training. We do so by randomly applying roll/pitch/yaw rotations sampled from a normal distribution with zero mean and a standard deviation of $45^{\circ}$ to the viewing direction, but reject augmentation in which the angle between the estimated surface normal and the negated augmented viewing direction is greater than 90 degrees.

For inference, we follow~\cite{peng2020neural,peng2021animatable} and crop an enlarged bounding box around the projected SMPL mesh on the image plane and render only pixels inside the bounding box. For unseen test poses we follow the practice of~\cite{peng2020neural,peng2021animatable} and use the latent code $\mathcal{Z}$ of the last training frame as the input. The rendering time of a $512 \times 512$ image is about 10-20 seconds, depending on the bounding box size of the person. In this process, the proposed joint root-finding algorithm takes about 1 second.
\section{Implementation Details for Baselines}
\label{appx:impl_baselines}
In this section, we describe the implementation details of the baselines from the main paper, \ie Neural Body~\cite{peng2020neural}, Ani-NeRF~\cite{peng2021animatable}, and A-NeRF~\cite{ANeRF:NeurIPS:2021}.

\subsection{Neural Body}
For quantitative evaluation, we use the official results provided by the Neural Body website. For generating rendering results and geometries, we use the official code of Neural Body and their pre-trained models without modification.  

\subsection{Animatable NeRF (Ani-NeRF)}
For quantitative evaluation, we use the official code and pre-trained models when possible, \ie for sequences 313, 315, 377, and 386. For the remaining sequences that the official code does not provide pre-trained models, we train models using the default hyperparameters that were applied to sequences 313, 315, 377, and 386.

We note that when reconstructing geometry on the training poses, Neural Body and Ani-NeRF compute visual hulls from ground-truth masks of training views and set density values outside the visual hulls to 0. This removes extraneous geometry blobs from reconstructions by Neural Body and Ani-NeRF. When testing on unseen poses, we disable the mask usage, as, by definition of the task, we do not have any image as input.

\subsection{A-NeRF}
For A-NeRF, we follow the author's suggestions to 1) use a bigger foreground mask for ray sampling, 2) enable background estimation in the official code, and 3) use L2 loss instead of L1 loss. The learned models give reasonable novel view synthesis results on training poses (Fig.~\ref{appx:fig:qualitative_results_train}) but cannot generalize to unseen poses (Fig.~\ref{appx:fig:qualitative_results_test}). We hypothesize that this is because training poses on the ZJU-MoCap dataset are extremely limited, and A-NeRF uses only keypoints instead of surface models to construct their conditional inputs to NeRF networks. The lack of a surface model makes it easy for A-NeRF to confuse background and foreground, resulting in obvious floating blob artifacts. These artifacts are amplified when training poses are limited, making the generalization result of A-NeRF on the ZJU-MoCap dataset the worst among the baselines.
\section{Ablation Study}
\label{appx:ablation}
In this section, we ablate on ray sampling strategies as well as canonicalization strategies. We conduct an ablation on sequence 313. Metrics on all novel views of training poses are reported.

\subsection{Ablation on Ray Sampling Strategies}
\label{appx:ablate_ray_sample}
\begin{figure}[t]
\centering
  \includegraphics[trim=0 40 0 0, clip, width=1.0\textwidth]{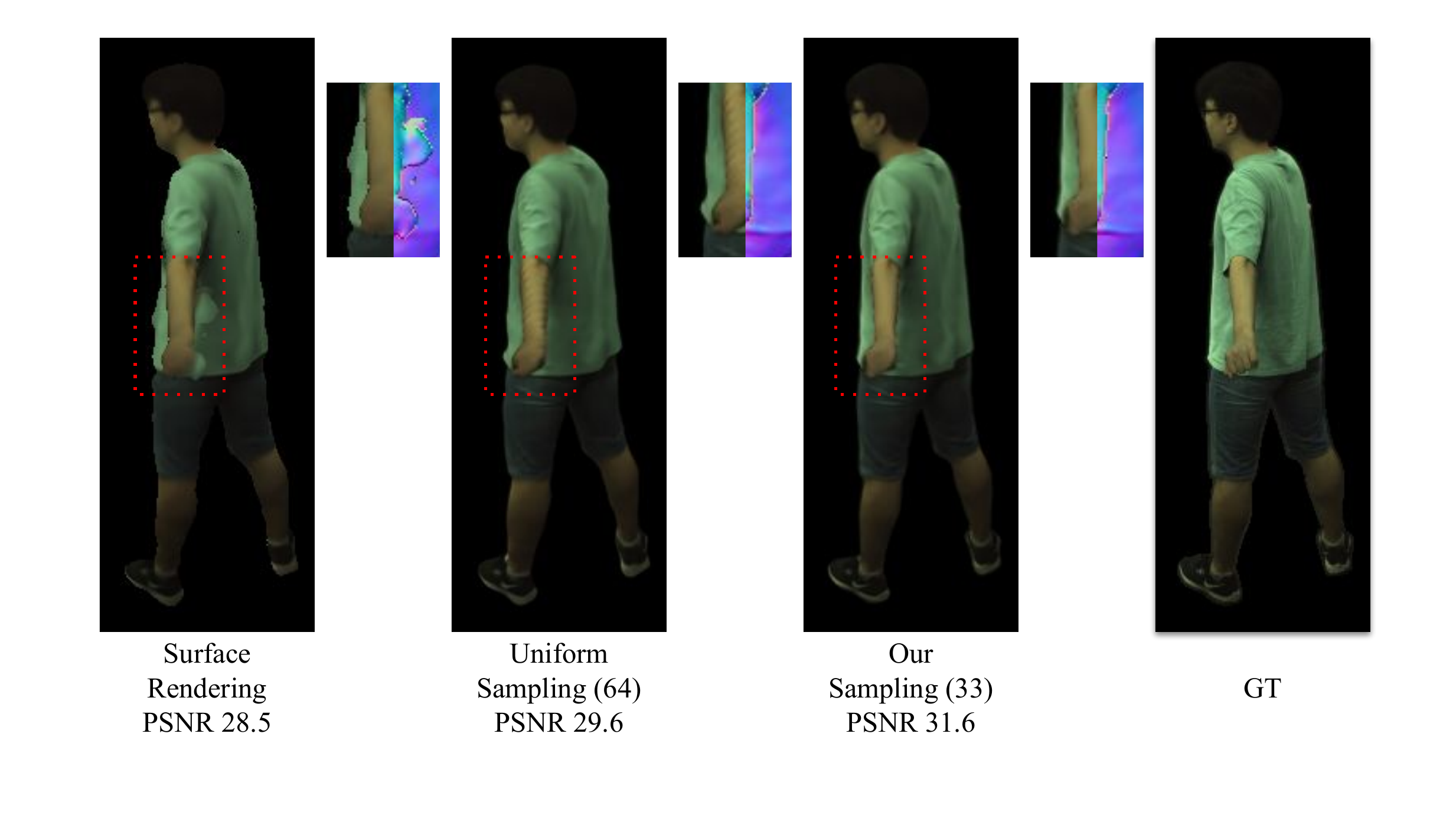}
  \caption{\textbf{Ablation on ray sampling strategies}. We observe severe geometric artifacts with models trained with surface rendering. A simple uniform sampling strategy (as used in~\cite{peng2020neural,peng2021animatable}) produces stratified artifacts due to the discretized sampling. In contrast, our proposed approach does not suffer from these problems and achieves better result.}
  \label{fig:ablate_ray_sample}
\end{figure}
We compare our proposed ray sampling strategy to surface rendering and uniform sampling with 64 samples on the novel view synthesis task (Fig~\ref{fig:ablate_ray_sample}). As discussed in the main paper, we did not use more sophisticated hierarchical sampling strategies~\cite{volsdf:NeurIPS:2021,wang2021neus,mildenhall2020nerf} due to the computational cost of running the iterative root-finding~\cite{Chen2021ICCV} on dense samples and the memory cost for running additional forward/backward passes through the LBS network.

\subsection{Ablation on Learned forward LBS}
\label{appx:ablate_lbs}
\begin{figure}[t]
\centering
  \includegraphics[trim=0 0 0 0, clip, width=1.0\textwidth]{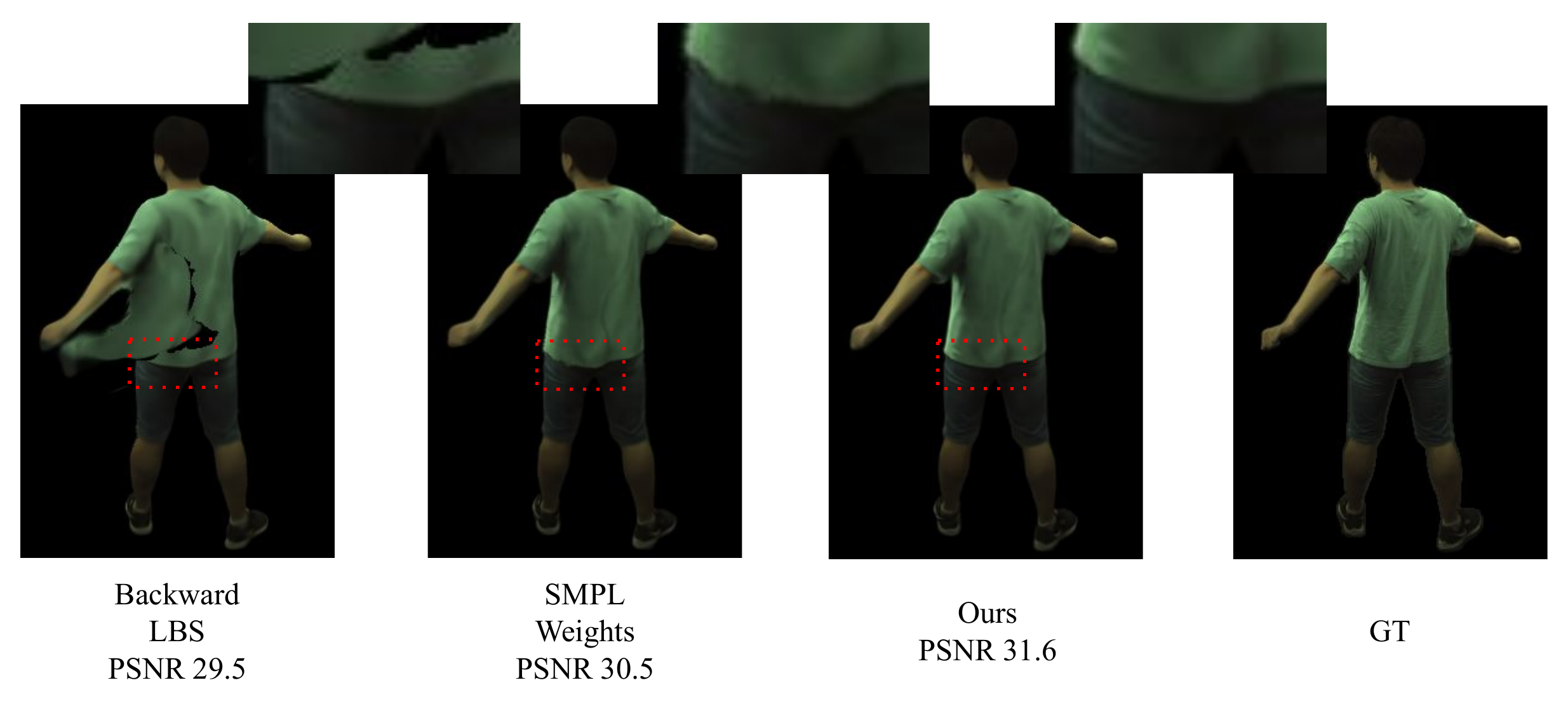}
  \caption{\textbf{Ablation on Learned LBS networks}. Backward LBS has difficulties with learning skinning weights for points far from the surface, resulting in artifacts under specific poses. Canonicalization with deterministic SMPL weights results in discretized artifacts on the cloth surface. In contrast, our approach does not suffer from these problems.}
  \label{appx:fig:ablate_skinning}
\end{figure}
In this subsection, we replace our learned forward LBS with (1) a backward LBS network that conditions on local body poses $\theta$, and (2) a deterministic LBS with nearest neighbor SMPL skinning weights. For the learned backward LBS, we always canonicalize the query points using the SMPL global translation and rotation before querying the LBS network. We also sample points on the transformed SMPL meshes and supervise the backward LBS network with corresponding skinning weights using Eq.~\eqref{appx:eqn:skin_loss}. We show qualitative results in Fig.~\ref{appx:fig:ablate_skinning}.

\subsection{Ablation on Root-finding Initialization}
\label{appx:ablate_init}
To ablate the effect of multiple initializations for root-finding, we add additional initializations from the nearest 2 SMPL bones but do not observe any noticeable change in metrics. We report PSNR/SSIM/LPIPS as: single initialization - 31.6/0.973/0.050, 2 more initializations: 31.5/0.972/0.049. Also, adding more initializations for root-finding drastically increases memory/time consumption, we thus decide to use only a single initialization for root-finding in our approach.
\section{Additional Quantitative Results}
\label{appx:additional_quantitative}
We present complete evaluation metrics including PSNR, SSIM, LPIPS on the test poses of the ZJU-MoCap~\cite{peng2020neural} dataset in Table~\ref{appx:tab:novel_pose}.
\begin{table}
\scriptsize
 \caption{\textbf{Complete evaluation results on novel pose synthesis.} PSNR, SSIM, LPIPS are reported for the test poses of the ZJU-MoCap dataset.}
 \iftoggle{arXiv}{\vspace{0.05in}}{}
 \label{appx:tab:novel_pose}
 \centering
 \begin{tabular}{ | c | c | c | c |  c | c | c  | c | c | c | }
 \hline
 & \multicolumn{3}{c|}{313} & \multicolumn{3}{c|}{315} & \multicolumn{3}{c|}{377}
 \\\hline
  Method & PSNR $\uparrow$ & SSIM $\uparrow$ & LPIPS $\downarrow$ & PSNR $\uparrow$ & SSIM $\uparrow$ & LPIPS $\downarrow$ & PSNR $\uparrow$ & SSIM $\uparrow$ & LPIPS $\downarrow$ \\
 \hline
 NB & 24.1 & 0.908 & 0.126 & 19.8 & 0.867 & 0.152 & 24.2 & 0.917 & 0.119 \\
 Ani-N & 23.9 & 0.909 & 0.115 & 19.2 & 0.855 & 0.167 & 22.6 & 0.900 & 0.153 \\
 A-NeRF & 22.0 & 0.855 & 0.209 & 18.7 & 0.810 & 0.232 & 22.6 & 0.890 & 0.165 \\
 Ours & \textbf{24.4} & \textbf{0.914} & \textbf{0.092} & \textbf{20.0} & \textbf{0.881} & \textbf{0.105} & \textbf{25.5} & \textbf{0.933} & \textbf{0.093}\\
 \hline
 & \multicolumn{3}{c|}{386} & \multicolumn{3}{c|}{387} & \multicolumn{3}{c|}{390}
 \\\hline
  Method & PSNR $\uparrow$ & SSIM $\uparrow$ & LPIPS $\downarrow$ & PSNR $\uparrow$ & SSIM $\uparrow$ & LPIPS $\downarrow$ & PSNR $\uparrow$ & SSIM $\uparrow$ & LPIPS $\downarrow$ \\
 \hline
 NB & 26.1 & 0.894 & 0.171 & 22.7 & 0.902 & 0.135 & 24.2 & 0.882 & 0.164 \\
 Ani-N & 25.5 & 0.884 & 0.187 & 23.1 & 0.906 & 0.145 & 23.9 & 0.887 & 0.173 \\
 A-NeRF & 24.8 & 0.858 & 0.241 & 22.4 & 0.885 & 0.162 & 22.6 & 0.846 & 0.226 \\
 Ours & \textbf{27.0} & \textbf{0.910} & \textbf{0.127} & \textbf{24.2} & \textbf{0.917} & \textbf{0.099} & \textbf{24.8} & \textbf{0.896} & \textbf{0.126} \\
 \hline
 & \multicolumn{3}{c|}{392} & \multicolumn{3}{c|}{393} & \multicolumn{3}{c|}{394}
 \\\hline
  Method & PSNR $\uparrow$ & SSIM $\uparrow$ & LPIPS $\downarrow$ & PSNR $\uparrow$ & SSIM $\uparrow$ & LPIPS $\downarrow$ & PSNR $\uparrow$ & SSIM $\uparrow$ & LPIPS $\downarrow$ \\
 \hline
 NB & 26.0 & 0.916 & 0.135 & 23.5 & 0.900 & 0.132 & 24.1 & 0.888 & 0.150 \\
 Ani-N & 24.3 & 0.900 & 0.169 & 23.8 & 0.897 & 0.155 & 24.1 & 0.887 & 0.171 \\
 A-NeRF & 23.7 & 0.886 & 0.183 & 22.1 & 0.875 & 0.175 & 22.7 & 0.861 & 0.199 \\
 Ours & \textbf{26.2} & \textbf{0.927} & \textbf{0.106} & \textbf{24.4} & \textbf{0.915} & \textbf{0.104} & \textbf{25.2} & \textbf{0.908} & \textbf{0.111} \\
 \hline
 \end{tabular}
\end{table}

We also report quantitative results on the H36M dataset~\cite{h36m_pami}, following the testing protocols proposed by~\cite{peng2021animatable} in Table~\ref{tab:eval_h36m}.

\begin{table}
 \begin{center}
\scriptsize
 \caption{\textbf{Evaluation results on the H36M dataset.} Numbers of NARF~\cite{NARF:ICCV:2021} and Ani-N~\cite{peng2021animatable} are reported in~\cite{xu2022sanerf}.}
 \iftoggle{arXiv}{\vspace{0.05in}}{}
 \label{tab:eval_h36m}
 \begin{tabular}{| c | c | c | c | c | c | c | c | c | c | c | c | c |}
 \hline
& \multicolumn{6}{c|}{\textbf{Training Poses}} & \multicolumn{6}{c|}{\textbf{Unseen Poses}} \\ \hline
& \multicolumn{3}{c|}{PSNR $\uparrow$} & \multicolumn{3}{c|}{SSIM $\uparrow$} & \multicolumn{3}{c|}{PSNR $\uparrow$} & \multicolumn{3}{c|}{SSIM $\uparrow$} \\ \hline
 & NARF & Ani-N & Ours & NARF & Ani-N & Ours & NARF & Ani-N & Ours & NARF & Ani-N & Ours \\ \hline
 S1 & 21.41 & 22.05 & \textbf{24.45} & 0.891 & 0.888 & \textbf{0.919} & 20.19 & 21.37 & \textbf{23.08} & 0.864 & 0.868 & \textbf{0.899} \\
 S5 & \textbf{25.24} & 23.27 & 24.54 & 0.914 & 0.892 & \textbf{0.918} & \textbf{23.91} & 22.29 & 22.79 & \textbf{0.891} & 0.875 & 0.890 \\
 S6 & 21.47 & 21.13 & \textbf{24.61} & 0.871 & 0.854 & \textbf{0.903} & 22.47 & 22.59 & \textbf{24.04} & 0.883 & 0.884 & \textbf{0.900} \\
 S7 & 21.36 & 22.50 & \textbf{24.31} & 0.899 & 0.890 & \textbf{0.919} & 20.66 & 22.22 & \textbf{22.58} & 0.876 & 0.878 & \textbf{0.891} \\
 S8 & 22.03 & 22.75 & \textbf{24.02} & 0.904 & 0.898 & \textbf{0.921} & 21.09 & 21.78 & \textbf{22.34} & 0.887 & 0.882 & \textbf{0.896} \\
 S9 & 25.11 & 24.72 & \textbf{26.20} & 0.906 & 0.908 & \textbf{0.924} & 23.61 & 23.72 & \textbf{24.36} & 0.881 & 0.886 & \textbf{0.894} \\
 S11 & 24.35 & 24.55 & \textbf{25.43} & 0.902 & 0.902 & \textbf{0.921} & 23.95 & 23.91 & \textbf{24.78} & 0.885 & 0.889 & \textbf{0.902} \\ \hline
 Average & 23.00 & 23.00 & \textbf{24.79} & 0.898 & 0.890 & \textbf{0.918} & 22.27 & 22.55 & \textbf{23.42} & 0.881 & 0.880 & \textbf{0.896} \\ \hline
 \end{tabular}
 \end{center}
\end{table}
\section{Additional Qualitative Results}
\label{appx:additional_qualitative}
\subsection{Qualitative Results on ZJU-MoCap Training Poses}
\label{appx:additional_qualitative_train}
\begin{figure}[t]
\captionsetup[subfigure]{labelformat=empty}
\centering
\begin{subfigure}[b]{0.17\textwidth}
    \includegraphics [trim=100 20 170 120, clip, width=1.0\textwidth]{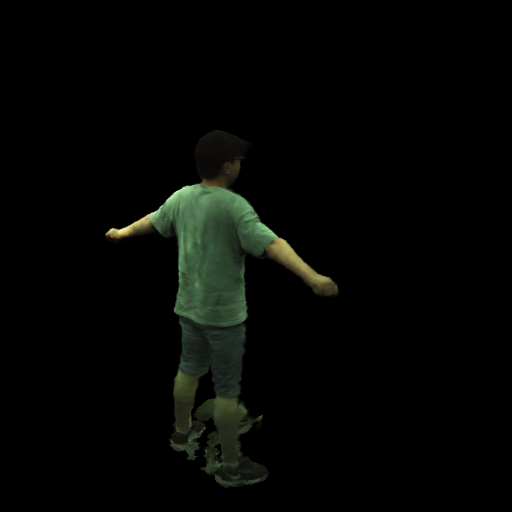}
\end{subfigure}
\begin{subfigure}[b]{0.17\textwidth}
    \includegraphics [trim=100 20 170 120, clip, width=1.0\textwidth]{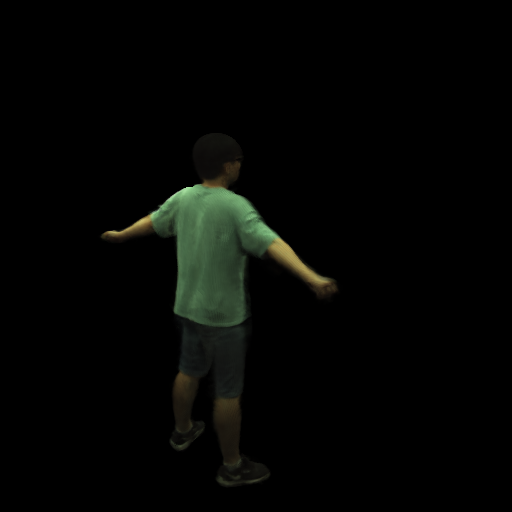}
\end{subfigure}
\begin{subfigure}[b]{0.17\textwidth}
    \includegraphics [trim=100 20 170 120, clip, width=1.0\textwidth]{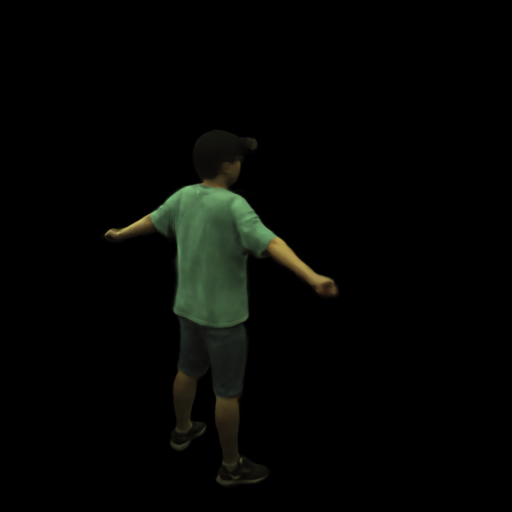}
\end{subfigure}
\begin{subfigure}[b]{0.17\textwidth}
    \includegraphics [trim=100 20 170 120, clip, width=1.0\textwidth]{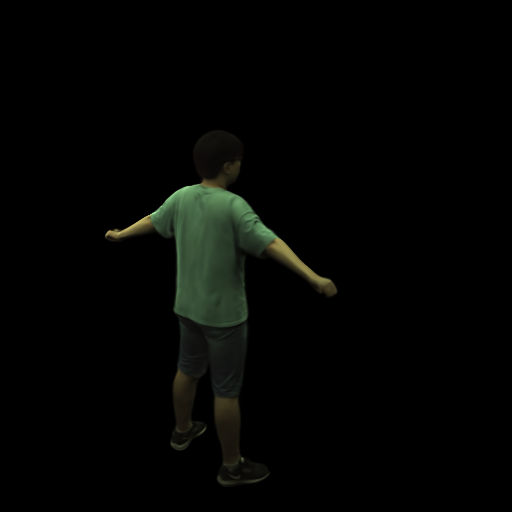}
\end{subfigure}
\begin{subfigure}[b]{0.17\textwidth}
    \includegraphics [trim=100 20 170 120, clip, width=1.0\textwidth]{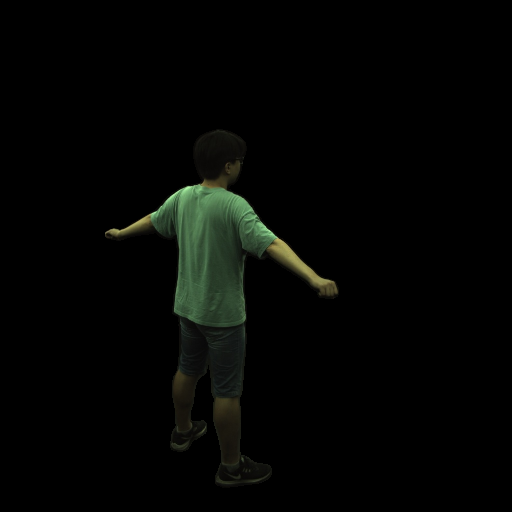}
\end{subfigure} \\
\begin{subfigure}[b]{0.17\textwidth}
    \includegraphics [trim=230 150 130 50, clip, width=1.0\textwidth]{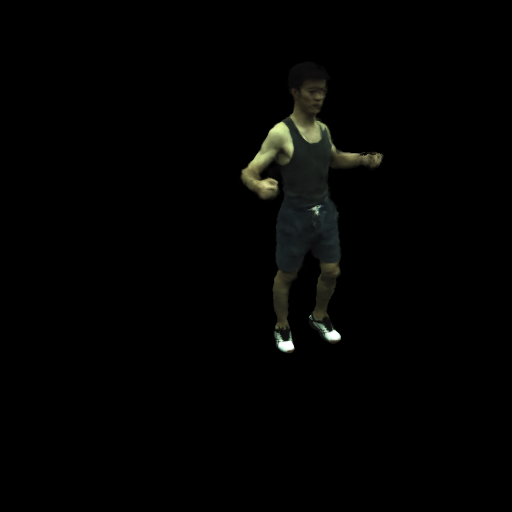}
\end{subfigure}
\begin{subfigure}[b]{0.17\textwidth}
    \includegraphics [trim=230 150 130 50, clip, width=1.0\textwidth]{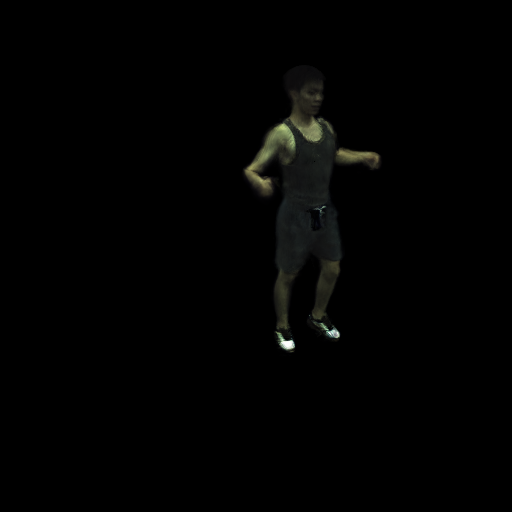}
\end{subfigure}
\begin{subfigure}[b]{0.17\textwidth}
    \includegraphics [trim=230 150 130 50, clip, width=1.0\textwidth]{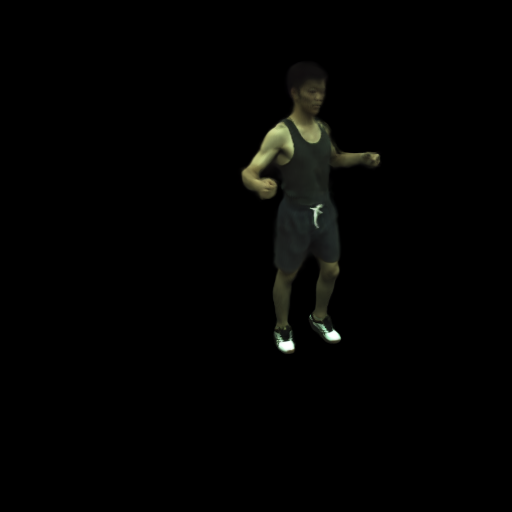}
\end{subfigure}
\begin{subfigure}[b]{0.17\textwidth}
    \includegraphics [trim=230 150 130 50, clip, width=1.0\textwidth]{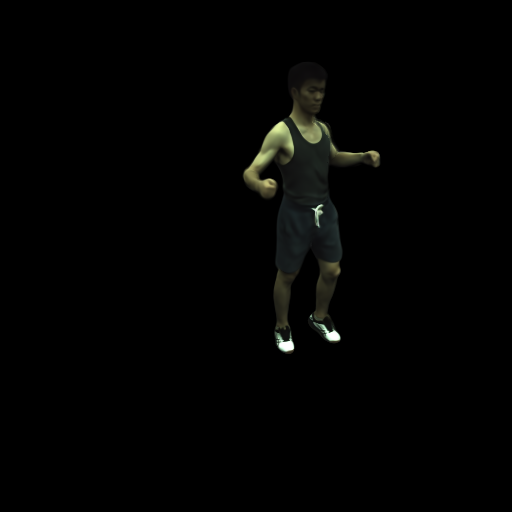}
\end{subfigure}
\begin{subfigure}[b]{0.17\textwidth}
    \includegraphics [trim=230 150 130 50, clip, width=1.0\textwidth]{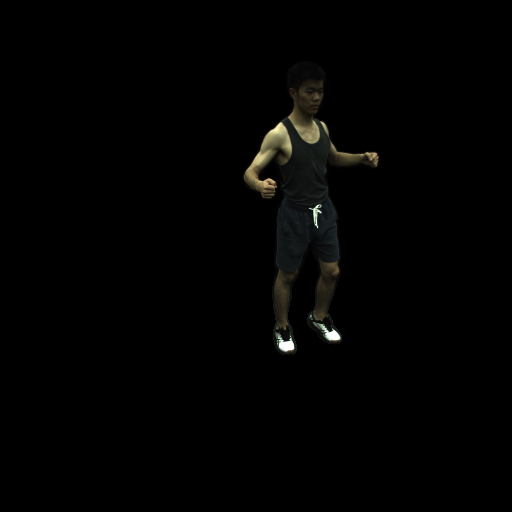}
\end{subfigure} \\
\begin{subfigure}[b]{0.17\textwidth}
    \includegraphics [trim=110 90 140 80, clip, width=1.0\textwidth]{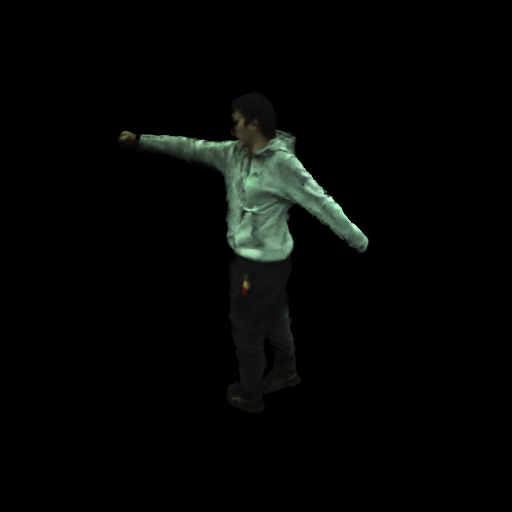}
\end{subfigure}
\begin{subfigure}[b]{0.17\textwidth}
    \includegraphics [trim=110 90 140 80, clip, width=1.0\textwidth]{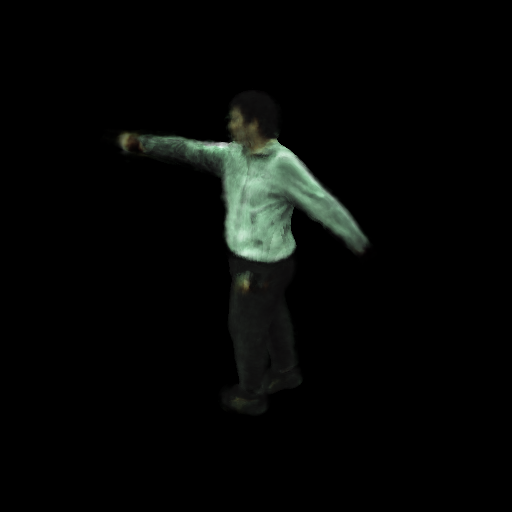}
\end{subfigure}
\begin{subfigure}[b]{0.17\textwidth}
    \includegraphics [trim=110 90 140 80, clip, width=1.0\textwidth]{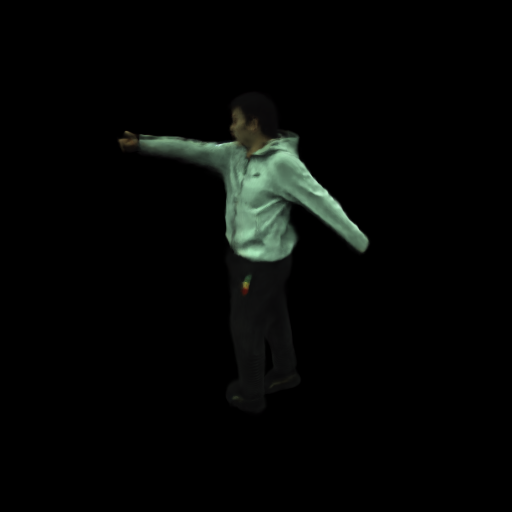}
\end{subfigure}
\begin{subfigure}[b]{0.17\textwidth}
    \includegraphics [trim=110 90 140 80, clip, width=1.0\textwidth]{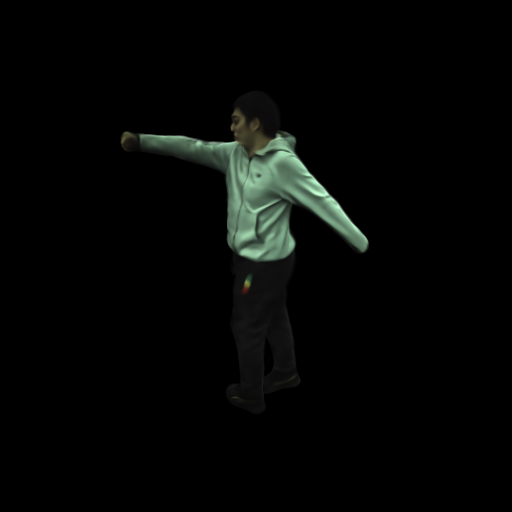}
\end{subfigure}
\begin{subfigure}[b]{0.17\textwidth}
    \includegraphics [trim=110 90 140 80, clip, width=1.0\textwidth]{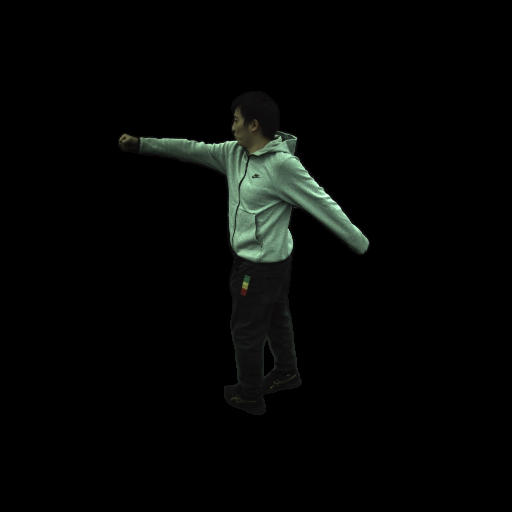}
\end{subfigure} \\
\begin{subfigure}[b]{0.17\textwidth}
    \includegraphics [trim=130 110 170 70, clip, width=1.0\textwidth]{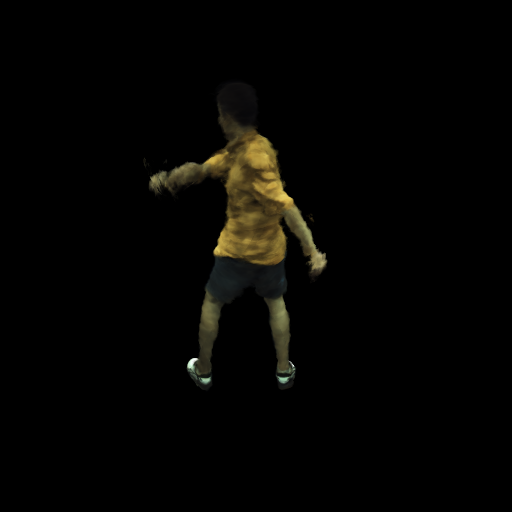}
\end{subfigure}
\begin{subfigure}[b]{0.17\textwidth}
    \includegraphics [trim=130 110 170 70, clip, width=1.0\textwidth]{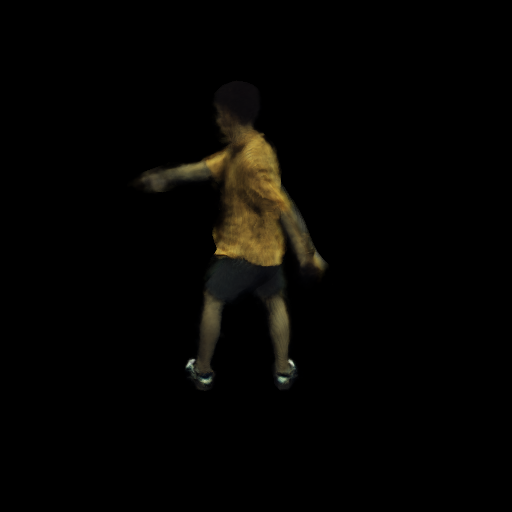}
\end{subfigure}
\begin{subfigure}[b]{0.17\textwidth}
    \includegraphics [trim=130 110 170 70, clip, width=1.0\textwidth]{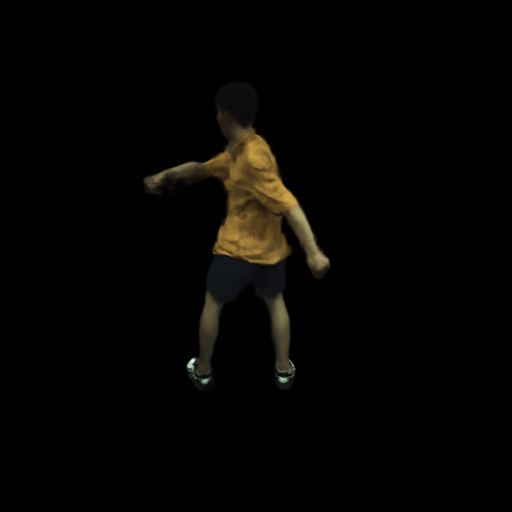}
\end{subfigure}
\begin{subfigure}[b]{0.17\textwidth}
    \includegraphics [trim=130 110 170 70, clip, width=1.0\textwidth]{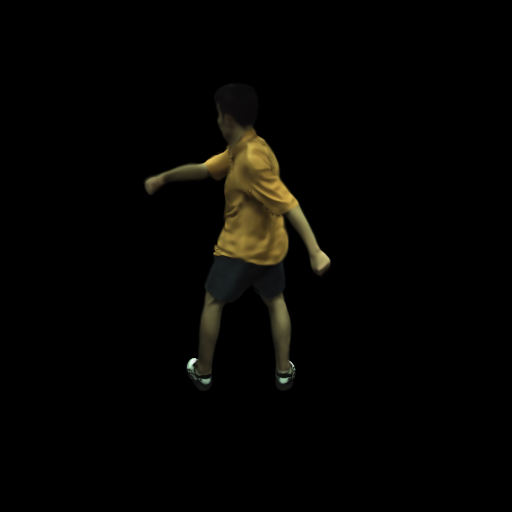}
\end{subfigure}
\begin{subfigure}[b]{0.17\textwidth}
    \includegraphics [trim=130 110 170 70, clip, width=1.0\textwidth]{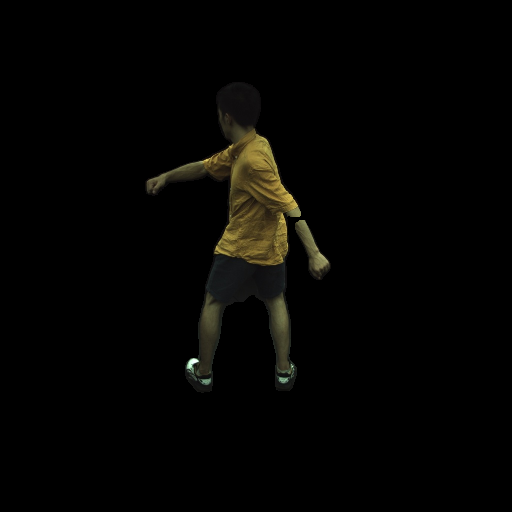}
\end{subfigure} \\
\begin{subfigure}[b]{0.17\textwidth}
    \includegraphics [trim=200 120 160 90, clip, width=1.0\textwidth]{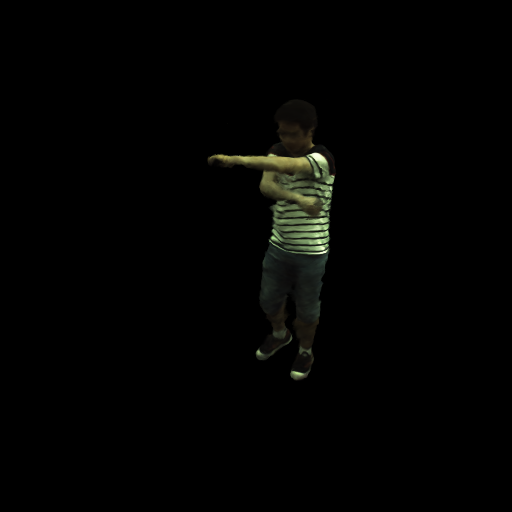}
    \caption{A-NeRF}
\end{subfigure}
\begin{subfigure}[b]{0.17\textwidth}
    \includegraphics [trim=200 120 160 90, clip, width=1.0\textwidth]{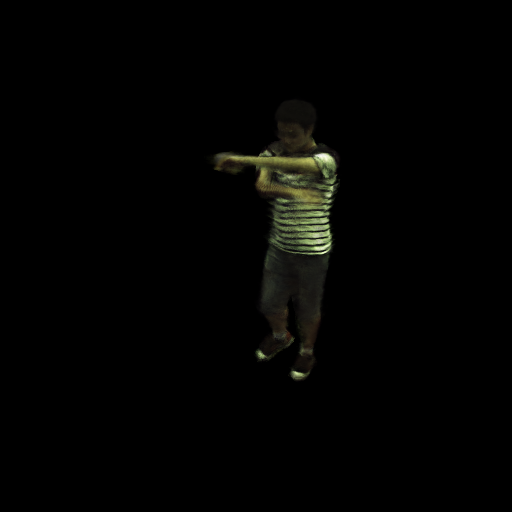}
    \caption{Ani-NeRF}
\end{subfigure}
\begin{subfigure}[b]{0.17\textwidth}
    \includegraphics [trim=200 120 160 90, clip, width=1.0\textwidth]{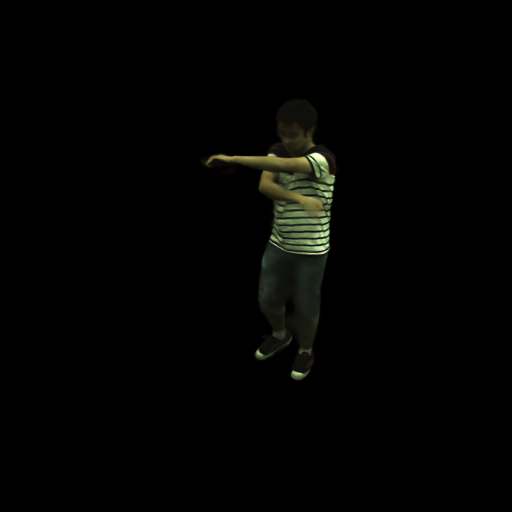}
    \caption{Neural Body}
\end{subfigure}
\begin{subfigure}[b]{0.17\textwidth}
    \includegraphics [trim=200 120 160 90, clip, width=1.0\textwidth]{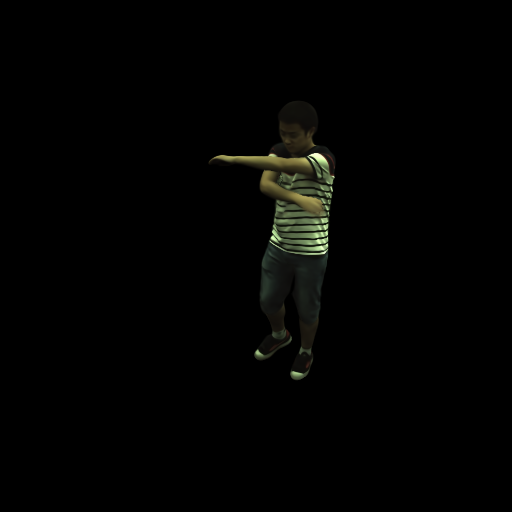}
    \caption{Ours}
\end{subfigure}
\begin{subfigure}[b]{0.17\textwidth}
    \includegraphics [trim=200 120 160 90, clip, width=1.0\textwidth]{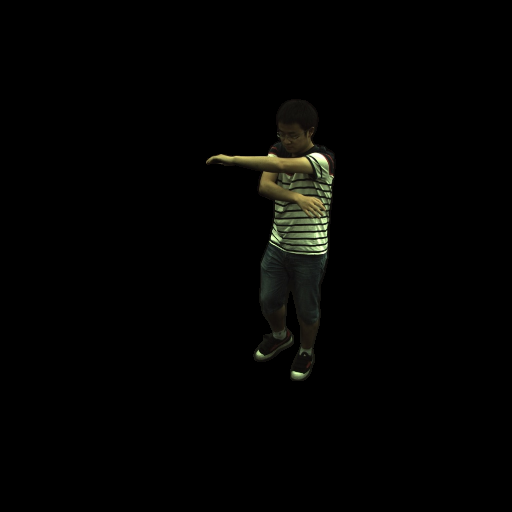}
    \caption{GT}
\end{subfigure}
\caption{\textbf{Novel View Synthesis Results} on the training poses of ZJU-MoCap.}
\label{appx:fig:qualitative_results_train}
\end{figure}
We present additional qualitative results on ZJU-MoCap training poses in Fig.~\ref{appx:fig:qualitative_results_train}. Due to better geometry constraints, our approach better captures cloth wrinkles, textures, and face details. We also avoid extraneous color blobs under novel views which all baselines suffer from.

\subsection{Additional Qualitative Results on ZJU-MoCap Test Poses}
\label{appx:additional_qualitative_test}
\begin{figure}[t]
\captionsetup[subfigure]{labelformat=empty}
\centering
\begin{subfigure}[b]{0.17\textwidth}
    \includegraphics [trim=190 80 180 70, clip, width=1.0\textwidth]{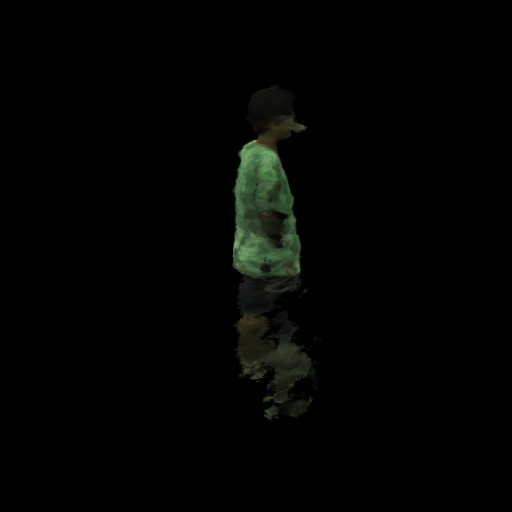}
\end{subfigure}
\begin{subfigure}[b]{0.17\textwidth}
    \includegraphics [trim=190 80 180 70, clip, width=1.0\textwidth]{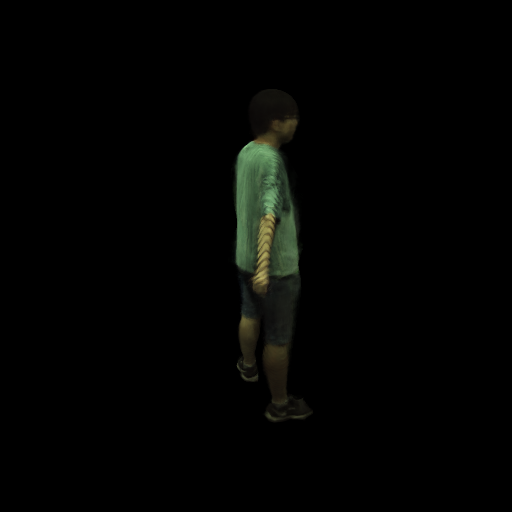}
\end{subfigure}
\begin{subfigure}[b]{0.17\textwidth}
    \includegraphics [trim=190 80 180 70, clip, width=1.0\textwidth]{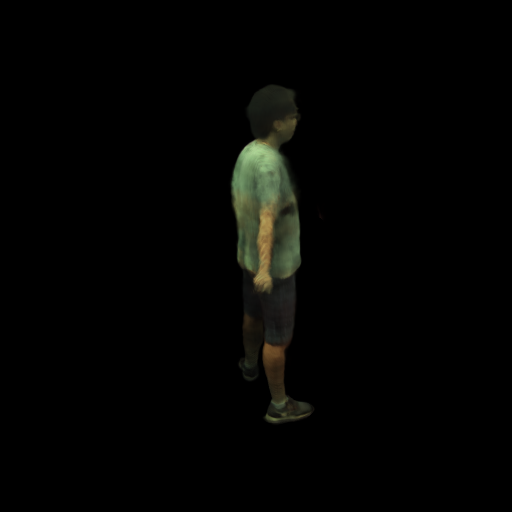}
\end{subfigure}
\begin{subfigure}[b]{0.17\textwidth}
    \includegraphics [trim=190 80 180 70, clip, width=1.0\textwidth]{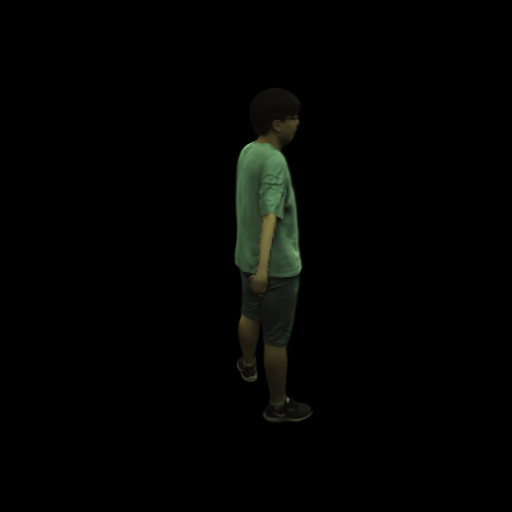}
\end{subfigure}
\begin{subfigure}[b]{0.17\textwidth}
    \includegraphics [trim=190 80 180 70, clip, width=1.0\textwidth]{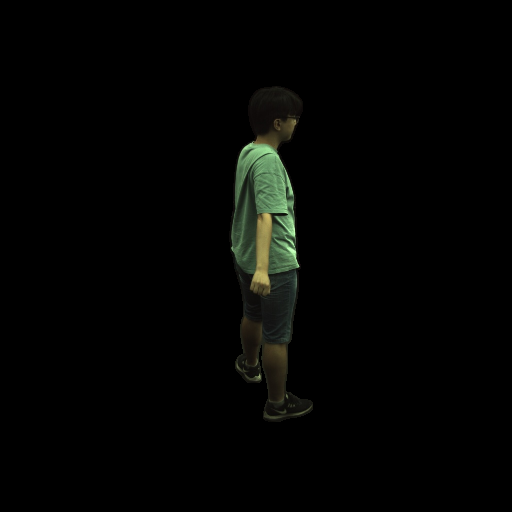}
\end{subfigure} \\
\begin{subfigure}[b]{0.17\textwidth}
    \includegraphics [trim=175 120 190 80, clip, width=1.0\textwidth]{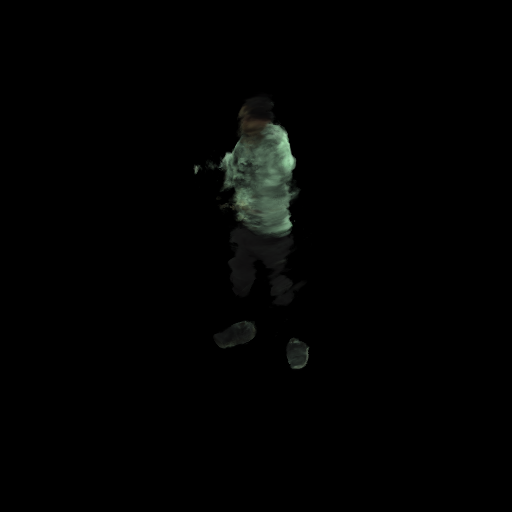}
\end{subfigure}
\begin{subfigure}[b]{0.17\textwidth}
    \includegraphics [trim=175 120 190 80, clip, width=1.0\textwidth]{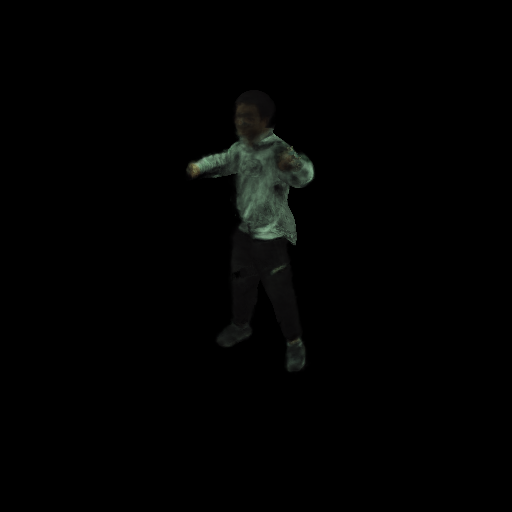}
\end{subfigure}
\begin{subfigure}[b]{0.17\textwidth}
    \includegraphics [trim=175 120 190 80, clip, width=1.0\textwidth]{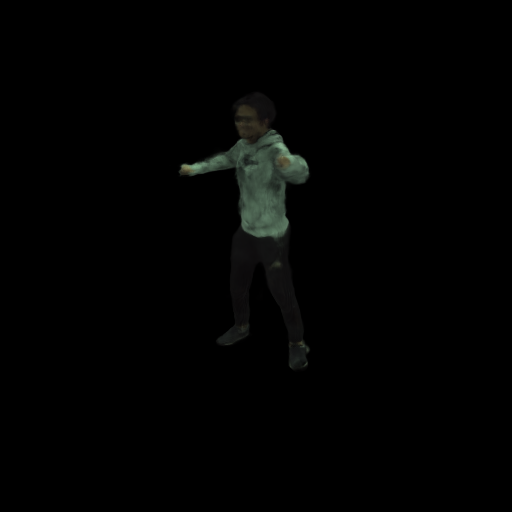}
\end{subfigure}
\begin{subfigure}[b]{0.17\textwidth}
    \includegraphics [trim=175 120 190 80, clip, width=1.0\textwidth]{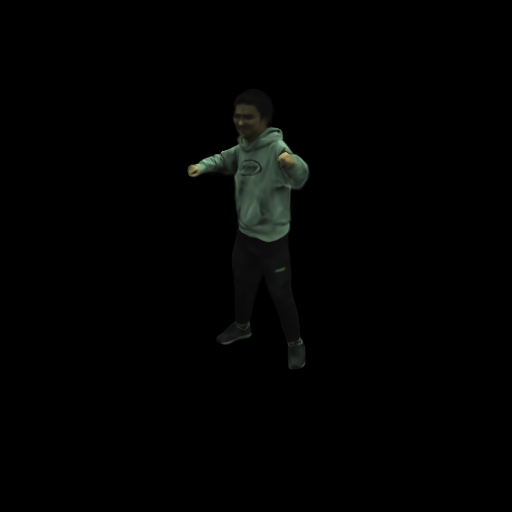}
\end{subfigure}
\begin{subfigure}[b]{0.17\textwidth}
    \includegraphics [trim=175 120 190 80, clip, width=1.0\textwidth]{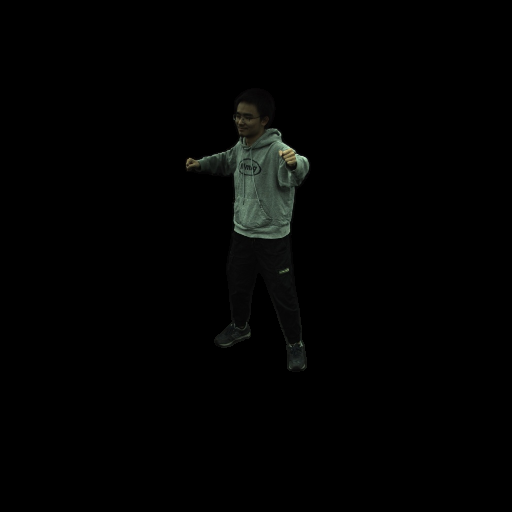}
\end{subfigure} \\
\begin{subfigure}[b]{0.17\textwidth}
    \includegraphics [trim=160 110 190 49, clip, width=1.0\textwidth]{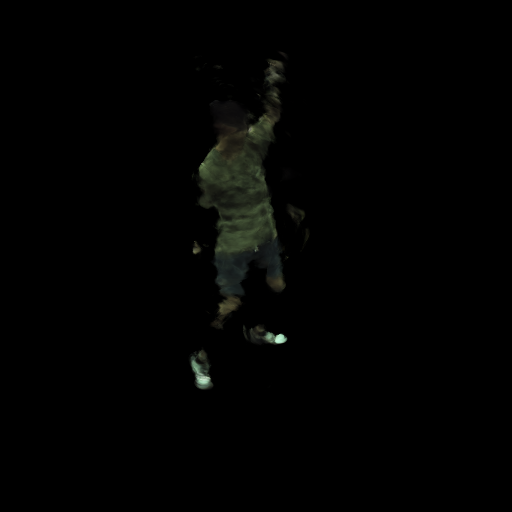}
    \caption{A-NeRF}
\end{subfigure}
\begin{subfigure}[b]{0.17\textwidth}
    \includegraphics [trim=160 110 190 49, clip, width=1.0\textwidth]{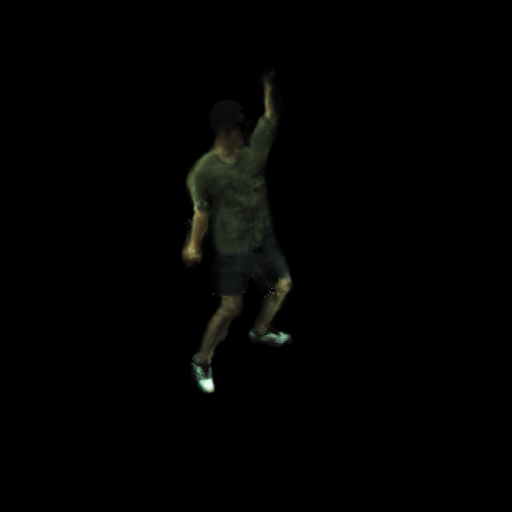}
    \caption{Ani-NeRF}
\end{subfigure}
\begin{subfigure}[b]{0.17\textwidth}
    \includegraphics [trim=160 110 190 49, clip, width=1.0\textwidth]{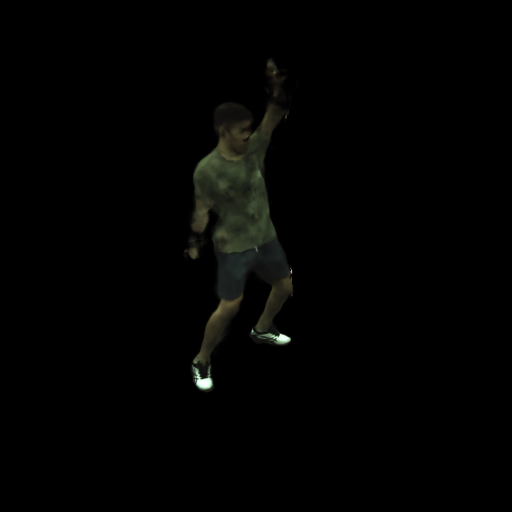}
    \caption{Neural Body}
\end{subfigure}
\begin{subfigure}[b]{0.17\textwidth}
    \includegraphics [trim=160 110 190 49, clip, width=1.0\textwidth]{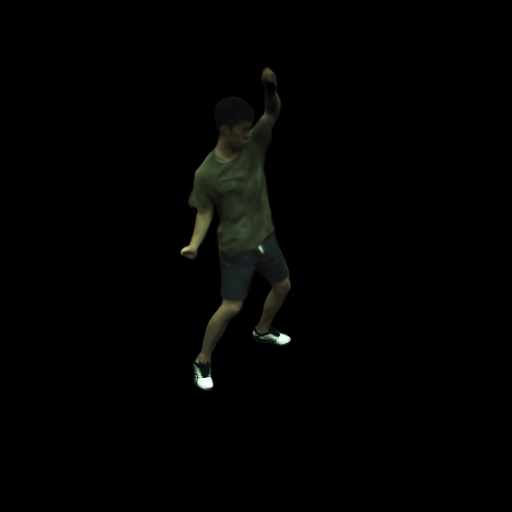}
    \caption{Ours}
\end{subfigure}
\begin{subfigure}[b]{0.17\textwidth}
    \includegraphics [trim=160 110 190 49, clip, width=1.0\textwidth]{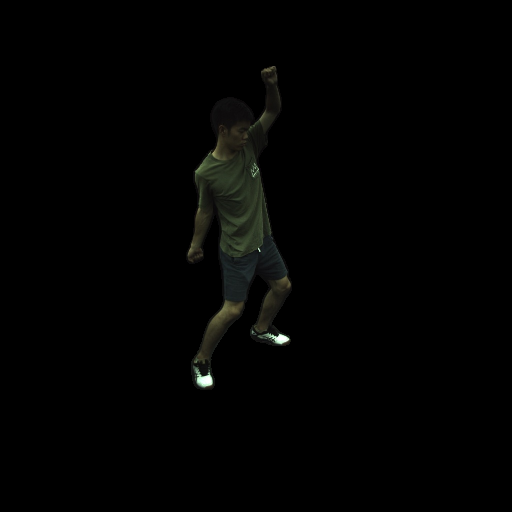}
    \caption{GT}
\end{subfigure}
\caption{\textbf{Additional Generalization Results on ZJU-MoCap Test Poses.}}
\label{appx:fig:qualitative_results_test}
\end{figure}

We show additional qualitative results on ZJU-MoCap test poses in Fig.~\ref{appx:fig:qualitative_results_test}. Similar to the results presented in the main paper, A-NeRF and Neural Body do not generalize to these within-distribution poses. Ani-NeRF produces noisy rendering due to its inaccurate backward LBS network. Note that since these results are pose extrapolations, it is not possible to reproduce the exact color and texture of ground-truth images. Still, our approach does not suffer from the artifacts that baselines have demonstrated, resulting in better metrics, especially for LPIPS (Table~\ref{appx:tab:novel_pose}). We present more qualitative results in the supplementary video.

\subsection{Additional Qualitative Results on Out-of-distribution Poses}
\label{appx:additional_qualitative_odp}

\begin{figure}[t]
\captionsetup[subfigure]{labelformat=empty}
\centering
\begin{subfigure}[b]{0.17\textwidth}
    \includegraphics [trim=170 90 170 80, clip, width=1.0\textwidth]{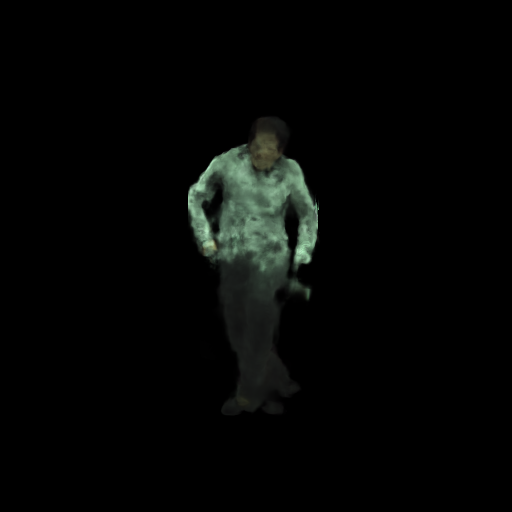}
\end{subfigure}
\begin{subfigure}[b]{0.17\textwidth}
    \includegraphics [trim=170 90 170 80, clip, width=1.0\textwidth]{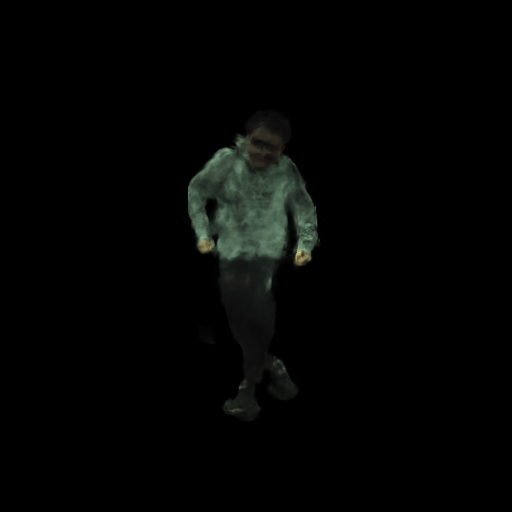}
\end{subfigure}
\begin{subfigure}[b]{0.17\textwidth}
    \includegraphics [trim=130 60 200 90, clip, width=1.0\textwidth]{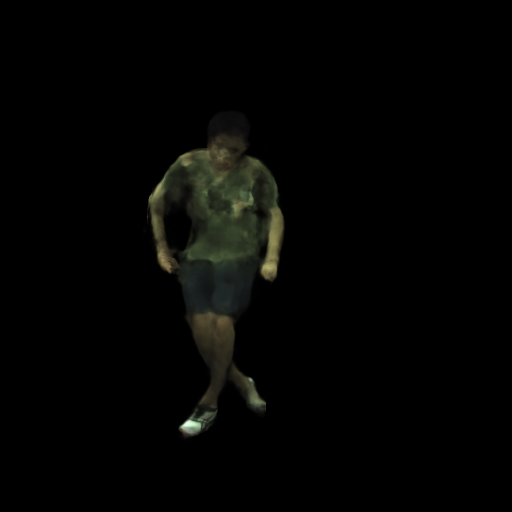}
\end{subfigure}
\begin{subfigure}[b]{0.17\textwidth}
    \includegraphics [trim=165 80 175 90, clip, width=1.0\textwidth]{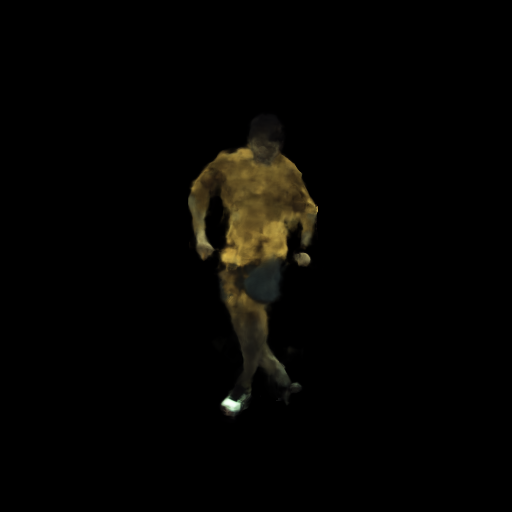}
\end{subfigure}
\begin{subfigure}[b]{0.17\textwidth}
    \includegraphics [trim=165 80 175 90, clip, width=1.0\textwidth]{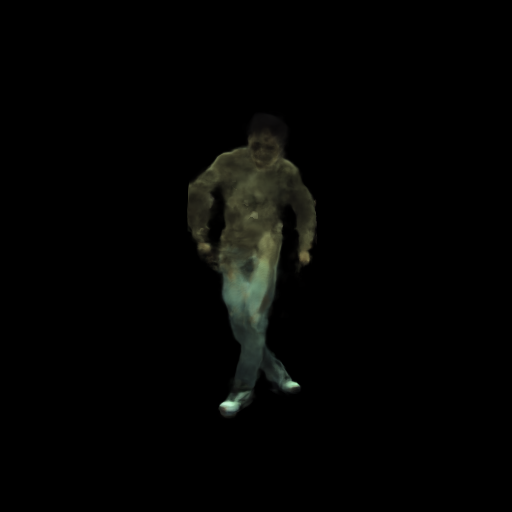}
\end{subfigure} \\
\begin{subfigure}[b]{0.17\textwidth}
    \includegraphics [trim=170 90 170 80, clip, width=1.0\textwidth]{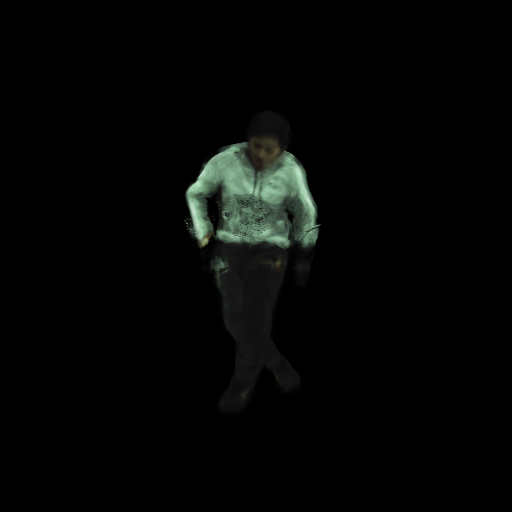}
\end{subfigure}
\begin{subfigure}[b]{0.17\textwidth}
    \includegraphics [trim=170 90 170 80, clip, width=1.0\textwidth]{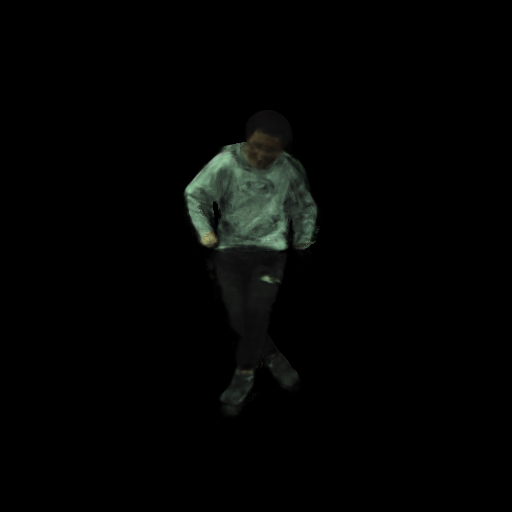}
\end{subfigure}
\begin{subfigure}[b]{0.17\textwidth}
    \includegraphics [trim=130 60 200 90, clip, width=1.0\textwidth]{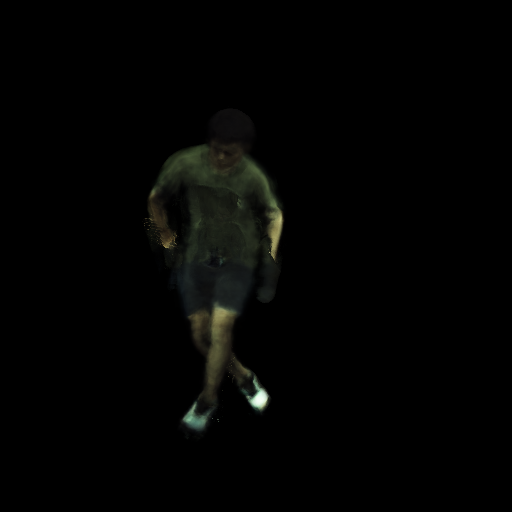}
\end{subfigure}
\begin{subfigure}[b]{0.17\textwidth}
    \includegraphics [trim=165 80 175 90, clip, width=1.0\textwidth]{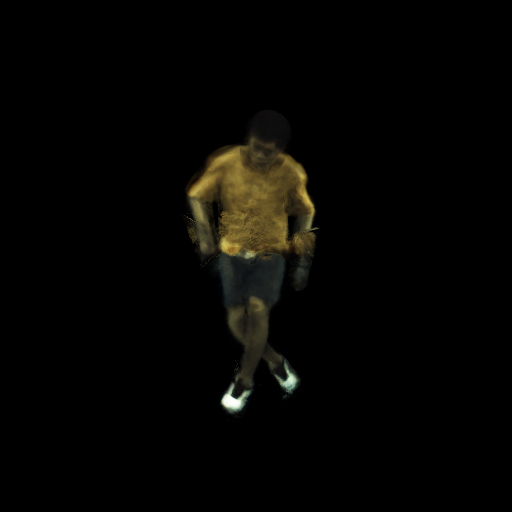}
\end{subfigure}
\begin{subfigure}[b]{0.17\textwidth}
    \includegraphics [trim=165 80 175 90, clip, width=1.0\textwidth]{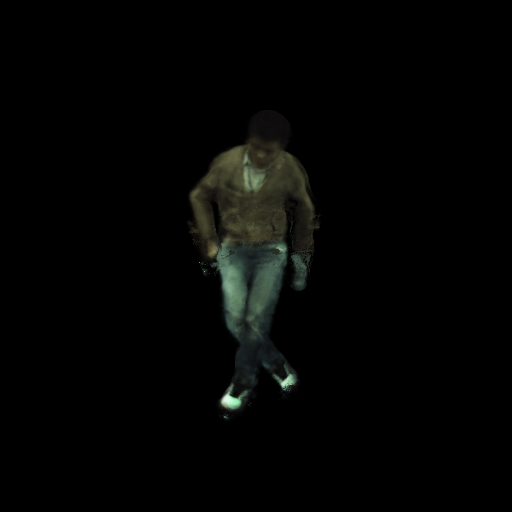}
\end{subfigure} \\
\begin{subfigure}[b]{0.17\textwidth}
    \includegraphics [trim=170 90 170 80, clip, width=1.0\textwidth]{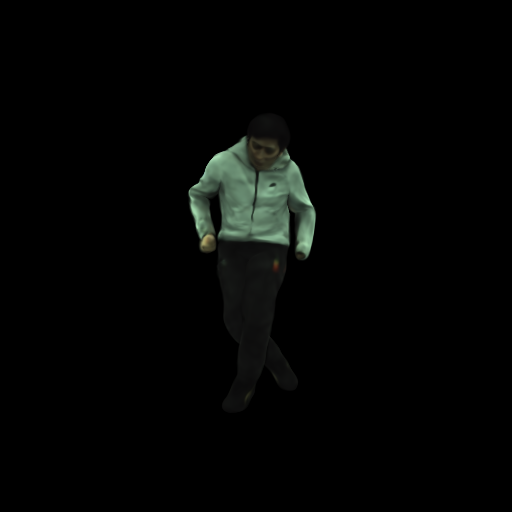}
\end{subfigure}
\begin{subfigure}[b]{0.17\textwidth}
    \includegraphics [trim=170 90 170 80, clip, width=1.0\textwidth]{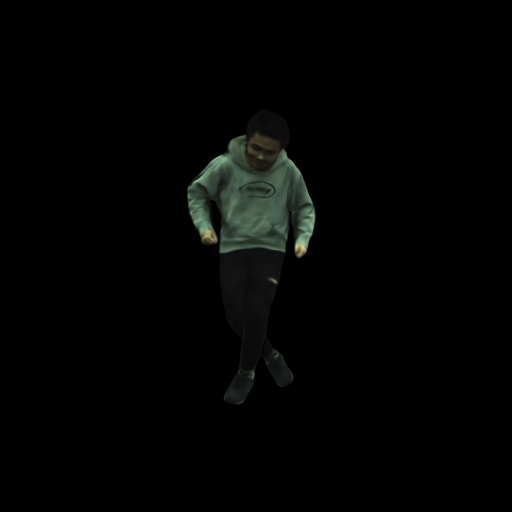}
\end{subfigure}
\begin{subfigure}[b]{0.17\textwidth}
    \includegraphics [trim=130 60 200 90, clip, width=1.0\textwidth]{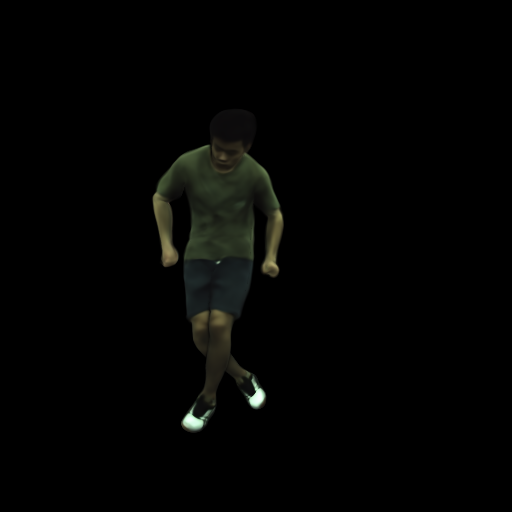}
\end{subfigure}
\begin{subfigure}[b]{0.17\textwidth}
    \includegraphics [trim=165 80 175 90, clip, width=1.0\textwidth]{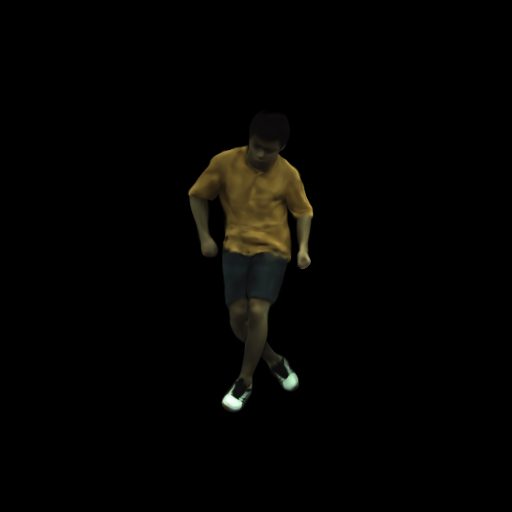}
\end{subfigure}
\begin{subfigure}[b]{0.17\textwidth}
    \includegraphics [trim=165 80 175 90, clip, width=1.0\textwidth]{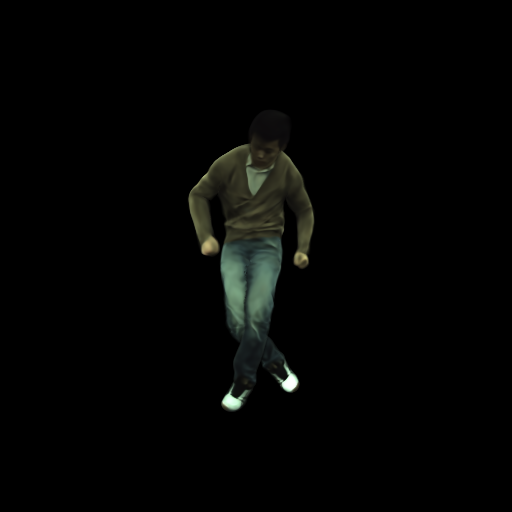}
\end{subfigure} \\
\begin{subfigure}[b]{0.17\textwidth}
    \includegraphics [trim=170 90 170 80, clip, width=1.0\textwidth]{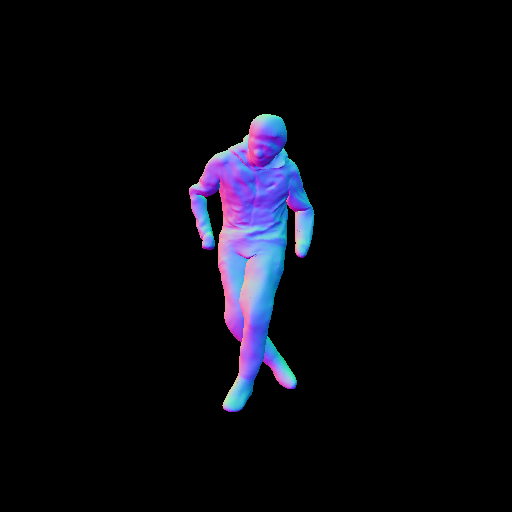}
\end{subfigure}
\begin{subfigure}[b]{0.17\textwidth}
    \includegraphics [trim=170 90 170 80, clip, width=1.0\textwidth]{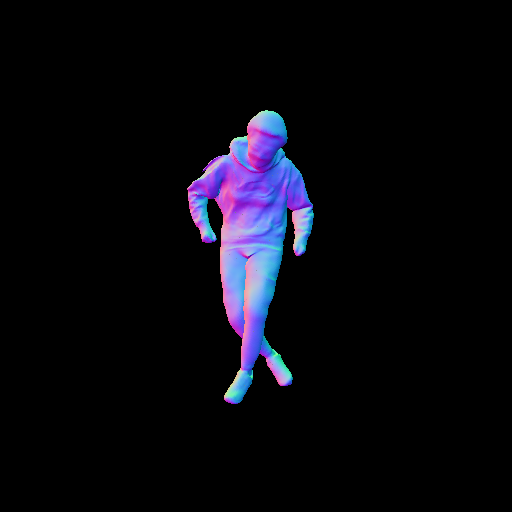}
\end{subfigure}
\begin{subfigure}[b]{0.17\textwidth}
    \includegraphics [trim=130 60 200 90, clip, width=1.0\textwidth]{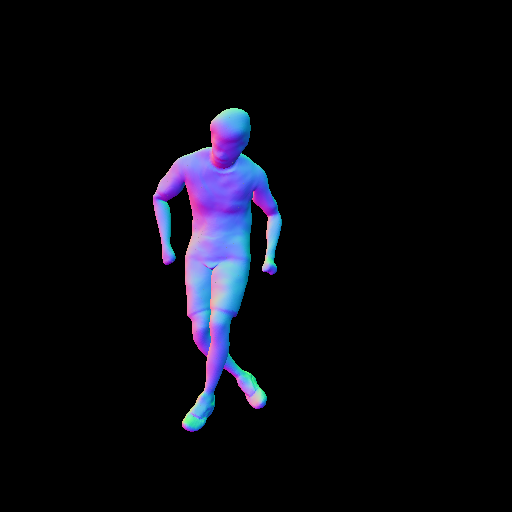}
\end{subfigure}
\begin{subfigure}[b]{0.17\textwidth}
    \includegraphics [trim=165 80 175 90, clip, width=1.0\textwidth]{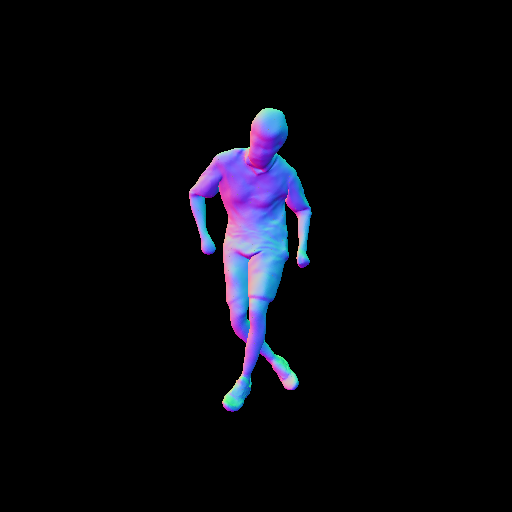}
\end{subfigure}
\begin{subfigure}[b]{0.17\textwidth}
    \includegraphics [trim=165 80 175 90, clip, width=1.0\textwidth]{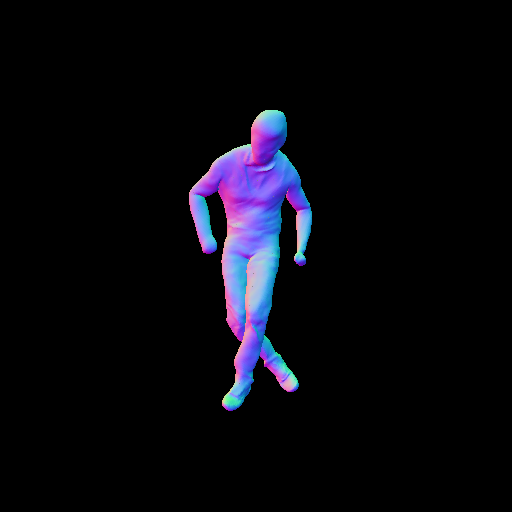}
\end{subfigure}
\caption{\textbf{Additional Generalization Results on Out-of-distribution Poses.} From top to bottom: Neural Body, Ani-NeRF, ours, and our geometry.}
\label{appx:fig:qualitative_results_odp}
\end{figure}
We show additional qualitative results on out-of-distribution poses~\cite{aist++:ICCV:2021} in Fig.~\ref{appx:fig:qualitative_results_odp}. We present more results in the supplementary video.

\subsection{Closest Training Poses to Out-of-distribution Poses}
\label{appx:additional_qualitative_closest_training}
\begin{figure}[t]
\captionsetup[subfigure]{labelformat=empty}
\centering
\begin{subfigure}[b]{1.0\textwidth}
    \includegraphics [width=1.0\textwidth]{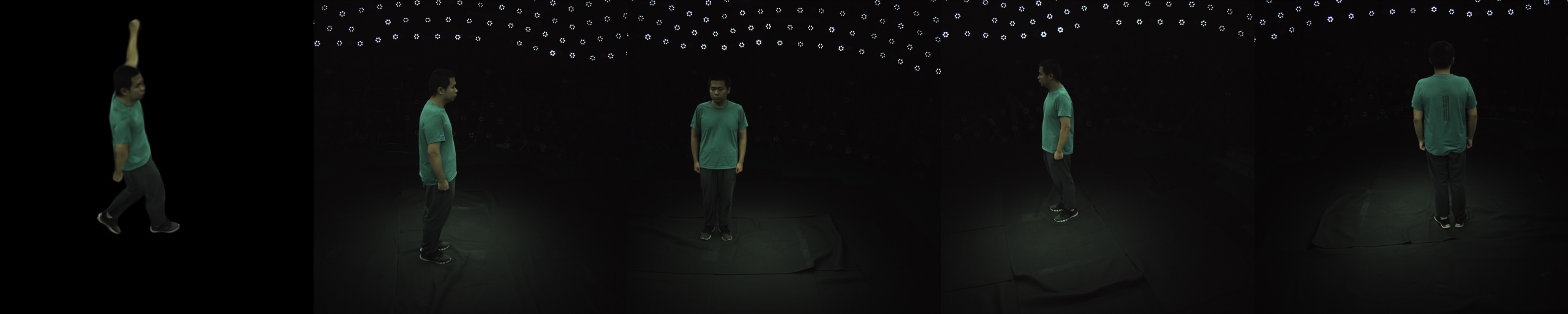}
\end{subfigure}
\begin{subfigure}[b]{1.0\textwidth}
    \includegraphics [width=1.0\textwidth]{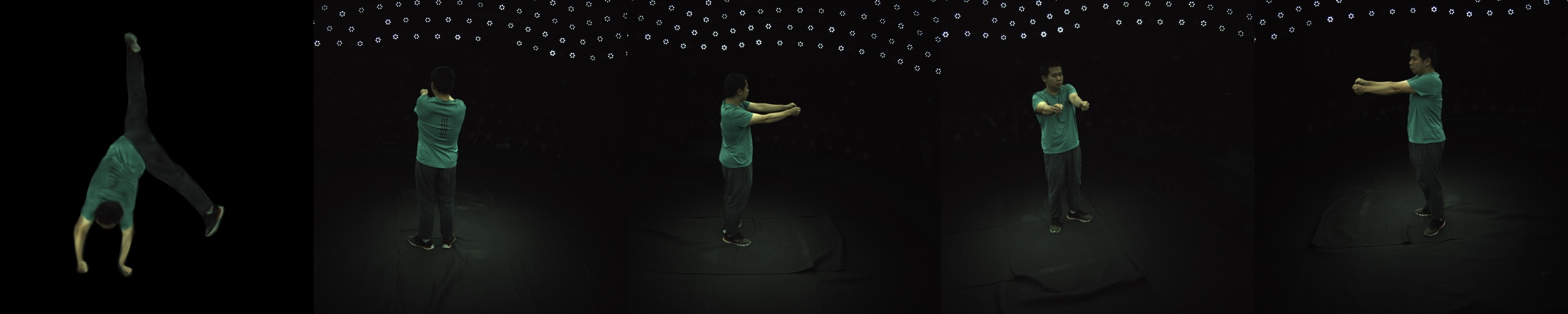}
\end{subfigure}
\begin{subfigure}[b]{1.0\textwidth}
    \includegraphics [width=1.0\textwidth]{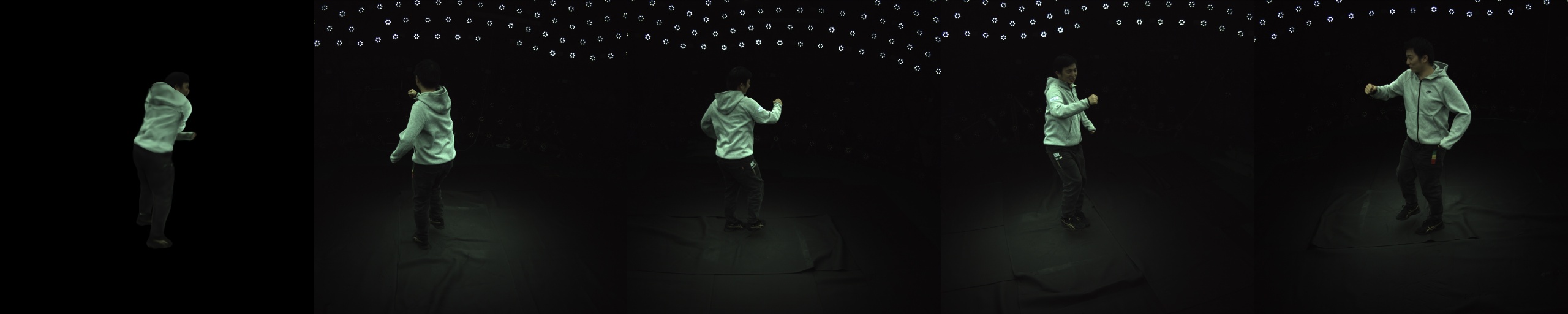}
\end{subfigure}
\begin{subfigure}[b]{1.0\textwidth}
    \includegraphics [width=1.0\textwidth]{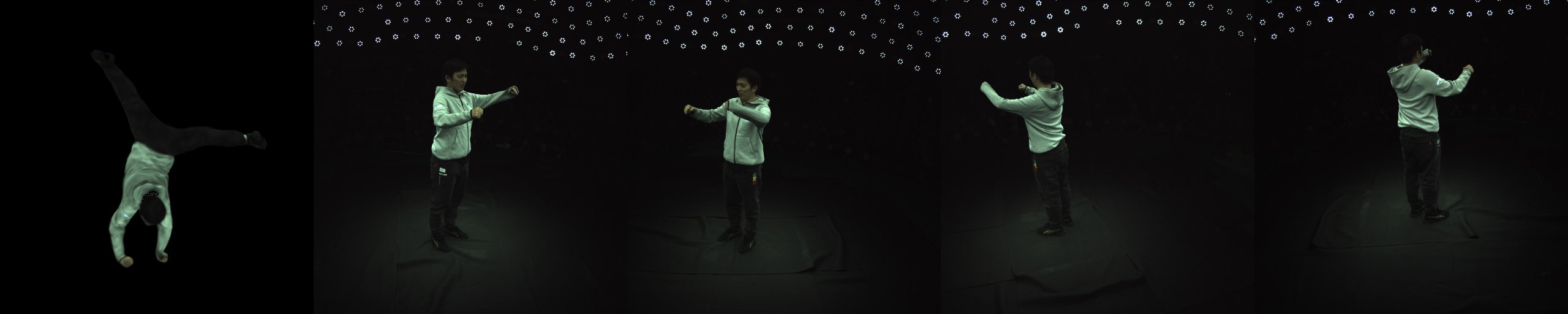}
\end{subfigure}
\caption{\textbf{Closest Training Poses to Out-of-distribution Test Poses.} We show rendering results of out-of-distribution poses on the left-most column, while demonstrating 4 training images of the closest training pose to each of the test poses.}
\label{appx:fig:qualitative_results_closest_training}
\end{figure}

To further demonstrate the generalization ability of our approach, we also visualize the closest training pose from the ZJU-MoCap dataset to out-of-distribution test poses from the AIST++ dataset and the AMASS dataset in Fig.~\ref{appx:fig:qualitative_results_closest_training}. To find the closest training pose to a test pose, we convert local poses (\ie\ all pose vectors excluding global orientation) to a matrix representation and find the closest training pose with nearest neighbor search using the converted matrix representation.

\subsection{Qualitative Results on Models Trained with Monocular Videos}
\label{appx:additional_qualitative_mono}
\begin{figure}[t]
\captionsetup[subfigure]{labelformat=empty}
\centering
\begin{subfigure}[b]{0.11\textwidth}
    \includegraphics [trim=180 100 140 50, clip, width=1.0\textwidth]{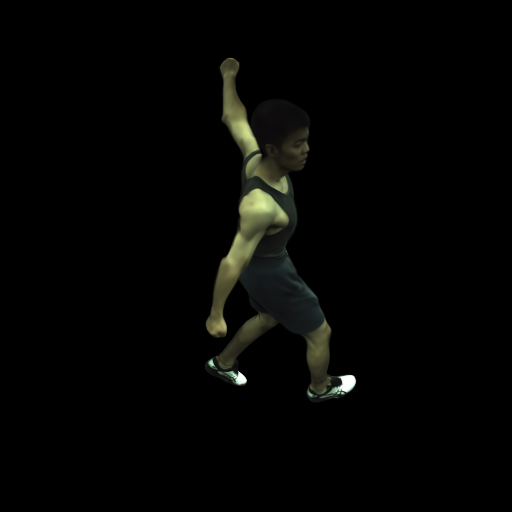}
\end{subfigure}
\begin{subfigure}[b]{0.11\textwidth}
    \includegraphics [trim=180 100 140 50, clip, width=1.0\textwidth]{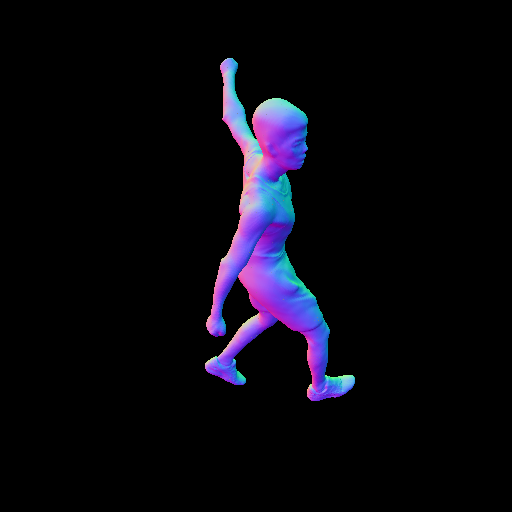}
\end{subfigure}
\begin{subfigure}[b]{0.11\textwidth}
    \includegraphics [trim=140 110 180 40, clip, width=1.0\textwidth]{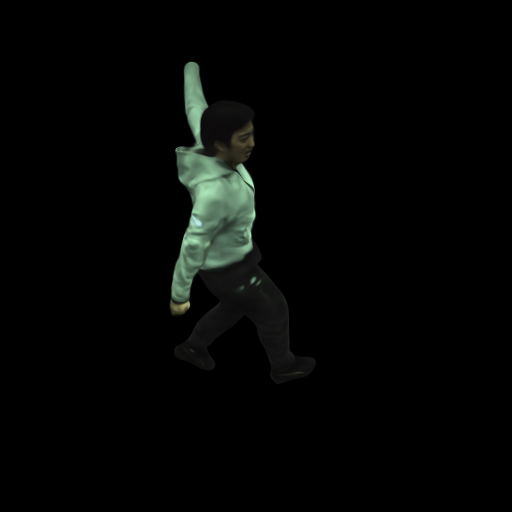}
\end{subfigure}
\begin{subfigure}[b]{0.11\textwidth}
    \includegraphics [trim=140 110 180 40, clip, width=1.0\textwidth]{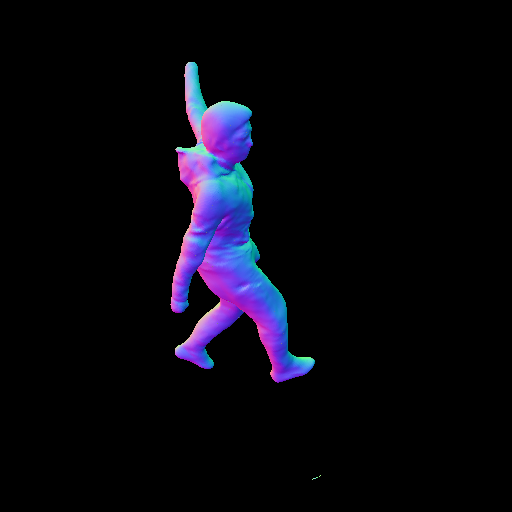}
\end{subfigure}
\begin{subfigure}[b]{0.11\textwidth}
    \includegraphics [trim=180 100 140 50, clip, width=1.0\textwidth]{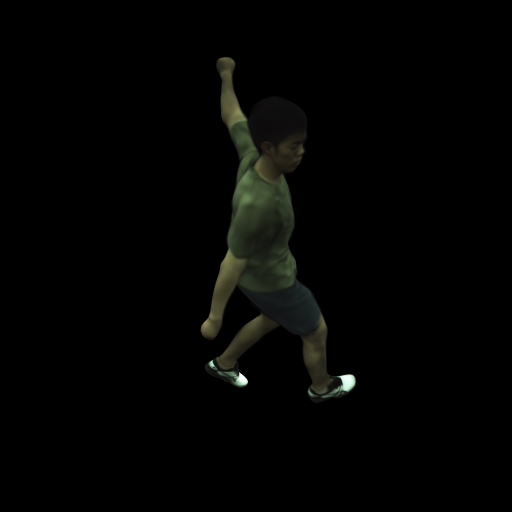}
\end{subfigure}
\begin{subfigure}[b]{0.11\textwidth}
    \includegraphics [trim=180 100 140 50, clip, width=1.0\textwidth]{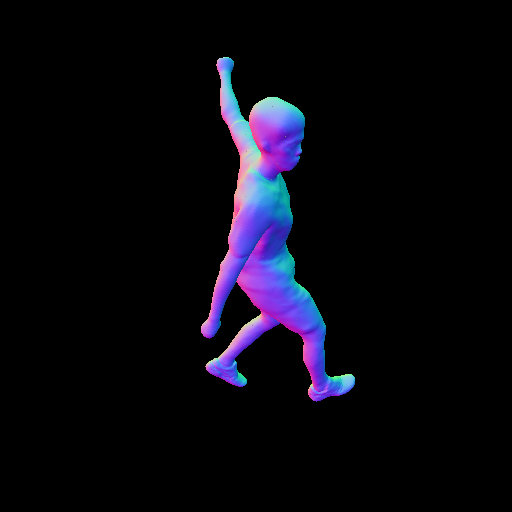}
\end{subfigure}
\begin{subfigure}[b]{0.11\textwidth}
    \includegraphics [trim=140 110 180 40, clip, width=1.0\textwidth]{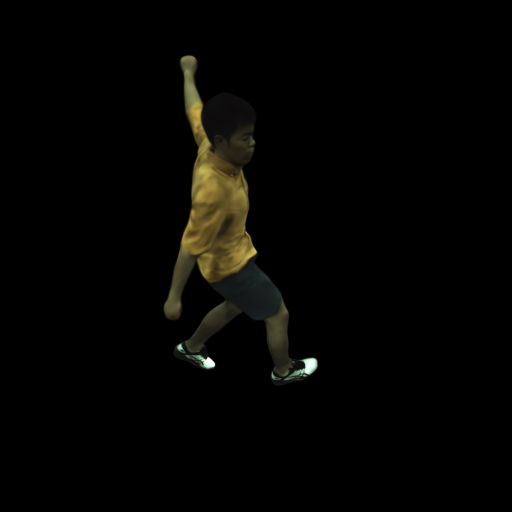}
\end{subfigure}
\begin{subfigure}[b]{0.11\textwidth}
    \includegraphics [trim=140 110 180 40, clip, width=1.0\textwidth]{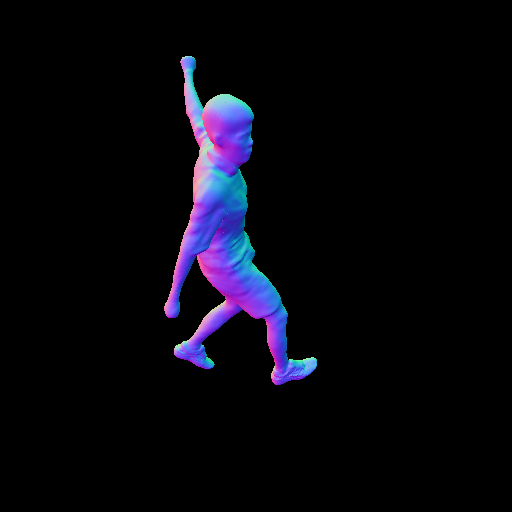}
\end{subfigure} \\
\begin{subfigure}[b]{0.11\textwidth}
    \includegraphics [trim=140 40 150 40, clip, width=1.0\textwidth]{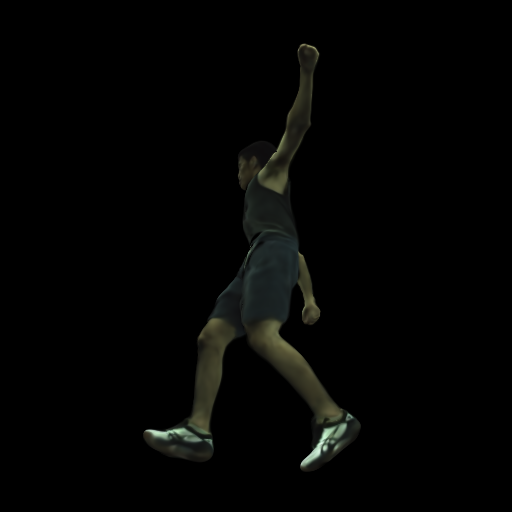}
\end{subfigure}
\begin{subfigure}[b]{0.11\textwidth}
    \includegraphics [trim=140 40 150 40, clip, width=1.0\textwidth]{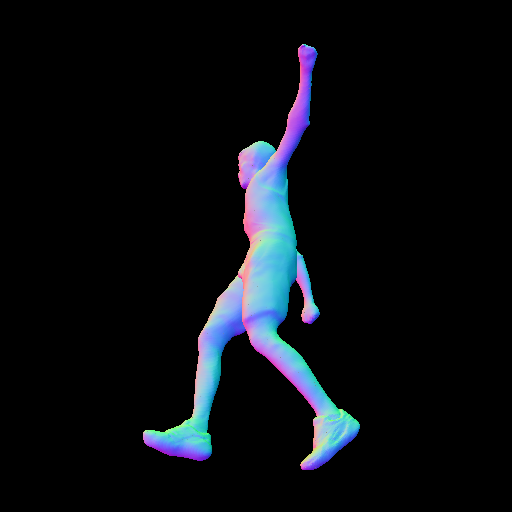}
\end{subfigure}
\begin{subfigure}[b]{0.11\textwidth}
    \includegraphics [trim=100 40 190 40, clip, width=1.0\textwidth]{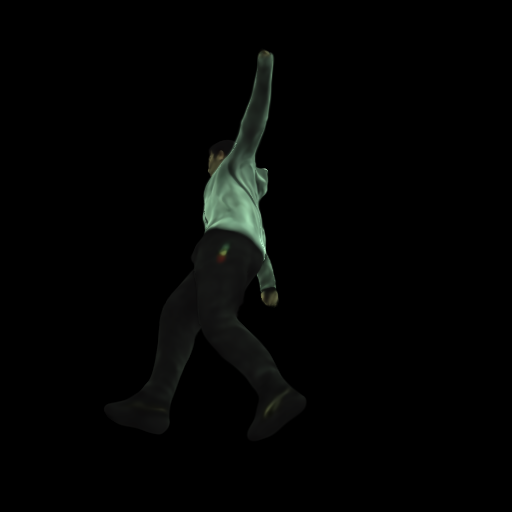}
\end{subfigure}
\begin{subfigure}[b]{0.11\textwidth}
    \includegraphics [trim=100 40 190 40, clip, width=1.0\textwidth]{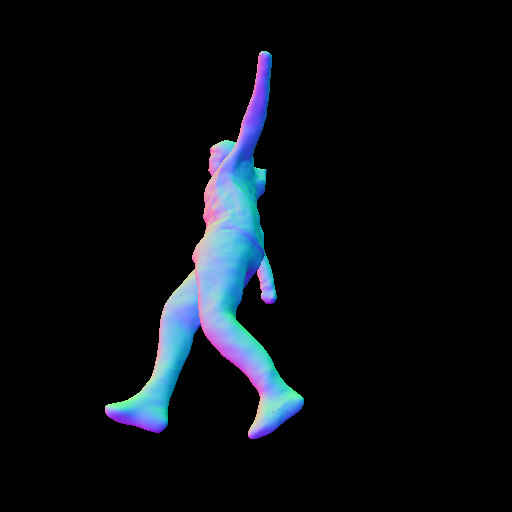}
\end{subfigure}
\begin{subfigure}[b]{0.11\textwidth}
    \includegraphics [trim=140 40 150 40, clip, width=1.0\textwidth]{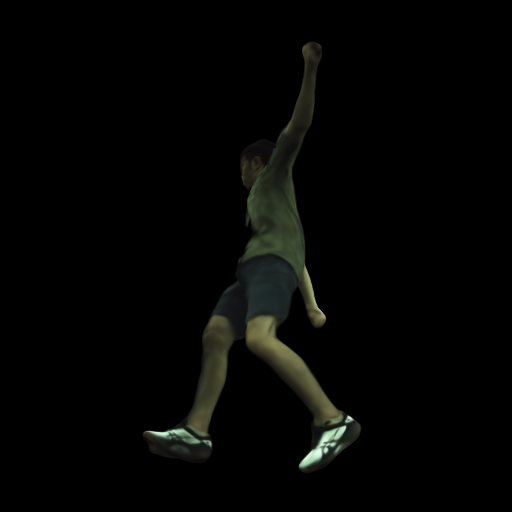}
\end{subfigure}
\begin{subfigure}[b]{0.11\textwidth}
    \includegraphics [trim=140 40 150 40, clip, width=1.0\textwidth]{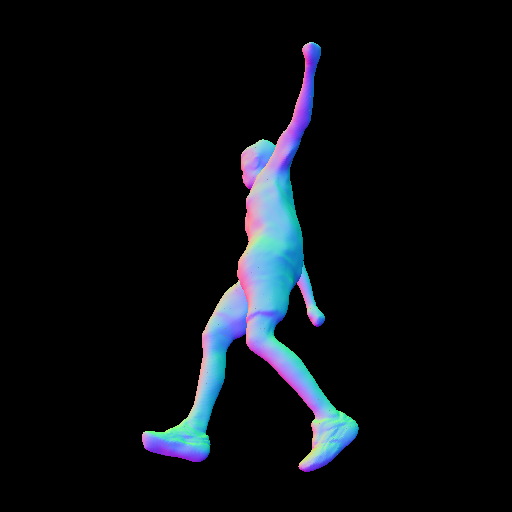}
\end{subfigure}
\begin{subfigure}[b]{0.11\textwidth}
    \includegraphics [trim=100 50 190 30, clip, width=1.0\textwidth]{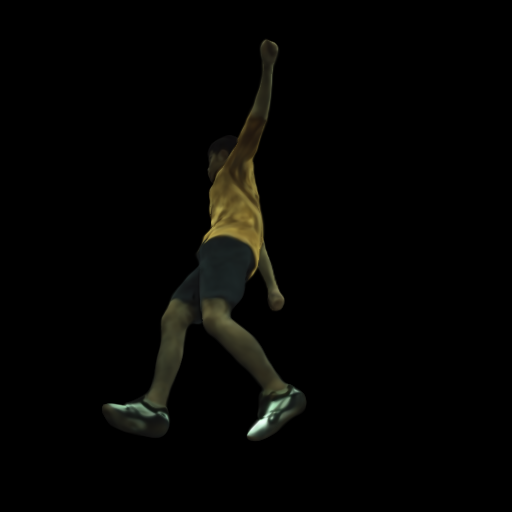}
\end{subfigure}
\begin{subfigure}[b]{0.11\textwidth}
    \includegraphics [trim=100 50 190 30, clip, width=1.0\textwidth]{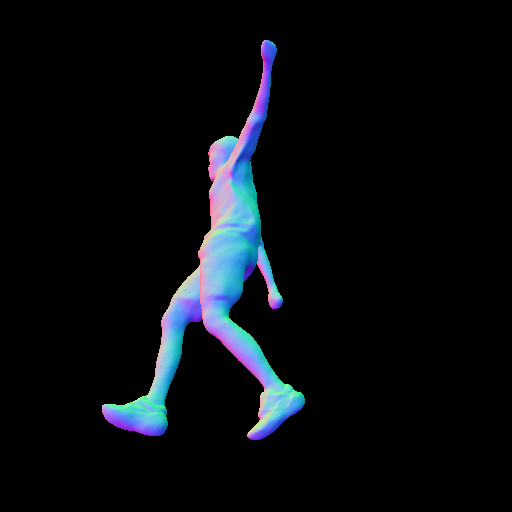}
\end{subfigure}
\caption{\textbf{Generalization to AIST++~\cite{aist++:ICCV:2021} Poses with Models Trained from Monocular Videos.}}
\label{appx:fig:qualitative_results_mono}
\end{figure}
In this subsection, we present models trained on monocular videos. For this monocular setup, we use only the first camera of the ZJU-MoCap dataset to train our models. We do not modify our approach and all hyperparameters remain the same as the multi-view setup. We train each model for 500 epochs on 500 frames of selected sequences in which the subjects do repetitive motions while rotating roughly 360 degrees. We animate the trained model with out-of-distribution poses from AIST++~\cite{aist++:ICCV:2021}. Qualitative results are shown in Fig.~\ref{appx:fig:qualitative_results_mono}. Even under this extreme setup, our approach can still learn avatars with plausible geometry/appearance and the avatars still generalize to out-of-distribution poses. For the complete animation sequences, please see our supplementary video.
\section{Limitations}
\label{appx:limitations}
As reported in Section~\ref{appx:implementation}, our approach is relatively slow at inference time. The major bottlenecks are the iterative root-finding~\cite{Chen2021ICCV} and the volume rendering.

\begin{figure}[t]
\captionsetup[subfigure]{labelformat=empty}
\scriptsize
\centering
\begin{subfigure}[b]{0.16\textwidth}
    \includegraphics [trim=130 170 190 70, clip, width=1.0\textwidth]{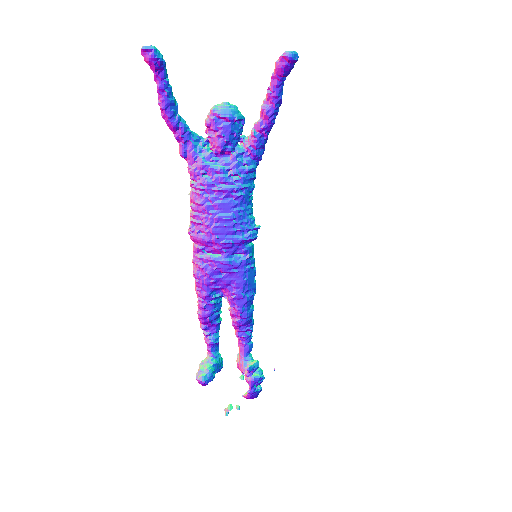}
    \caption{A-NeRF}
\end{subfigure}
\begin{subfigure}[b]{0.16\textwidth}
    \includegraphics [trim=130 170 190 70, clip, width=1.0\textwidth]{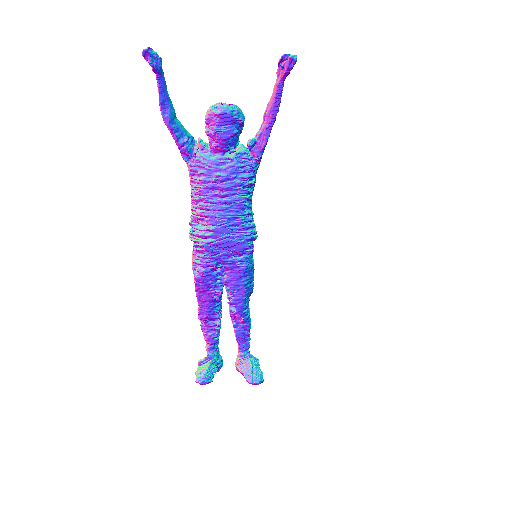}
    \caption{Ani-NeRF}
\end{subfigure}
\begin{subfigure}[b]{0.16\textwidth}
    \includegraphics [trim=130 170 190 70, clip, width=1.0\textwidth]{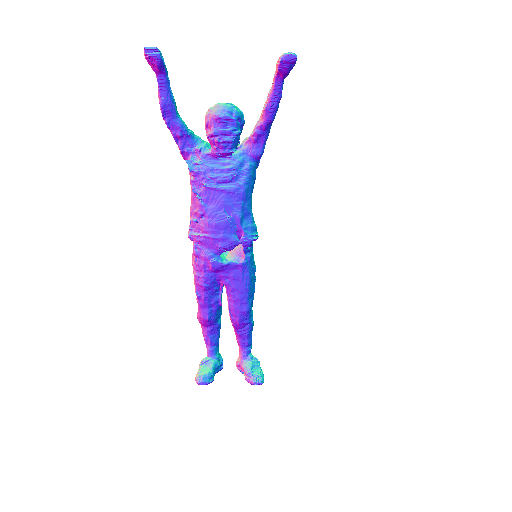}
    \caption{Neural Body}
\end{subfigure}
\begin{subfigure}[b]{0.16\textwidth}
    \includegraphics [trim=130 170 190 70, clip, width=1.0\textwidth]{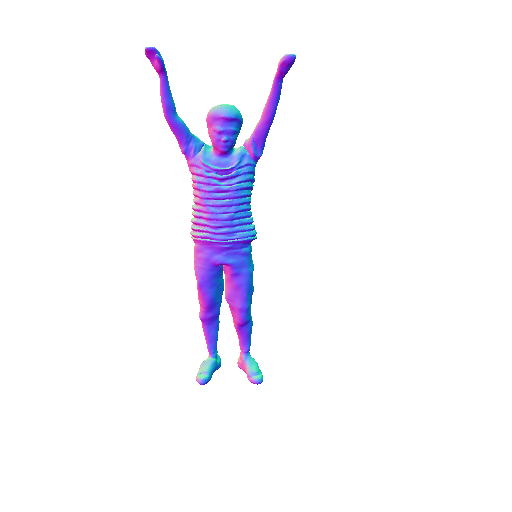}
    \caption{Ours}
\end{subfigure}
\begin{subfigure}[b]{0.16\textwidth}
    \includegraphics [trim=130 170 190 70, clip, width=1.0\textwidth]{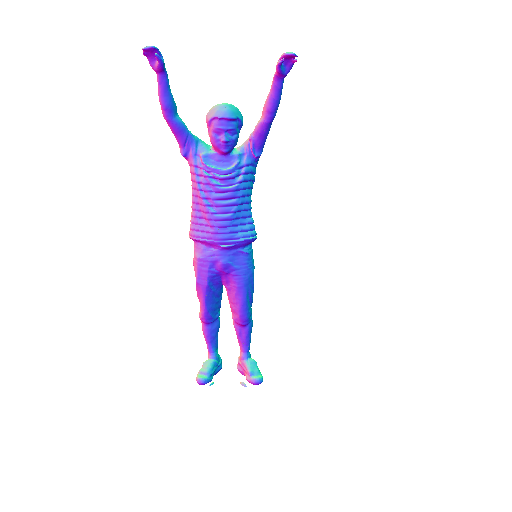}
    \caption{GT (NeuS)}
\end{subfigure}
\begin{subfigure}[b]{0.16\textwidth}
    \includegraphics [trim=130 170 190 70, clip, width=1.0\textwidth]{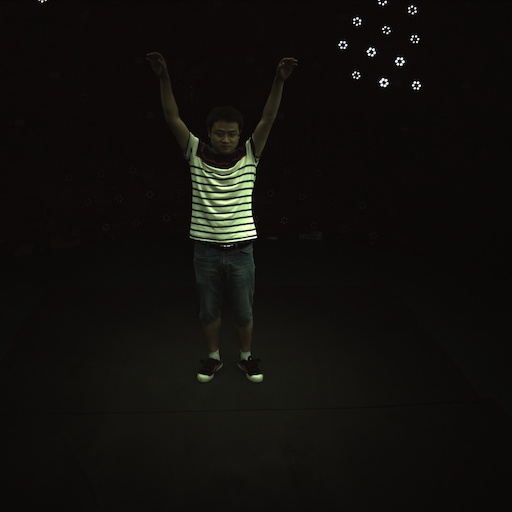}
    \caption{Image}
\end{subfigure}
\caption{\textbf{Shape-Appearance Ambiguity}. The Neural Rendering-based reconstruction tends to bake complex textures into the geometry, resulting in a biased geometry reconstruction.}
\label{appx:fig:limitation}
\end{figure}

Another limitation is that neural rendering-based reconstruction methods tend to overfit the geometry to the texture, resulting in a reconstruction bias. As shown in Fig.~\ref{appx:fig:limitation}, while NeRF-based baselines are unable to recover detailed wrinkles, SDF-based rendering (ours and NeuS) wrongfully reconstructs the stripes on the shirt as part of the geometry. Note that A-NeRF and Ani-NeRF also suffer from this kind of bias. Neural Body demonstrates less overfitting effects. We hypothesize that this is because the structured latent codes in Neural Body are local in space and thus give the color network more flexibility, making the density network less prone to overfitting. Still, Neural Body gives noisy reconstructions and cannot generalize to unseen poses. Resolving this reconstruction bias while maintaining a clean geometry is an interesting avenue for future research.

\end{document}